\documentclass{article}

\newif\ifvenuestyle
\IfFileExists{iclr2027_conference.sty}{
  \venuestyletrue
  \usepackage{iclr2027_conference,times}
}{
  \venuestylefalse
  \usepackage[margin=1in]{geometry}
  \usepackage{times}
}
\usepackage{amsmath,amssymb}
\ifdefined\XeTeXversion
  \InputIfFileExists{omscmsy.fd}{}{}
  \DeclareFontShape{OMS}{cmsy}{m}{n}{%
    <5><6><7><8><9><10>cmsy10
    <10.95><12><14.4><17.28><20.74><24.88>cmsy10}{}
  \InputIfFileExists{omlcmm.fd}{}{}
  \DeclareFontShape{OML}{cmm}{m}{it}{%
    <7>cmmi7
    <5><6><8><9><10>cmmi10
    <10.95><12><14.4><17.28><20.74><24.88>cmmi10}{}
\fi
\usepackage{booktabs,graphicx}
\usepackage{needspace}
\usepackage{subcaption}
\usepackage{xcolor}
\usepackage{tabularx,multirow,colortbl}
\usepackage{microtype}
\usepackage{url}
\usepackage[hidelinks]{hyperref}

\ifvenuestyle
  \iclrfinalcopy
\fi
\hypersetup{
  pdftitle={Reconstruction as Spatial Evaluation: Diagnosing Cross-View Spatial Intelligence in Large Vision-Language Models},
  pdfauthor={Junfei Wu, Qiuyue Wang, Xuhong Huang, Qihang Peng, Mingsheng Li, Jian Guan, Qiang Liu, Shu Wu, Liang Wang, Shuai Bai, Tieniu Tan}
}

\newcommand{\bench}{\textsc{ReconSpatial}}

\newcommand{\numatomic}{1,388}
\newcommand{\numrecon}{840}
\newcommand{\numnavigation}{346}

\newcommand{\nummodels}{18}
\newcommand{\diagmodel}{GPT-5.6-Sol}

\definecolor{caseInk}{HTML}{263E55}
\definecolor{caseMuted}{HTML}{68798A}
\definecolor{caseImage}{HTML}{EEF4FA}
\definecolor{casePrompt}{HTML}{F4F1F8}
\definecolor{caseAnswer}{HTML}{EDF6F0}
\definecolor{caseRule}{HTML}{D4DEE8}
\DeclareFontFamily{OT1}{casecm}{}
\DeclareFontShape{OT1}{casecm}{m}{n}{<->cmr10}{}
\DeclareFontShape{OT1}{casecm}{b}{n}{<->cmbx10}{}
\DeclareFontShape{OT1}{casecm}{bx}{n}{<->cmbx10}{}
\newcommand{\casefont}{\fontencoding{OT1}\fontfamily{casecm}\selectfont}
\newcounter{caseexample}
\newsavebox{\caseexamplebox}

\definecolor{casePaper}{HTML}{F5F6F8}
\definecolor{caseOk}{HTML}{1F7A52}
\definecolor{caseLabel}{HTML}{6B7A8A}
\newcommand{\casemono}{\fontencoding{OT1}\fontfamily{cmtt}\fontsize{6.9}{8.3}\selectfont}

\newcommand{\casesection}[1]{%
  \par\vspace{4pt}\noindent{\sffamily\scriptsize\bfseries\color{caseLabel}\MakeUppercase{#1}}\par\vspace{1pt}}
\newenvironment{promptbox}{%
  \par\noindent\begin{lrbox}{\caseexamplebox}%
  \begin{minipage}{\dimexpr\linewidth-12pt\relax}%
  \sffamily\fontsize{7.6}{9.2}\selectfont\color{caseInk}\raggedright
  \setlength{\parindent}{0pt}\setlength{\parskip}{2pt}}%
  {\end{minipage}\end{lrbox}%
  {\setlength{\fboxsep}{6pt}\colorbox{casePaper}{\usebox{\caseexamplebox}}}\par}
\newcommand{\caseoptions}[1]{\par\vspace{1pt}#1\par}
\newcommand{\caseright}[1]{{\setlength{\fboxsep}{1.5pt}\hspace*{-1.5pt}\colorbox{caseOk!14}{\color{caseOk}#1\,\checkmark}}}
\newcommand{\caseanswer}[1]{\noindent{\sffamily\small\bfseries\color{caseOk}#1}\par}
\newcommand{\caseanswercode}[1]{\noindent{\casemono\color{caseOk}#1}\par}
\newcommand{\reconobj}[1]{{\setlength{\fboxsep}{0.8pt}\colorbox{reconobj#1}{\makebox[0.9em]{%
  \sffamily\bfseries\tiny\color{white}\raisebox{0.1ex}{#1}}}}\,}
\newsavebox{\casecardbox}
\newenvironment{casecard}[3]{%
  \par\addvspace{10pt}\noindent
  \begin{lrbox}{\casecardbox}\begin{minipage}{\linewidth}
  \setlength{\parindent}{0pt}%
  {\color{caseInk}\sffamily\bfseries\normalsize #1\hspace{6pt}#2}\hfill
  {\sffamily\scriptsize\color{caseLabel}\MakeUppercase{#3}}\par
  \vspace{1pt}{\color{caseInk}\rule{\linewidth}{0.8pt}}\par\vspace{4pt}}%
  {\par\vspace{4pt}{\color{caseInk!30}\rule{\linewidth}{0.4pt}}%
  \end{minipage}\end{lrbox}%
  \typeout{CASECARD-HEIGHT: \the\dimexpr\ht\casecardbox+\dp\casecardbox\relax}%
  \usebox{\casecardbox}\par\addvspace{10pt}}

\input{layout_helpers}

\title{Reconstruction as Spatial Evaluation:\\
Diagnosing Cross-View Spatial Intelligence in Large Vision-Language Models}
\author{%
\textbf{Junfei Wu$^{1,2}$\thanks{Work done during an internship at Alibaba Token Hub, Alibaba Group.}\hspace{0.45em},
Qiuyue Wang$^{2}$, Xuhong Huang$^{2,3}$\footnotemark[1]\hspace{0.45em},
Qihang Peng$^{2,4}$\footnotemark[1]\hspace{0.45em}, Mingsheng Li$^{2}$,} \\
\textbf{Jian Guan$^{2}$, Qiang Liu$^{1}$, Shu Wu$^{1}$, Liang Wang$^{1}$, Shuai Bai$^{2}$, Tieniu Tan$^{1,5}$} \\[4pt]
{\normalfont $^{1}$NLPR, MAIS, Institute of Automation, Chinese Academy of Sciences} \\
{\normalfont $^{2}$Alibaba Token Hub, Alibaba Group} \\
{\normalfont $^{3}$The University of Hong Kong} \\
{\normalfont $^{4}$The Chinese University of Hong Kong} \\
{\normalfont $^{5}$Nanjing University}
}

\date{}

\begin{document}
\maketitle
\ifvenuestyle
  \lhead{ReconSpatial: Reconstruction as Spatial Evaluation}
  \thispagestyle{fancy}
\fi
\begin{abstract}

The spatial intelligence of large vision-language models (LVLMs) requires the
ability to combine local visual observations into a geometrically consistent
representation of the physical world. Many static benchmarks assess isolated
spatial judgments, such as relative direction or distance, while interactive embodied
evaluations combine perception with exploration, planning, control, and
simulator-specific execution.
To bridge this gap, we introduce \textbf{\bench{}}\footnote{Project repository: \url{https://github.com/Hyperwjf/ReconSpatial}.}, a benchmark centered on multi-view 3D
scene reconstruction across indoor, tabletop, and outdoor environments. Given real
multi-view images and target-object queries, LVLMs are required to jointly predict
oriented 3D bounding boxes in a shared, camera-anchored metric coordinate frame.
Reconstructions are evaluated along two complementary axes: \emph{Structure} for relative
layout and \emph{Metric} for physical scale and camera-anchored placement.
Nine atomic spatial QA tasks, grouped into \emph{Geometric Perception} and \emph{Spatial
Relations}, probe spatial capabilities involved in joint reconstruction. Offline navigation
extends evaluation to path planning.
Evaluation of 18 frontier LVLMs reveals a consistent gap between relative scene structure
and metric reconstruction; even GPT-6-Astra scores 81.4\% on Structure but only 58.7\% on
Metric. Tracing this gap, diagnostic analyses link atomic spatial capabilities to
reconstruction quality and show that relational consistency alone does not ensure
metric accuracy. Experiments with 3D map injection further separate scene construction
from path planning, revealing how these geometric errors limit navigation success.
Finally, spatial fine-tuning yields
broad transfer across both spatial and general vision-language benchmarks.

\end{abstract}

\section{Introduction}
\label{sec:introduction}

\begin{figure}[t]
  \centering
  \includegraphics[width=\linewidth]{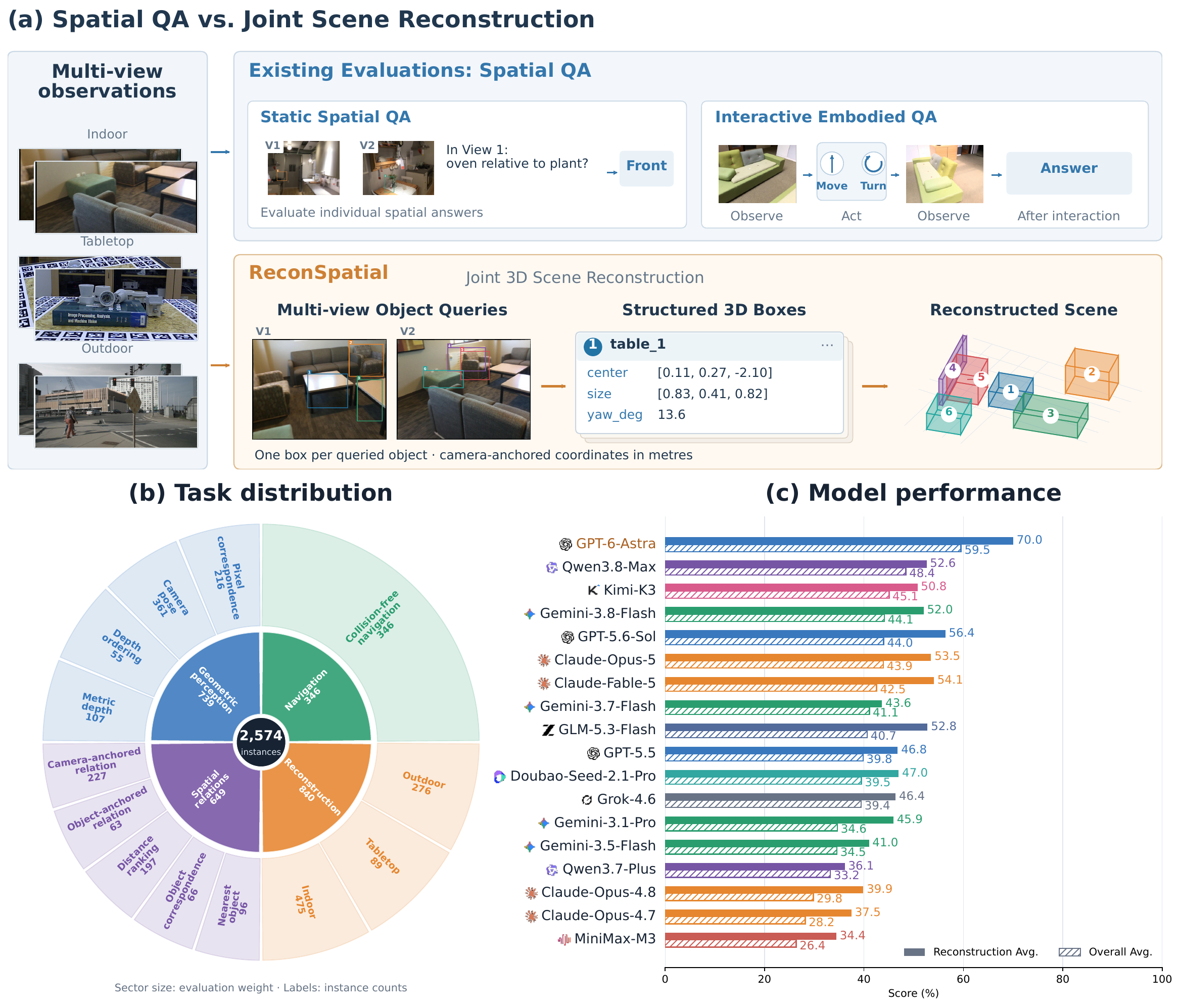}
  \caption{From spatial QA to joint 3D box prediction for evaluating LVLM spatial reasoning.
  \textbf{(a)} Static and interactive embodied QA compared with reconstruction
  from multi-view object queries via structured 3D boxes, visualized as a joint
  scene; reference geometry illustrates the target output.
  \textbf{(b)} Task distribution across four evaluation levels.
  \textbf{(c)} Eighteen models ranked by Overall, alongside their
  reconstruction score.}
  \label{fig:overview}
\end{figure}

A central requirement of spatial reasoning in large vision-language models
(LVLMs) is the ability to combine partial, viewpoint-dependent observations
into a coherent understanding of three-dimensional space. Although contemporary
LVLMs have made substantial progress in visual-semantic understanding
~\citep{bai2025qwen3,comanici2025gemini,achiam2023gpt,fu2025video,fu2026mme},
their spatial intelligence remains severely limited, 
particularly in understanding complex 3D environments~\citep{yang2025thinking,yang2026mmsi,liu2025ir3d}.

Contemporary spatial benchmarks primarily fall into two main paradigms. 
Static benchmarks assess individual spatial judgments, such as relative direction
and distance~\citep{yang2025thinking, yang2026mmsi, wang2025site,jia2026omnispatial,li2025viewspatial}.
However, independent queries fail to test whether a model can jointly
recover a coherent scene: for instance, knowing which object is closer does not determine whether their 3D positions and extents form a
valid layout.
Conversely, embodied benchmarks evaluate spatial understanding 
through interaction in simulated environments~\citep{majumdar2024openeqa,hong2026esi}, where
outcomes jointly depend on perception, exploration, planning, control, and
simulator-specific execution.
This leaves an unmet need: evaluating whether LVLMs can reason jointly about
object geometry across views, independently of interactive execution.

To bridge this evaluation gap, we introduce \bench{}, a benchmark for
evaluating cross-view spatial reasoning in LVLMs through joint prediction
of object-level 3D bounding boxes.
Given multi-view observations and target object queries, an LVLM is tasked
with predicting oriented 3D bounding boxes in a shared, camera-anchored metric coordinate
frame. This joint objective requires models to associate objects across viewpoints, resolve
relative camera geometry, and estimate metric extents simultaneously. Providing queried 2D
regions controls for open-set detection, isolating spatial reasoning as the primary point of
evaluation. We assess each reconstruction along two complementary axes: \emph{Structure},
which evaluates relative layout and topological arrangement, and \emph{Metric},
which measures physical scale and camera-anchored 3D placement across indoor, tabletop, and
outdoor scenes.

\bench{} organizes evaluation into four levels: \emph{Geometric Perception},
\emph{Spatial Relations}, \emph{Reconstruction}, and \emph{Navigation}
(Figure~\ref{fig:overview}). The first two levels comprise nine spatial QA tasks
that probe individual spatial capabilities, including cross-view correspondence,
camera pose estimation, and relative distance judgments. Their scores help us
examine how these capabilities relate to joint reconstruction performance.
\emph{Navigation} independently evaluates path planning: given
multi-view observations, a starting viewpoint, and a visual target, the model predicts
a path that reaches the target while avoiding obstacles.

Evaluation of 18 frontier LVLMs reveals a consistent gap between 
relative scene structure and metric reconstruction; even the 
top-performing GPT-6-Astra scores 81.4\% on Structure but only 58.7\% 
on Metric. Diagnostic analyses connect individual spatial capabilities to 
reconstruction quality, showing that relational consistency alone does 
not ensure metric accuracy. Experiments with 3D map injection show that replacing
self-generated maps with ground-truth geometry boosts navigation success
by 55.2 to 69.7 percentage points.
This gap indicates that downstream navigation is primarily bounded
by perceptual accuracy rather than planning capability.
Finally, spatial fine-tuning confers broad performance gains across 
both spatial and general vision-language benchmarks.

The contributions of this work are summarized as follows:
\begin{itemize}
  \item \textbf{Joint 3D reconstruction for spatial evaluation.}
  We introduce \bench{}, which establishes a new evaluation paradigm that benchmarks LVLMs on 
  jointly recovering object-level 3D bounding boxes from real multi-view observations.
  Complementary \emph{Structure} and \emph{Metric} scores distinguish relative scene
  coherence from physical scale and placement.
  \item \textbf{A hierarchical diagnostic suite across diverse domains.} 
  Covering indoor, tabletop, and outdoor scenes, \bench{} unifies \numatomic{} atomic items, 
  \numrecon{} reconstruction instances, and \numnavigation{} offline navigation episodes 
  to evaluate local spatial capabilities, joint scene reconstruction, and navigation.
   \item \textbf{Diagnostic insights and transferable supervision.}
  Evaluating eighteen frontier models uncovers a pervasive Structure--Metric
  disparity. Controlled interventions show that object-level metric depth
  yields larger Metric gains than relational cues. Downstream map substitution decouples spatial perception from
  planning competence, and spatial fine-tuning confirms transferable reasoning
  benefits.
\end{itemize}

\section{Related Work}
\label{sec:related_work}

\paragraph{Benchmarking Spatial Reasoning.}
Spatial reasoning benchmarks have primarily targeted local geometric 
queries~\citep{yang2025thinking,ma20253dsrbench,wang2025site,jia2026omnispatial,li2025viewspatial,yang2026mmsi} or 
embodied task execution~\citep{majumdar2024openeqa,hong2026esi,zhao2026espire}.
Recent efforts also explore 3D room generation~\citep{yang2026codeasroom},
inverse rendering of synthetic single-image scenes in
IR3D-Bench~\citep{liu2025ir3d}, top-down metric layouts in
Video2Layout~\citep{huang2025video2layout}, and 2D floor plans from apartment
photographs in Blueprint-Bench~\citep{petersson2025blueprintbench}.
In contrast, \bench{} evaluates joint object geometry from real-world multi-view observations,
directly scoring predicted 3D bounding boxes against annotated ground truth.
Complementary \emph{Structure} and \emph{Metric} scores explicitly disentangle relative scene layout from physical scale and camera-anchored placement.

\paragraph{Spatial Representations for Reasoning and Planning.}
Explicit spatial representations provide essential intermediate abstractions between perception
and action. Recent agentic approaches support spatial reasoning
with intermediate information obtained through code execution, tool calls, or visual sketching~\citep{marsili2025visualagenticaispatial, wu2025vilasr,chen2026spacetools,dai2026sagentspatialtooluseelicits,cho2026spatialclaw}, while cognitive-mapping approaches study structured
representations that scaffold mental navigation~\citep{yang2025thinking,zhu2026mind_over_space}. Complementary work examines data
and training strategies for spatial reasoning~\citep{ouyang2025spacer,zhao2026ouroborosspatial,chen2024spatialvlmendowingvisionlanguagemodels,li2026spatialevoselfevolvingspatialintelligence,li2025spatialladderprogressivetrainingspatial}.
We build upon this insight by treating reconstructed 3D bounding box maps as an explicit spatial 
reasoning representation, directly probing how their geometric fidelity supports path planning.

\section{Reconstruction as Spatial Evaluation}
\label{sec:benchmark}

\begin{figure}[t]
  \centering
  \includegraphics[width=\linewidth]{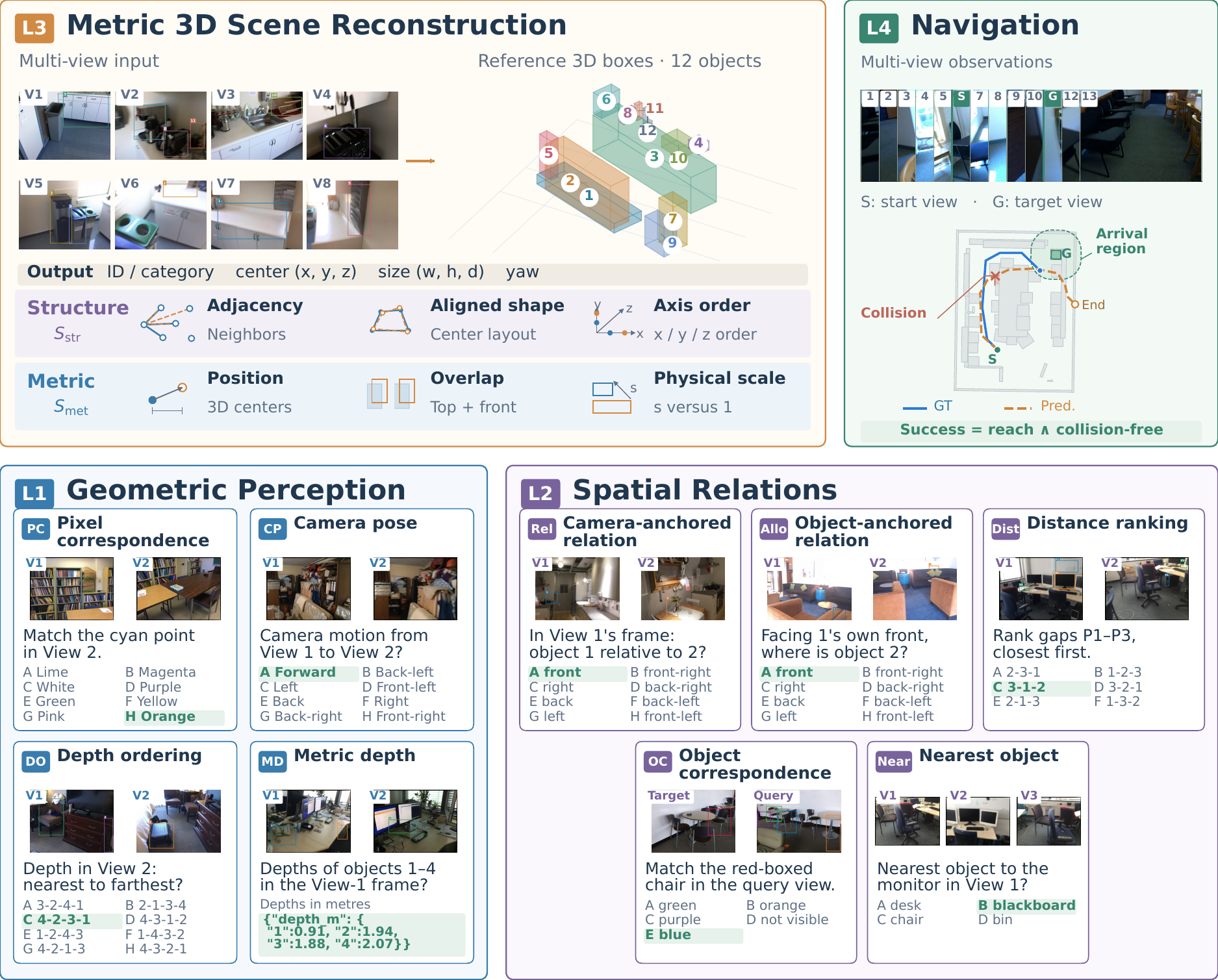}
  \caption{Examples across the four levels of \bench{}, with reconstruction
  scoring schematics and navigation annotations on the shown scene.
  Bold green options mark reference answers.
  Prompts are abbreviated; complete examples appear in
  Appendix~\ref{app:task_examples}.}
  \label{fig:task_suite}
\end{figure}

\bench{} evaluates cross-view spatial reasoning in LVLMs by requiring them
to jointly predict oriented 3D bounding boxes from multi-view images and
object queries.
Spatial QA tasks provide complementary measurements of individual spatial capabilities,
while an independent navigation task evaluates path planning
(Figure~\ref{fig:task_suite}).

\subsection{Problem formulation}

Let $\mathcal{V}=(I_1,\ldots,I_N)$ denote an ordered sequence of multi-view
observations, with $I_1$ serving as the reference view. Let
$\mathcal{Q}=\{q_k\}_{k=1}^{K}$ denote the target-object queries, where
\[
q_k=(u_k,j_k,\mathbf{b}_k,\ell_k)
\]
specifies an object identifier $u_k$, a designated view $I_{j_k}$, its 2D
bounding box $\mathbf{b}_k$ in that view, and its category label $\ell_k$.
Given $\mathcal{V}$ and $\mathcal{Q}$, the evaluated LVLM $F$ is required
to output one oriented 3D bounding box for each queried object:
\begin{equation}
  \widehat{\mathcal{S}}=F(\mathcal{V},\mathcal{Q})
  =\{(u_k,\hat{\mathbf{c}}_k,\hat{\mathbf{s}}_k,
  \hat{\psi}_k)\}_{k=1}^{K},
  \label{eq:scene}
\end{equation}
where $\hat{\mathbf{c}}_k\in\mathbb{R}^3$ and
$\hat{\mathbf{s}}_k\in\mathbb{R}_{+}^3$ denote the center and side lengths of
the predicted 3D box, and $\hat{\psi}_k$ denotes its yaw about the gravity axis.
All quantities are expressed in metres in a gravity-aligned frame whose
horizontal orientation is anchored to $I_1$. Its origin is the vertical
projection of the reference-camera center onto the support surface.

Providing object identifiers, category labels, and image regions controls for
open-set detection while preserving the need to associate targets across views
and predict a mutually compatible scene. The LVLM reports object IDs,
centers, dimensions, and yaw angles as structured numerical outputs, which
are parsed and scored against reference annotations. Oriented 3D boxes make the inferred
geometry directly measurable without assuming that the model internally
represents the scene in the same form. Predictions are matched to queried
objects by identifier under the fixed parsing and missing-output protocol
detailed in Appendix~\ref{app:scoring}.
\subsection{Complementary Dimensions of Spatial Fidelity}
\label{sec:fidelity_metrics}

A reconstruction may preserve relative scene organization while remaining
inaccurate in physical scale and camera-anchored placement. We therefore
evaluate \emph{Structure} and \emph{Metric} separately:
\begin{align}
  S_{\mathrm{ord}} &= \tfrac13(S_{\mathrm{lr}}+S_{\mathrm{fb}}+S_{\mathrm{up}}),\\
  S_{\mathrm{str}} &= \tfrac13(S_{\mathrm{adj}}+S_{\mathrm{shape}}+S_{\mathrm{ord}}),\\
  S_{\mathrm{met}} &= \tfrac13(S_{\mathrm{pos}}+S_{\mathrm{viewIoU}}+S_{\mathrm{scale}}).
\end{align}
All component scores lie in $[0,1]$, with higher values indicating better
agreement. Table~\ref{tab:score_anatomy} summarizes what each component tests;
Figure~\ref{fig:task_suite} illustrates the six quantities.

\begin{table}[t]
  \centering
  \small
  \caption{Anatomy of the reconstruction scores. Metric separates scale-corrected
  geometry from the size of the required scale correction.}
  \label{tab:score_anatomy}
  \begin{tabular}{lll}
    \toprule
    Score & Component & Quantity tested \\
    \midrule
    Structure & Adjacency & Jaccard agreement of local neighbor graphs \\
              & Shape & Residual after planar similarity alignment \\
              & Axis order & Chance-corrected Kendall order on three axes \\
    \midrule
    Metric & Position & AP over normalized localization thresholds \\
           & View IoU & Mean top-down and front-view box overlap \\
           & Scale & Penalty for the fitted global scale correction \\
    \bottomrule
  \end{tabular}
\end{table}

\paragraph{Relative layout and directional consistency.}
Let $g_i,p_i\in\mathbb{R}^2$ be the horizontal $(x,z)$ centers of the $n$
matched ground-truth and predicted objects. For their undirected
$k$-nearest-neighbor edge sets $E_G,E_P$, with $k=\min(3,n-1)$, define
\begin{align}
  r_{\mathrm{adj}} &= \frac{|E_G\cap E_P|}{|E_G\cup E_P|},\\
  r_{\mathrm{shape}} &= \max\left\{0,1-
  \frac{\sum_{i=1}^{n}\|g_i-\phi(p_i)\|_2}{n(D/2+\epsilon)}\right\}.
\end{align}
Here $D$ is the diagonal of the ground-truth centers' axis-aligned bounding
box, and $\phi(p)=aRp+t$, with $R\in\mathrm{SO}(2)$, is a least-squares
planar similarity fit using the fixed object-ID correspondences. Adjacency
and shape are invariant to planar translation, rotation, and uniform scale.
Directional ordering retains constraints in the camera-anchored frame that
this fit removes. On each of the left--right, front--back, and vertical axes,
we use Kendall's tie-adjusted correlation:
\begin{equation}
  r_{\mathrm{axis}}=\tau_b=
  \frac{N_c-N_d}{\sqrt{(N_c+N_d+T_G)(N_c+N_d+T_P)}}.
\end{equation}
$N_c,N_d$ count concordant and discordant object pairs, and $T_G,T_P$ count
pairs tied only in the ground-truth or predicted ordering. Constant or
undefined orderings receive raw score zero.

\paragraph{Agreement beyond a random layout.}
Each raw Structure similarity $r_j$ is corrected against an empirical
random-layout reference $b_j$:
\begin{equation}
  S_j=\operatorname{clip}_{[0,1]}
  \left(\frac{r_j-b_j}{1-b_j+\epsilon}\right),
  \qquad \epsilon=1/10^9.
  \label{eq:structure_calibration}
\end{equation}
The reference averages 20 draws of $n$ horizontal points within the
ground-truth bounding box, or $n$ heights within its vertical range.
Thus, Structure measures agreement beyond chance rather than rewarding
relations that arise in arbitrary layouts. Uninformative channels with
$r_j=b_j=1$, such as a necessarily complete neighbor graph, receive zero.
Per-instance seeds and numerical details are specified in
Appendix~\ref{app:scoring}.

\paragraph{Camera-anchored placement and physical scale.}
Let $G_i,P_i\in\mathbb{R}^3$ denote matched 3D centers. Metric uses a single
scale-only least-squares fit,
\begin{equation}
  s=\frac{\sum_i\langle G_i,P_i\rangle}{\sum_i\|P_i\|_2^2+\epsilon},
  \label{eq:metric_scale_fit}
\end{equation}
with no fitted rotation or translation. Position accuracy averages the
fraction of matched objects localized within four thresholds:
\begin{equation}
  S_{\mathrm{pos}}=\frac{1}{4n}
  \sum_{t\in\{0.1,0.2,0.4,0.8\}}\sum_{i=1}^{n}
  \mathbf{1}\!\left[\|G_i-sP_i\|_2<tR_g\right],
  \label{eq:position_score}
\end{equation}
where $R_g=\sqrt{\frac1n\sum_i\|G_i-\bar G\|_2^2}+\epsilon$ is the
ground-truth radius of gyration. The same fitted scale is applied to
predicted centers and box extents for View IoU, which averages per-object
polygon overlap in the ground-plane $(x,z)$ and front-plane $(x,y)$
projections. It jointly reflects errors in position, extent, and orientation.
The third component penalizes the scale correction itself:
\begin{equation}
  S_{\mathrm{scale}}=
  \max\left\{0,1-\frac{|\log\max(s,1/10^6)|}{\log 2}\right\}.
  \label{eq:scale_score}
\end{equation}
Metric combines scale-corrected placement and overlap with an explicit scale
penalty. Absolute
position, size, and yaw errors are retained as additional diagnostics.

\subsection{Diagnostic Hierarchy and Evaluation Domains}
  \label{sec:diagnostic_hierarchy}

\paragraph{Task design and diagnostic hierarchy.}
The benchmark tasks are designed to distribute relevant spatial evidence
across multiple views, requiring LVLMs to compare observations and reason
about geometry in shared reference frames. Because an aggregate
reconstruction score cannot isolate constituent failure modes, \bench{}
complements holistic reconstruction with nine atomic spatial QA diagnostics
organized into two task families (Figure~\ref{fig:task_suite}).
\emph{Geometric Perception} examines cross-view correspondence, camera pose,
depth ordering, and metric depth. \emph{Spatial Relations} assesses camera-
and object-anchored directions, distance ranking, object correspondence,
and proximity. In total, \bench{} includes \numrecon{} joint reconstruction
instances, \numatomic{} atomic spatial QA items, and \numnavigation{}
independent offline navigation episodes (Figure~\ref{fig:overview}(b)).
Counts refer to task instances rather than distinct physical scenes.
Detailed task formulations appear in Appendix~\ref{app:tasks}.

\paragraph{Evaluation domains.}
Reconstruction draws on ScanNet and CA-1M for indoor scenes~\citep{dai2017scannet,lazarow2025cubify};
YCB-Video, T-LESS, HomebrewedDB, and NOCS for tabletop scenes~\citep{xiang2018posecnn,hodan2017tless,kaskman2019homebreweddb,wang2019nocs};
and nuScenes for outdoor scenes~\citep{caesar2020nuscenes}. Reconstruction scores
are first averaged within each domain and then equally across the three
domain means, preventing larger domains from dominating the aggregate score.
Data composition and construction details are given in Appendix~\ref{app:tasks}
and Table~\ref{tab:composition}.

\subsection{Offline Navigation and Success Criteria}
\label{sec:navigation_metrics}

Navigation evaluates path planning in LVLMs independently
of the reconstruction task. Each single-room episode provides multi-view
observations, a starting viewpoint, and a visual target.
The LVLM is required to predict
an egocentric path $\widehat{\mathcal P}=\{\widehat{\mathbf p}_t\}_{t=1}^{T}$
on the ground plane relative to the starting viewpoint.

\paragraph{Arrival and collision avoidance.}
For the evaluated path $\mathcal Q=\{\mathbf q_t\}_{t=1}^{T}$ and target
footprint $\mathcal G$, success is
\begin{equation}
  \operatorname{Success}(\mathcal Q)=
  \mathbf{1}\!\left[d(\mathbf q_T,\mathcal G)\leq 1\,\mathrm{m}\right]
  \mathbf{1}\!\left[C_{\mathrm{coll}}(\mathcal Q)=1\right].
  \label{eq:navigation_success}
\end{equation}
Here $C_{\mathrm{coll}}$ is an avoidability-aware collision check using
annotated ground-plane obstacles. It tolerates small lateral errors when
free floor is reachable within 0.2 metres perpendicular to travel, but
rejects unavoidable obstacle crossings and all room-boundary crossings.
The target object is excluded from collision checking. The reported score
is the fraction of episodes satisfying both arrival and collision
conditions; any valid path is accepted, and unparseable paths receive zero.
Coordinate conventions and collision-checking details are provided in
Appendix~\ref{app:navigation_scoring}.

\paragraph{Scale-normalized evaluation.}
The navigation leaderboard evaluates scale-normalized paths. For a
nonzero predicted endpoint, a scalar aligns its distance from the origin
with that of the reference endpoint:
\begin{equation}
  s_{\mathrm{nav}}=
  \frac{\|\mathbf p_T^{*}\|_2}{\|\widehat{\mathbf p}_T\|_2},
  \qquad \mathbf q_t=s_{\mathrm{nav}}\widehat{\mathbf p}_t.
  \label{eq:navigation_scale}
\end{equation}
Here $\mathbf p_T^{*}$ denotes the reference path endpoint. Arrival and
collision are assessed on this transformed path. The
map-injection experiment in Section~\ref{sec:map_injection} instead
evaluates paths without scale normalization, with protocol details in
Appendix~\ref{app:map_injection}.

\section{Experiments and Analysis}
\label{sec:evaluation}

In this section, we evaluate \nummodels{} frontier LVLMs across the four
levels of \bench{}. We examine the overall benchmark landscape, the disparity
between Structure and Metric fidelity, and the gap between
atomic capabilities and joint scene reconstruction.
We further investigate these findings through information injection,
verification probes, and qualitative examples, and examine whether
cross-view spatial supervision transfers to broader multimodal tasks.

\subsection{Evaluation Setup}

The cohort comprises GPT-6-Astra \citep{openai2026astra}, GPT-5.6-Sol, and GPT-5.5;
Gemini-3.1-Pro \citep{comanici2025gemini},
Gemini-3.5-Flash \citep{google2026gemini35flash}, Gemini-3.7-Flash, and Gemini-3.8-Flash;
Claude-Opus-4.7, Claude-Opus-4.8, Claude-Opus-5, and Claude-Fable-5;
Qwen3.7-Plus \citep{bai2025qwen3} and Qwen3.8-Max \citep{qwen2026qwen38max};
Kimi-K3 \citep{team2026kimi};
Doubao-Seed-2.1-Pro \citep{bytedance2026seed21};
MiniMax-M3 \citep{minimax2026m3}; Grok-4.6; and GLM-5.3-Flash~\citep{zai2026glm53flash}.
All models use the same task definitions, parser, and scoring protocol.
Output-format and reproducibility details are given in
Appendix~\ref{app:scoring}.
The Overall score equally weights the four capability levels—Geometric Perception, Spatial Relations, Reconstruction, and Navigation—and Reconstruction equally averages Structure and Metric across the three physical domains, preventing split sizes from biasing aggregate results.

\begin{table*}[t]
  \centering
  \small
  \ifdefined\XeTeXversion
    \renewcommand{\bfdefault}{bx}
    \fontencoding{OT1}\fontfamily{cmr}\fontsize{10}{12}\selectfont
  \fi
  \setlength{\tabcolsep}{1.8pt}
  \setlength{\fboxsep}{1.2pt}
  \renewcommand{\arraystretch}{1.12}
  \definecolor{benchGeo}{HTML}{365E82}
  \definecolor{benchRel}{HTML}{73558F}
  \definecolor{benchRecon}{HTML}{A15B26}
  \definecolor{benchNav}{HTML}{367262}
  \definecolor{benchOverall}{HTML}{384959}
  \newcommand{\benchgroup}[2]{%
    \colorbox{#1!15}{\strut\textcolor{#1}{\textbf{#2}}}}
  \newcommand{\benchavg}[2]{%
    \colorbox{#1!8}{\makebox[2.1em][c]{\strut\textcolor{#1!65!black}{#2}}}}
  \caption{Main results for eighteen frontier LVLMs. Atomic-task abbreviations
  are defined in Figure~\ref{fig:task_suite}; Reconstruction reports three-domain
  Structure (S), Metric (M), and their mean. Scores are percentages;
  Navigation uses scale-normalized paths and Overall equally averages the four
  levels. Best and second-best values are bold and underlined, respectively.}
  \label{tab:main_results}
  \newcommand{\generatedmainresultrows}{%
    \raisebox{-0.15em}{\includegraphics[width=1em,height=1em,keepaspectratio]{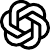}}\hspace{0.25em}\textbf{GPT-6-Astra}
      & \textbf{87.9} & \textbf{70.8} & \textbf{89.1} & \textbf{53.6} & \benchavg{benchGeo}{\textbf{75.3}} & \textbf{53.3} & \underline{50.8} & \underline{61.4} & \underline{72.7} & \textbf{74.0} & \benchavg{benchRel}{\textbf{62.4}} & \textbf{81.4} & \textbf{58.7} & \benchavg{benchRecon}{\textbf{70.0}} & \benchavg{benchNav}{\textbf{30.3}} & \benchavg{benchOverall}{\textbf{59.5}} \\
    \raisebox{-0.15em}{\includegraphics[width=1em,height=1em,keepaspectratio]{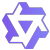}}\hspace{0.25em}Qwen3.8-Max
      & \underline{74.0} & \underline{56.2} & \underline{83.3} & 47.1 & \benchavg{benchGeo}{\underline{65.2}} & \underline{50.2} & \textbf{60.3} & 54.3 & 67.7 & 61.5 & \benchavg{benchRel}{\underline{58.8}} & 68.5 & 36.7 & \benchavg{benchRecon}{52.6} & \benchavg{benchNav}{17.1} & \benchavg{benchOverall}{\underline{48.4}} \\
    \raisebox{-0.15em}{\includegraphics[width=1em,height=1em,keepaspectratio]{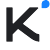}}\hspace{0.25em}Kimi-K3
      & 63.3 & 50.6 & 72.7 & 48.0 & \benchavg{benchGeo}{58.6} & 36.1 & 42.9 & 49.2 & 65.2 & 52.1 & \benchavg{benchRel}{49.1} & 68.9 & 32.7 & \benchavg{benchRecon}{50.8} & \benchavg{benchNav}{\underline{21.7}} & \benchavg{benchOverall}{45.1} \\
    \raisebox{-0.15em}{\includegraphics[width=1em,height=1em,keepaspectratio]{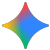}}\hspace{0.25em}Gemini-3.8-Flash
      & 64.2 & 54.1 & 70.9 & \underline{49.5} & \benchavg{benchGeo}{59.7} & 41.9 & 23.8 & \textbf{65.0} & 66.7 & \underline{65.6} & \benchavg{benchRel}{52.6} & 66.4 & 37.6 & \benchavg{benchRecon}{52.0} & \benchavg{benchNav}{12.1} & \benchavg{benchOverall}{44.1} \\
    \raisebox{-0.15em}{\includegraphics[width=1em,height=1em,keepaspectratio]{figures/model_logos/openai.pdf}}\hspace{0.25em}GPT-5.6-Sol
      & 56.7 & 50.9 & 72.7 & 41.8 & \benchavg{benchGeo}{55.5} & 42.5 & 34.9 & 34.0 & 53.0 & 46.9 & \benchavg{benchRel}{42.3} & \underline{71.3} & \underline{41.5} & \benchavg{benchRecon}{\underline{56.4}} & \benchavg{benchNav}{\underline{21.7}} & \benchavg{benchOverall}{44.0} \\
    \raisebox{-0.15em}{\includegraphics[width=1em,height=1em,keepaspectratio]{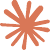}}\hspace{0.25em}Claude-Opus-5
      & 58.1 & 51.9 & 74.6 & 29.6 & \benchavg{benchGeo}{53.6} & 48.0 & 47.6 & 54.3 & 47.0 & 59.4 & \benchavg{benchRel}{51.3} & 68.1 & 38.9 & \benchavg{benchRecon}{53.5} & \benchavg{benchNav}{17.3} & \benchavg{benchOverall}{43.9} \\
    \raisebox{-0.15em}{\includegraphics[width=1em,height=1em,keepaspectratio]{figures/model_logos/claude.pdf}}\hspace{0.25em}Claude-Fable-5
      & 55.4 & 46.4 & 69.1 & 38.8 & \benchavg{benchGeo}{52.4} & 39.2 & 44.4 & 45.7 & 42.4 & \underline{65.6} & \benchavg{benchRel}{47.5} & 68.7 & 39.4 & \benchavg{benchRecon}{54.1} & \benchavg{benchNav}{16.2} & \benchavg{benchOverall}{42.5} \\
    \raisebox{-0.15em}{\includegraphics[width=1em,height=1em,keepaspectratio]{figures/model_logos/gemini.pdf}}\hspace{0.25em}Gemini-3.7-Flash
      & 63.7 & 52.7 & 58.2 & 46.6 & \benchavg{benchGeo}{55.3} & 42.3 & 25.4 & 53.3 & 69.7 & 55.2 & \benchavg{benchRel}{49.2} & 53.7 & 33.6 & \benchavg{benchRecon}{43.6} & \benchavg{benchNav}{16.2} & \benchavg{benchOverall}{41.1} \\
    \raisebox{-0.15em}{\includegraphics[width=1em,height=1em,keepaspectratio]{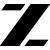}}\hspace{0.25em}GLM-5.3-Flash
      & 44.6 & 45.9 & 74.6 & 49.2 & \benchavg{benchGeo}{53.6} & 31.7 & 23.8 & 50.8 & 47.0 & 50.0 & \benchavg{benchRel}{40.7} & 68.3 & 37.2 & \benchavg{benchRecon}{52.8} & \benchavg{benchNav}{15.6} & \benchavg{benchOverall}{40.7} \\
    \raisebox{-0.15em}{\includegraphics[width=1em,height=1em,keepaspectratio]{figures/model_logos/openai.pdf}}\hspace{0.25em}GPT-5.5
      & 40.9 & 45.8 & 74.6 & 46.1 & \benchavg{benchGeo}{51.9} & 34.4 & 41.3 & 40.6 & 51.5 & 45.8 & \benchavg{benchRel}{42.7} & 65.0 & 28.6 & \benchavg{benchRecon}{46.8} & \benchavg{benchNav}{17.9} & \benchavg{benchOverall}{39.8} \\
    \raisebox{-0.15em}{\includegraphics[width=1em,height=1em,keepaspectratio]{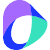}}\hspace{0.25em}Doubao-Seed-2.1-Pro
      & 47.9 & 46.0 & 63.6 & 45.3 & \benchavg{benchGeo}{50.7} & 33.0 & 7.9 & 44.2 & \textbf{74.2} & 56.2 & \benchavg{benchRel}{43.1} & 63.8 & 30.2 & \benchavg{benchRecon}{47.0} & \benchavg{benchNav}{17.1} & \benchavg{benchOverall}{39.5} \\
    \raisebox{-0.15em}{\includegraphics[width=1em,height=1em,keepaspectratio]{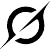}}\hspace{0.25em}Grok-4.6
      & 32.1 & 45.9 & 80.0 & 43.6 & \benchavg{benchGeo}{50.4} & 33.9 & 41.3 & 42.1 & 59.1 & 50.0 & \benchavg{benchRel}{45.3} & 63.3 & 29.4 & \benchavg{benchRecon}{46.4} & \benchavg{benchNav}{15.6} & \benchavg{benchOverall}{39.4} \\
    \raisebox{-0.15em}{\includegraphics[width=1em,height=1em,keepaspectratio]{figures/model_logos/gemini.pdf}}\hspace{0.25em}Gemini-3.1-Pro
      & 44.6 & 43.6 & 60.0 & 36.3 & \benchavg{benchGeo}{46.1} & 28.2 & 38.1 & 26.9 & 43.9 & 36.5 & \benchavg{benchRel}{34.7} & 62.3 & 29.6 & \benchavg{benchRecon}{45.9} & \benchavg{benchNav}{11.8} & \benchavg{benchOverall}{34.6} \\
    \raisebox{-0.15em}{\includegraphics[width=1em,height=1em,keepaspectratio]{figures/model_logos/gemini.pdf}}\hspace{0.25em}Gemini-3.5-Flash
      & 48.4 & 46.9 & 43.6 & 36.9 & \benchavg{benchGeo}{44.0} & 33.9 & 31.8 & 37.1 & 63.6 & 49.0 & \benchavg{benchRel}{43.1} & 49.1 & 32.9 & \benchavg{benchRecon}{41.0} & \benchavg{benchNav}{10.1} & \benchavg{benchOverall}{34.5} \\
    \raisebox{-0.15em}{\includegraphics[width=1em,height=1em,keepaspectratio]{figures/model_logos/qwen.pdf}}\hspace{0.25em}Qwen3.7-Plus
      & 37.7 & 41.3 & 65.5 & 33.4 & \benchavg{benchGeo}{44.5} & 30.4 & 46.0 & 32.5 & 53.0 & 33.3 & \benchavg{benchRel}{39.1} & 52.7 & 19.6 & \benchavg{benchRecon}{36.1} & \benchavg{benchNav}{13.3} & \benchavg{benchOverall}{33.2} \\
    \raisebox{-0.15em}{\includegraphics[width=1em,height=1em,keepaspectratio]{figures/model_logos/claude.pdf}}\hspace{0.25em}Claude-Opus-4.8
      & 29.8 & 38.0 & 56.4 & 31.3 & \benchavg{benchGeo}{38.9} & 26.1 & 25.4 & 37.6 & 28.8 & 35.4 & \benchavg{benchRel}{30.7} & 55.5 & 24.3 & \benchavg{benchRecon}{39.9} & \benchavg{benchNav}{9.8} & \benchavg{benchOverall}{29.8} \\
    \raisebox{-0.15em}{\includegraphics[width=1em,height=1em,keepaspectratio]{figures/model_logos/claude.pdf}}\hspace{0.25em}Claude-Opus-4.7
      & 29.3 & 36.6 & 40.0 & 34.1 & \benchavg{benchGeo}{35.0} & 29.2 & 23.8 & 31.0 & 40.9 & 26.0 & \benchavg{benchRel}{30.2} & 53.9 & 21.1 & \benchavg{benchRecon}{37.5} & \benchavg{benchNav}{10.1} & \benchavg{benchOverall}{28.2} \\
    \raisebox{-0.15em}{\includegraphics[width=1em,height=1em,keepaspectratio]{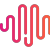}}\hspace{0.25em}MiniMax-M3
      & 30.7 & 37.5 & 50.9 & 40.5 & \benchavg{benchGeo}{39.9} & 25.6 & 11.1 & 26.4 & 18.2 & 28.1 & \benchavg{benchRel}{21.9} & 50.7 & 18.2 & \benchavg{benchRecon}{34.4} & \benchavg{benchNav}{9.2} & \benchavg{benchOverall}{26.4} \\
}

  \resizebox{\textwidth}{!}{%
  \begin{tabular}{l*{5}{c}@{\hspace{5pt}}*{6}{c}@{\hspace{5pt}}*{3}{c}@{\hspace{5pt}}cc}
    \toprule
    \textbf{Model}
      & \multicolumn{5}{c}{\benchgroup{benchGeo}{Geometric Perception}}
      & \multicolumn{6}{c}{\benchgroup{benchRel}{Spatial Relations}}
      & \multicolumn{3}{c}{\benchgroup{benchRecon}{Reconstruction}}
      & \benchgroup{benchNav}{Navigation}
      & \benchgroup{benchOverall}{Overall} \\
    \cmidrule(lr){2-6}\cmidrule(lr){7-12}\cmidrule(lr){13-15}\cmidrule(lr){16-16}\cmidrule(lr){17-17}
      & PC & CP & DO & MD & \benchavg{benchGeo}{\textbf{Avg.}}
      & Rel & Allo & Dist & OC & Near & \benchavg{benchRel}{\textbf{Avg.}}
      & S & M & \benchavg{benchRecon}{\textbf{Avg.}}
      & \benchavg{benchNav}{\textbf{Avg.}}
      & \benchavg{benchOverall}{\textbf{Avg.}} \\
    \midrule
    \generatedmainresultrows
    \bottomrule
  \end{tabular}
  }
\end{table*}

\subsection{Main Results}

Table~\ref{tab:main_results} illustrates the overall benchmark landscape. 
GPT-6-Astra sets the performance frontier, leading all four levels with 59.5 
Overall, followed by Qwen3.8-Max at 48.4. Across capability tiers, runner-up 
rankings vary: Qwen3.8-Max ranks second in both atomic levels, 
while GPT-5.6-Sol places second in reconstruction and ties with Kimi-K3 
for second in navigation. Farther down the leaderboard, Claude-Fable-5 ranks 
third in reconstruction (54.1) but drops to seventh Overall (42.5). 
Close aggregate scores thus frequently mask divergent competencies across 
local geometric reasoning, scene reconstruction, and planning. 
Across all models, navigation remains universally difficult: 
even GPT-6-Astra succeeds in only 30.3\% of episodes under scale-normalized 
evaluation, motivating our map-injection experiments to probe the origins
of planning failures in Section~\ref{sec:map_injection}.

\begin{figure}[t]
  \centering
  \includegraphics[width=\linewidth]{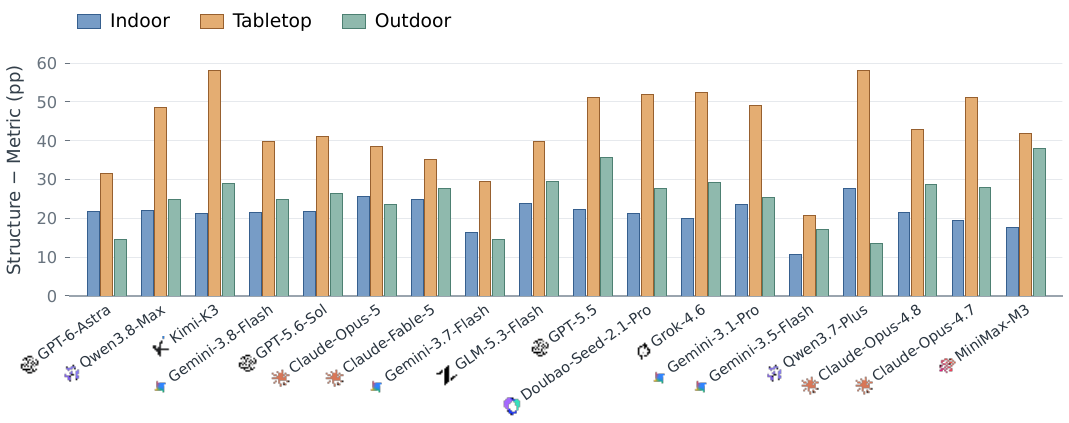}
  \caption{Structure--Metric gaps by domain across eighteen models, ordered
  as in Table~\ref{tab:main_results}. Bars show Structure minus Metric in
  percentage points. All 54 gaps are positive, with tabletop scenes showing
  the largest gap for every model.}
  \label{fig:structure_metric_gap}
\end{figure}

Table~\ref{tab:main_results} shows a persistent gap between relative layout
recovery and metric reconstruction, even for the strongest models.
Figure~\ref{fig:structure_metric_gap} shows that this disparity holds in all
54 model--domain combinations and is largest on tabletop scenes for every
model. Indoor gaps are comparatively uniform across models, while tabletop
and outdoor gaps vary more substantially. Recovering recognizable object
relations therefore does not guarantee calibrated geometry for
distance-sensitive planning. Per-domain results and scorer robustness checks
are provided in Appendices~\ref{app:domain_results}
and~\ref{app:scorer_sensitivity}.

\subsection{Diagnostic Analysis}
\label{sec:analysis}

\paragraph{Atomic capabilities and reconstruction.}
\label{sec:cross_level}

To examine task-specific associations across the eighteen LVLMs, we compute
partial correlations controlling for the mean score on the other eight
atomic tasks. Camera pose is associated with
projected box overlap ($r=0.62$), while depth ordering tracks front--back
layout consistency ($r=0.63$). Pixel correspondence is associated with
Metric ($r=0.46$) but barely with adjacency ($r=0.01$), suggesting that
detailed cross-view matching is more closely linked to metric geometry than
to local neighbor relations. These patterns distinguish the geometric
demands of joint reconstruction beyond an overall spatial score.
The full correlation matrix and stability analyses are provided in
Appendix~\ref{app:cross_level}. Building on this analysis,
we use information-injection experiments to examine
how supplied geometry affects reconstruction and navigation.

\raggedbottom

\begin{figure}[tbp]
  \centering
  \includegraphics[width=\linewidth]{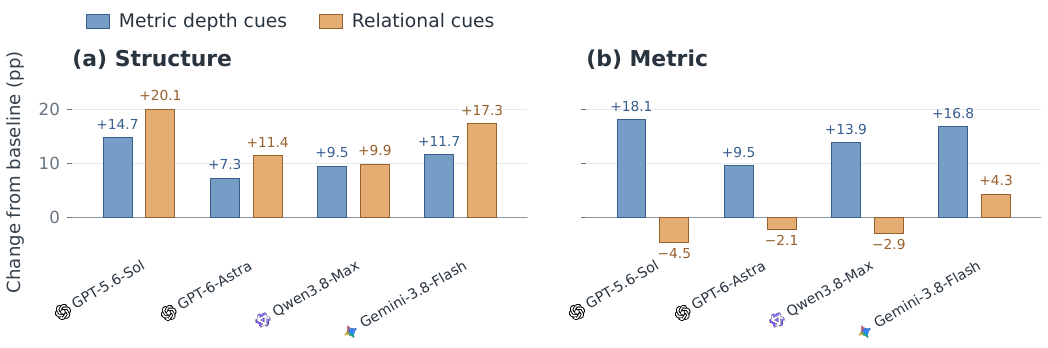}
  \caption{Geometric information injection for reconstruction. Changes are
  relative to the same prompts without added cues.}
  \label{fig:metric_injection}
\end{figure}

\paragraph{Geometric information injection.}
\label{sec:interventions}
Using the ScanNet two-view reconstruction set, we augment prompts for
GPT-5.6-Sol, GPT-6-Astra, Qwen3.8-Max, and Gemini-3.8-Flash with ground-truth
metric depth cues or relational cues. Metric depth cues specify object depths
in metres along the reference camera's viewing axis; relational cues describe
depth order, pairwise directions, distance ranks, and camera motion.
Compared with the same prompts without added cues, both cue types improve
Structure, while metric depth cues produce larger
and more consistent Metric gains across all four models
(Figure~\ref{fig:metric_injection}). For \diagmodel{}, depth alone
raises Metric by 18.1 percentage points. A further
condition supplies both true depth and camera pose, yet leaves a mean
object-position error of 0.28~m, down from 0.65~m without added cues.
Depth anchors camera-axis placement, but lateral position, size, and
orientation still require inference. Appendix~\ref{app:interventions}
details the conditions and comparisons.

\paragraph{Layout verification and revision.}
\label{sec:verification_revision}
We investigate whether visual feedback can help models assess candidate
layouts and improve their own reconstructions. In the verification task,
GPT-5.6-Sol, Claude-Opus-4.8, Gemini-3.1-Pro, and Gemini-3.5-Flash choose
between two rendered layouts: one preserves the reference geometry, while
the other perturbs object arrangement. Providing the scene photographs
raises selection accuracy to 95.0--97.5\% across the four models, showing
that they can use image evidence to judge the candidate layouts.

We then examine whether rendered feedback also helps a model produce a
better reconstruction. \diagmodel{} revisits its prediction produced without
added cues, using the original photographs and its previous scene estimate;
one condition additionally provides a rendering of that estimate. Both
revision conditions leave Metric essentially unchanged from the original
prediction, and adding the rendering provides no additional gain over
text-only revision, with a difference of $-0.3$ percentage points. Revision
requires coordinated changes to object positions, sizes, and orientations,
while verification supplies two fixed alternatives. A rendering makes the
current estimate visible without adding external geometric measurements.
High verification accuracy alone therefore does not establish that rendered
feedback is sufficient for effective metric correction.

\par\addvspace{1.5ex plus .5ex minus .2ex}
\phantomsection
\label{sec:map_injection}
\begin{rightinsettext}{table}{0.54\linewidth}{%
  \setlength{\abovecaptionskip}{0pt}
  \caption{\raggedright Navigation success (\%) with 3D box maps,
  without path rescaling.}
  \label{tab:map_intervention}
  \centering
  \ifdefined\XeTeXversion
    \renewcommand{\bfdefault}{bx}
    \fontencoding{OT1}\fontfamily{casecm}
  \fi
  \fontsize{9}{11}\selectfont
  \setlength{\tabcolsep}{1pt}
  \scalebox{0.9}{\begin{tabular}{@{}lrrrr@{}}
\toprule
Model & Images & Self-map & GT map & \shortstack{GT +\\target ID} \\
\midrule
GPT-5.6-Sol & 15.6 & 15.6 & 85.3 & \textbf{90.5} \\
GPT-6-Astra & 22.5 & 35.5 & 97.1 & \textbf{97.4} \\
Qwen3.8-Max & 9.5 & 11.0 & 66.2 & \textbf{78.6} \\
Gemini-3.8-Flash & 11.8 & 18.2 & \textbf{82.1} & 79.8 \\
\bottomrule
\end{tabular}
}
}
\textbf{Scene representations for navigation.}\quad
We assess the role of explicit scene representations in path planning by
comparing image-only planning (\emph{Images}), planning from model-generated
3D box maps (\emph{Self-map}), and planning from reference maps
(\emph{GT map}). Self-maps improve navigation success in three of the four
models; \diagmodel{} is the exception, with unchanged success
(Table~\ref{tab:map_intervention}). Thus, making a predicted scene
representation explicit already benefits navigation in most evaluated
models. Replacing these predictions with reference geometry produces a
further gain of 55.2--69.7 percentage points, highlighting scene representation
quality as a major constraint on navigation. Explicitly identifying the
target map entry (\emph{GT + target ID}) makes target arrival nearly perfect,
while the remaining failures mainly involve collisions. The residual
difficulty therefore concerns planning feasible routes through the supplied
scene, even when the target entry is given. Detailed protocols and component
results are provided in Appendix~\ref{app:map_injection}.
\end{rightinsettext}

\Needspace{14\baselineskip}
\subsection{Qualitative Analysis}
\label{sec:case_studies}

Figure~\ref{fig:case_reconstruction} compares indoor, tabletop, and outdoor
reconstructions using common metric axes within each scene, without per-model
alignment or rescaling. In the indoor case, Qwen3.8-Max preserves recognizable
object clusters (Structure: 75.7) but scores only 5.3 on Metric, with a mean
position error of 1.82~m. The tabletop case shows plausible arrangements
across all four models, while GPT-6-Astra is markedly more accurate in metric
placement. Outdoors, models preserve the barrier ordering but differ in
absolute placement; Qwen3.8-Max has a mean position error of 9.02~m. These
examples show that recognizable scene organization can coexist with large
camera-relative errors across different environments. Larger views and
additional cases are provided in Appendix~\ref{app:case_studies}.

\par\noindent\begin{minipage}{\linewidth}
  \centering
  \includegraphics[width=\linewidth]{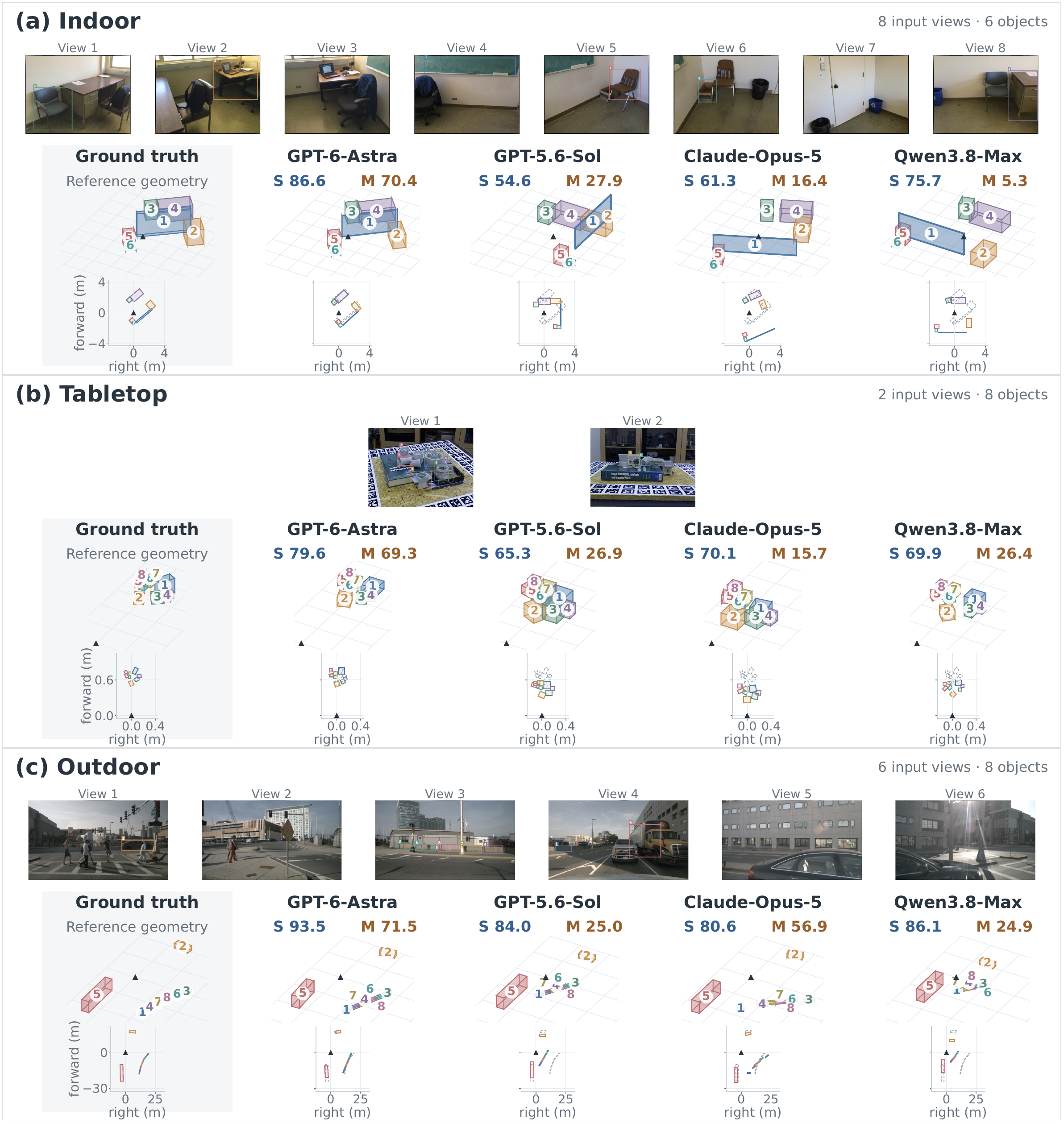}
  \captionof{figure}{Reconstruction examples across three domains. Columns
  compare reference geometry and four models; each scene uses shared metric
  axes. All 8 indoor, 2 tabletop, and 6 outdoor input views are shown in order
  above the corresponding reconstructions. Upper and lower geometry views
  show 3D boxes and ground-plane footprints.
  Dashed footprints denote GT and triangles mark the reference origin.
  S/M denote Structure/Metric.}
  \label{fig:case_reconstruction}
  \label{fig:case_indoor}
\end{minipage}\par

\Needspace{22\baselineskip}
\subsection{Spatial Supervision and Transfer}
\label{sec:training}

To evaluate the effectiveness and transfer of our spatial supervision, we
fine-tune Qwen3.6-35B-A3B\footnote{\url{https://huggingface.co/Qwen/Qwen3.6-35B-A3B}}
on up to 50k examples spanning Geometric Perception, Spatial Relations, and
Reconstruction, using reasoning traces generated by Qwen3.8-Max and filtered
against the ground truth, without general-domain training data.
On external evaluations, the 50k
model improves nine of ten spatial and all eight general benchmarks
(Table~\ref{tab:training_transfer}). Across the ten public spatial benchmarks, including
SITE-Bench~\citep{wang2025site}, BLINK~\citep{fu2024blink},
ViewSpatial-Bench~\citep{li2025viewspatial}, and ERQA~\citep{gemini2025gemini_robotics},
the mean score rises from 60.98 to 65.13.
General benchmark gains include 7.9 points on VisualPuzzles~\citep{song2025visualpuzzles}
and 3.6 points on BabyVision~\citep{chen2026babyvision}.
Most external transfer gains are already present at 10k examples, with smaller
gains from further scaling. The full benchmark list and evaluation settings
are provided in Appendix~\ref{app:training_benchmarks}.

On \bench{}, performance improves steadily with training scale, from 24.39
to 37.17 Overall, with monotonic gains in the aggregate Geometric Perception,
Spatial Relations, and Reconstruction scores. Navigation success rises from
an 8.1\% baseline to at most 11.8\% and varies non-monotonically with data
scale. Cross-view spatial supervision thus yields broad external transfer
and steady gains in perception and reconstruction on \bench{}, while
translating these gains into reliable planning remains an open challenge.

\begingroup
\definecolor{trainSpatial}{HTML}{365E82}
\definecolor{trainGeneral}{HTML}{73558F}
\definecolor{trainRecon}{HTML}{A15B26}
\definecolor{trainNav}{HTML}{367262}
\definecolor{trainOverall}{HTML}{384959}
\definecolor{trainMuted}{HTML}{89939E}
\definecolor{trainGain}{HTML}{276B50}
\definecolor{trainDrop}{HTML}{A45032}
\newcommand{\trainingcell}[2]{%
  \colorbox{#1!9}{\makebox[3em][r]{\strut#2}}}
\newcommand{\traininghead}[2]{%
  \colorbox{#1!15}{\strut\textcolor{#1}{\textbf{#2}}}}
\newcommand{\trainingband}[2]{%
  \rlap{\colorbox{#1!13}{\makebox[\dimexpr\linewidth-2\fboxsep\relax]{\strut}}}%
  \hspace{\fboxsep}\strut\textcolor{#1}{\textbf{#2}}}

\par\noindent\begin{minipage}{\linewidth}
  \centering
  \small
  \renewcommand{\arraystretch}{1.32}
  \setlength{\tabcolsep}{2pt}
  \setlength{\fboxsep}{0.7pt}
  \captionof{table}{Spatial supervised fine-tuning (SFT) of Qwen3.6-35B-A3B on
  \textbf{(a)} spatial, \textbf{(b)} general, and \textbf{(c)} \bench{}
  evaluations. The external and \bench{} panels report all three spatial-data
  doses and their starting-checkpoint reference. Scores
  are percentages and column-best values are bold; GP/SR and S/M denote
  Geometric Perception/Spatial Relations and Structure/Metric.}
  \label{tab:training_transfer}
  \ifdefined\XeTeXversion
    \renewcommand{\bfdefault}{bx}
    \fontencoding{OT1}\fontfamily{cmr}\fontsize{10}{12}\selectfont
  \fi
  \newcommand{\trainingcompactcell}[2]{%
    \colorbox{#1!10}{\makebox[2.4em][r]{\strut#2}}}
  \resizebox{\linewidth}{!}{%
  \begin{minipage}{1.05\linewidth}
  \begin{tabular*}{\linewidth}{@{\extracolsep{\fill}}l@{\hspace{7pt}}*{11}{r}@{}}
\multicolumn{12}{@{}l@{}}{\trainingband{trainSpatial}{(a) Spatial benchmarks}} \\[3pt]
\toprule
\textbf{Variant} & SITE & BLINK & \shortstack{View\\Spatial} & ERQA & \shortstack{Omni\\Spatial} & 3DSR & MMSI & \shortstack{Emb\\Spatial} & \shortstack{SITE\\Video} & \shortstack{Ref\\Spatial} & \textcolor{trainSpatial}{\textbf{Avg.}} \\
\midrule
Base & \trainingcompactcell{white}{56.48} & \trainingcompactcell{white}{68.15} & \trainingcompactcell{white}{48.72} & \trainingcompactcell{white}{58.25} & \trainingcompactcell{white}{57.99} & \trainingcompactcell{white}{64.41} & \trainingcompactcell{white}{38.50} & \trainingcompactcell{white}{82.72} & \trainingcompactcell{white}{71.26} & \trainingcompactcell{white}{63.36} & \trainingcompactcell{trainSpatial}{60.98} \\
SFT 10k & \trainingcompactcell{white}{63.12} & \trainingcompactcell{white}{73.82} & \trainingcompactcell{white}{52.92} & \trainingcompactcell{white}{59.75} & \trainingcompactcell{white}{\textbf{62.95}} & \trainingcompactcell{white}{67.09} & \trainingcompactcell{white}{39.82} & \trainingcompactcell{white}{83.74} & \trainingcompactcell{white}{72.12} & \trainingcompactcell{white}{\textbf{65.46}} & \trainingcompactcell{trainSpatial}{64.08} \\
SFT 20k & \trainingcompactcell{white}{64.71} & \trainingcompactcell{white}{\textbf{76.62}} & \trainingcompactcell{white}{53.57} & \trainingcompactcell{white}{60.75} & \trainingcompactcell{white}{62.56} & \trainingcompactcell{white}{\textbf{68.66}} & \trainingcompactcell{white}{42.42} & \trainingcompactcell{white}{84.18} & \trainingcompactcell{white}{\textbf{73.22}} & \trainingcompactcell{white}{63.45} & \trainingcompactcell{trainSpatial}{65.01} \\
\textcolor{trainSpatial}{\textbf{SFT 50k}} & \trainingcompactcell{white}{\textbf{65.54}} & \trainingcompactcell{white}{75.42} & \trainingcompactcell{white}{\textbf{54.20}} & \trainingcompactcell{white}{\textbf{62.75}} & \trainingcompactcell{white}{62.30} & \trainingcompactcell{white}{68.53} & \trainingcompactcell{white}{\textbf{42.50}} & \trainingcompactcell{white}{\textbf{84.56}} & \trainingcompactcell{white}{72.84} & \trainingcompactcell{white}{62.64} & \trainingcompactcell{trainSpatial}{\textbf{65.13}} \\
\bottomrule
\end{tabular*}
\par\vspace{8pt}
\begin{tabular*}{\linewidth}{@{\extracolsep{\fill}}l@{\hspace{7pt}}*{9}{r}@{}}
\multicolumn{10}{@{}l@{}}{\trainingband{trainGeneral}{(b) General benchmarks}} \\[3pt]
\toprule
 & \multicolumn{2}{c}{\textcolor{trainGeneral}{\shortstack{Multimodal\\knowledge}}} & \multicolumn{2}{c}{\textcolor{trainGeneral}{\shortstack{Visual\\reasoning}}} & \multicolumn{2}{c}{\textcolor{trainGeneral}{\shortstack{Charts and\\diagrams}}} & \multicolumn{2}{c}{\textcolor{trainGeneral}{\shortstack{Scene / perceptual\\reliability}}} & \\
\cmidrule(lr){2-3}\cmidrule(lr){4-5}\cmidrule(lr){6-7}\cmidrule(lr){8-9}
\textbf{Variant} & MMMU & \shortstack{MMBench\\CN} & \shortstack{Visual\\Puzzles} & \shortstack{Baby\\Vision} & CharXiv & AI2D & \shortstack{Real\\WorldQA} & \shortstack{Hallusion\\Bench} & \textcolor{trainGeneral}{\textbf{Avg.}} \\
\midrule
Base & \trainingcompactcell{white}{80.89} & \trainingcompactcell{white}{89.94} & \trainingcompactcell{white}{53.42} & \trainingcompactcell{white}{24.23} & \trainingcompactcell{white}{74.70} & \trainingcompactcell{white}{92.26} & \trainingcompactcell{white}{82.75} & \trainingcompactcell{white}{78.55} & \trainingcompactcell{trainGeneral}{72.09} \\
SFT 10k & \trainingcompactcell{white}{81.56} & \trainingcompactcell{white}{90.79} & \trainingcompactcell{white}{\textbf{63.36}} & \trainingcompactcell{white}{\textbf{30.67}} & \trainingcompactcell{white}{76.30} & \trainingcompactcell{white}{92.55} & \trainingcompactcell{white}{82.88} & \trainingcompactcell{white}{76.34} & \trainingcompactcell{trainGeneral}{74.30} \\
SFT 20k & \trainingcompactcell{white}{82.00} & \trainingcompactcell{white}{90.94} & \trainingcompactcell{white}{62.24} & \trainingcompactcell{white}{27.06} & \trainingcompactcell{white}{\textbf{78.10}} & \trainingcompactcell{white}{\textbf{92.78}} & \trainingcompactcell{white}{83.01} & \trainingcompactcell{white}{78.13} & \trainingcompactcell{trainGeneral}{74.28} \\
\textcolor{trainGeneral}{\textbf{SFT 50k}} & \trainingcompactcell{white}{\textbf{82.56}} & \trainingcompactcell{white}{\textbf{91.56}} & \trainingcompactcell{white}{61.30} & \trainingcompactcell{white}{27.84} & \trainingcompactcell{white}{77.00} & \trainingcompactcell{white}{\textbf{92.78}} & \trainingcompactcell{white}{\textbf{83.66}} & \trainingcompactcell{white}{\textbf{79.92}} & \trainingcompactcell{trainGeneral}{\textbf{74.58}} \\
\bottomrule
\end{tabular*}

  \par\vspace{8pt}
  \begin{tabular*}{\linewidth}{@{\extracolsep{\fill}}lrrrrrrr@{}}
\multicolumn{8}{@{}l@{}}{\trainingband{trainRecon}{(c) \bench{}}} \\[3pt]
\toprule
\textbf{Configuration} & \traininghead{trainSpatial}{GP} & \traininghead{trainGeneral}{SR} & \multicolumn{3}{c}{\traininghead{trainRecon}{Reconstruction}} & \traininghead{trainNav}{Nav.} & \traininghead{trainOverall}{Overall} \\
\cmidrule(lr){4-6}
 & Avg. & Avg. & S & M & Avg. & & \\
\midrule
Base (reference) & \trainingcell{white}{38.44} & \trainingcell{white}{20.03} & \trainingcell{white}{42.52} & \trainingcell{white}{19.48} & \trainingcell{white}{31.00} & \trainingcell{white}{8.10} & \trainingcell{white}{24.39} \\
Spatial SFT (10k) & \trainingcell{white}{49.29} & \trainingcell{white}{35.91} & \trainingcell{white}{50.11} & \trainingcell{white}{26.48} & \trainingcell{white}{38.29} & \trainingcell{white}{\textbf{11.80}} & \trainingcell{white}{33.82} \\
Spatial SFT (20k) & \trainingcell{white}{50.43} & \trainingcell{white}{40.87} & \trainingcell{white}{52.70} & \trainingcell{white}{26.33} & \trainingcell{white}{39.52} & \trainingcell{white}{8.40} & \trainingcell{white}{34.80} \\
\textcolor{trainSpatial}{\textbf{Spatial SFT (50k)}} & \trainingcell{trainSpatial}{\textbf{54.22}} & \trainingcell{trainSpatial}{\textbf{43.37}} & \trainingcell{trainSpatial}{\textbf{54.74}} & \trainingcell{trainSpatial}{\textbf{26.60}} & \trainingcell{trainSpatial}{\textbf{40.67}} & \trainingcell{trainSpatial}{10.40} & \trainingcell{trainSpatial}{\textbf{37.17}} \\
\bottomrule
\end{tabular*}

  \end{minipage}}
\end{minipage}\par
\endgroup

\Needspace{14\baselineskip}
\section{Conclusion and Limitations}

In this work, \bench{} establishes multi-view 3D scene reconstruction as a
principled benchmark connecting local spatial perception with holistic scene
coherence and downstream planning. Across 18 frontier LVLMs, our evaluations
uncover a persistent gap between relative scene structure and metric
reconstruction. Diagnostic analyses show that relational consistency alone
cannot guarantee metric accuracy, while downstream map substitution
demonstrates that navigation failures stem primarily from perceptual geometry
rather than planning logic. Furthermore, cross-view spatial fine-tuning
confers broad transfer to both spatial and general vision-language benchmarks.

\paragraph{Limitations and Future Work.}
\label{sec:limitations}

Several limitations outline avenues for future work. First, physical objects
are represented via oriented 3D bounding boxes, omitting fine-grained surface
meshes, articulated parts, and continuous free-space occupancy. Second, our
navigation evaluation focuses on offline, open-loop planning within single
rooms to decouple execution noise, leaving closed-loop recovery and multi-room
exploration unexamined. Finally, spatial fine-tuning was evaluated on a single
model family. Scaling these evaluations to broader architectures and
interactive physical environments represents a critical frontier for building
dependable embodied spatial intelligence.

\section*{Ethics Statement}

\paragraph{Data Sourcing and Privacy.}
\bench{} builds upon existing open research datasets and in-house robot
recordings (Appendix~\ref{app:construction}). All asset curation and
redistribution strictly comply with the original licenses and permissible use
policies. To protect privacy, all real-world scenes have been screened to
ensure they contain no personally identifiable information.

\paragraph{Broader Impact and Safety.}
This benchmark aims to advance foundational 3D spatial perception and
reasoning. All navigation tasks are evaluated strictly offline using
pre-recorded observations and maps, involving no physical robotic actuation
or safety risks. While enhanced metric accuracy is essential for building
more dependable embodied agents, real-world deployment requires independent,
rigorous safety verification and failsafe mechanisms.

\section*{AI Use Statement}

We used LLMs to assist with manuscript editing, code development and figure
preparation. Separately, Qwen3.8-Max was employed to generate reasoning traces
and target responses for spatial fine-tuning
(Appendix~\ref{app:training_benchmarks}). All AI-assisted outputs, code, and
synthesized data were manually verified by the authors, who take full
responsibility for all methods, results, and claims presented in this work.

\bibliography{main}
\ifvenuestyle
  \bibliographystyle{iclr2027_conference}
\else
  \bibliographystyle{plainnat}
\fi

\clearpage
\appendix
\section{Task Construction, Evaluation, and Examples}
\label{app:tasks}
\label{app:construction}

\bench{} is constructed around cross-view scenes: answering a query requires
combining observations from multiple viewpoints, since no single view provides
the complete task evidence. We organize each source into scene records linking
images, camera geometry, and available point- or object-level annotations, from
which we derive task-specific questions and reference answers; all scores are
computed analytically from these annotations. For each task, we describe its
inputs, reference frame, answer space, and construction; for reconstruction and
navigation, we also give the evaluation protocol.
Table~\ref{tab:composition} reports task-level domains, counts, and input views,
and Section~\ref{app:task_examples} gives complete examples with visual inputs,
prompts, and reference answers.

\begin{table}[!h]
  \centering
  \small
  \setlength{\tabcolsep}{6pt}
  \renewcommand{\arraystretch}{1.17}
  \caption{Data composition of \bench{} across L1--L4.}
  \label{tab:composition}
  \begin{tabularx}{\linewidth}{@{}>{\hsize=1.2\hsize\linewidth=\hsize\raggedright\arraybackslash}X
  >{\hsize=.8\hsize\linewidth=\hsize\raggedright\arraybackslash}X r c r@{}}
\toprule
\textbf{Task} & \textbf{Domain} & \textbf{Instances} & \textbf{Views} & \textbf{Chance (\%)} \\
\midrule
\rowcolor{black!5}[0pt][\tabcolsep]
\multicolumn{5}{@{}l}{\strut\textbf{L1: Geometric Perception}} \\
\multirow{3}{=}{Pixel correspondence} & Indoor & 56 & \multirow{3}{*}{2} & \multirow{3}{*}{16.8} \\
 & Tabletop & 118 & & \\
 & Outdoor & 42 & & \\
\addlinespace[2pt]
Camera pose & Indoor, outdoor & 361 & 2 & 15.6 \\
Depth ordering & Indoor & 55 & 2 & 12.5 \\
Metric depth & Indoor & 107 & 2 & 31.2 \\
\addlinespace[4pt]
\rowcolor{black!5}[0pt][\tabcolsep]
\multicolumn{5}{@{}l}{\strut\textbf{L2: Spatial Relations}} \\
Camera-anchored relation & Indoor & 227 & 2 & 12.5 \\
Object-anchored relation & Indoor & 63 & 2 & 12.5 \\
Distance ranking & Indoor & 197 & 2 & 16.7 \\
Object correspondence & Indoor & 66 & 2--6 & 20.7 \\
Nearest object & Indoor & 96 & 3 & 25.0 \\
\addlinespace[4pt]
\rowcolor{black!5}[0pt][\tabcolsep]
\multicolumn{5}{@{}l}{\strut\textbf{L3: Scene Reconstruction}} \\
\multirow{3}{=}{Oriented-box reconstruction} & Indoor & 475 & 2 / 8 & \multirow{3}{*}{--} \\
 & Tabletop & 89 & 2 & \\
 & Outdoor & 276 & 6 & \\
\addlinespace[4pt]
\rowcolor{black!5}[0pt][\tabcolsep]
\multicolumn{5}{@{}l}{\strut\textbf{L4: Navigation}} \\
Offline path prediction & Indoor & 346 & 11--23 & -- \\
\bottomrule
\end{tabularx}

  \par\vspace{4pt}
  \begin{minipage}{\linewidth}
    \footnotesize
    \textit{Note.} Counts are task instances, not distinct physical scenes;
    views denote input frames. Chance (\%) is the mean inverse number of
    options per item; for metric depth, it is the score of a constant-depth
    prediction.
  \end{minipage}
\end{table}

\subsection{Geometric Perception}

\paragraph{Pixel correspondence.}
\label{app:construction_pc}
The input contains two views and a query marker in the first. The second view
contains five to eight colored candidate markers, and the model selects the
marker corresponding to the same physical surface point. Candidate locations
are sampled around the projected correspondence so that image proximity alone
does not identify the answer. Items are scored by exact multiple-choice
accuracy. The data cover indoor scenes from ScanNet~\citep{dai2017scannet},
outdoor scenes from Sekai~\citep{li2025sekai}, and robot observations from
AgiBot World~\citep{agibotworld2025}, BridgeData V2~\citep{walke2023bridgedatav2},
RoboMIND~\citep{wu2025robomind}, Galaxea Open-World~\citep{jiang2025galaxea},
RDT-1B-associated recordings~\citep{liu2025rdt}, and in-house data. Robot
examples pair distinct viewpoints from head-mounted, wrist-mounted, and
external third-person cameras, according to the cameras available in each
source. Existing correspondence annotations are used where available. For
robot observations, we detect target regions and use
MASt3R~\citep{mast3r_eccv24} to obtain cross-view matches and 3D point maps,
retaining candidate matches whose associated 3D points agree spatially.

\paragraph{Camera pose.}
\label{app:construction_cp}
Two views are presented without intrinsics or field of view. Translation is
quantized into eight egocentric directions relative to the first camera;
rotation is represented by its sense and magnitude. The reported score is the
mean of translation and rotation accuracy. This discrete formulation avoids
conflating geometric understanding with unstable free-form numeric formatting,
and querying translation as a direction accommodates sources with poses
defined only up to scale. View pairs and their relative camera transformations
come from ScanNet~\citep{dai2017scannet},
ScanNet++~\citep{yeshwanth2023scannetpp}, ETH3D~\citep{schops2017eth3d}, and
Sekai~\citep{li2025sekai}.

\paragraph{Depth ordering and metric depth.}
\label{app:construction_do}
\label{app:construction_md}
Both tasks use ScanNet scenes with EmbodiedScan
annotations~\citep{dai2017scannet,wang2024embodiedscan} and
CA-1M~\citep{lazarow2025cubify}. Target objects are selected across views, and
their annotated 3D centres are transformed into a designated reference camera
frame; their coordinates along its viewing axis define the reference depths.
In depth ordering, four objects are designated by boxes across two views, and
the model orders their centres by depth along the second camera's viewing
axis. The output is one of eight candidate permutations and is scored by exact
accuracy; every adjacent pair in the sorted sequence differs by at least
$0.3\,\mathrm{m}$ to avoid ambiguous orderings. In metric depth, the model
receives two views with four named, boxed objects and predicts each object's
depth in metres along a designated reference camera axis, as a JSON mapping
from object identifiers to real-valued depths. At least one object is marked
in the other view, requiring its depth to be expressed in the reference camera
frame. We report the mean accuracy at relative-error thresholds
$\{5,10,20,40\}\%$; the ordering-specific separation criterion is not imposed.

\subsection{Spatial Relations}

Spatial relation tasks draw on object-level geometry from ScanNet
images~\citep{dai2017scannet} with EmbodiedScan oriented-box
annotations~\citep{wang2024embodiedscan}, and from
CA-1M~\citep{lazarow2025cubify}. Scene records associate object centres,
orientations, instance identities, and visibility across views, and each query
distributes the required object evidence across views.

\paragraph{Camera-anchored relation.}
\label{app:construction_rel}
Two objects are selected from different views so that neither view contains
both. The model adopts the first camera's orientation and classifies the
direction of one object relative to the other into eight horizontal sectors;
the reference answer maps their relative displacement in that frame to a
sector. The reference-frame requirement distinguishes cross-view perspective
taking from a comparison of image-plane positions.

\paragraph{Object-anchored relation.}
\label{app:construction_allo}
The reference frame is defined by an anchor object's own front direction rather
than by a camera. The position of a target observed in another view is
transformed into this frame to determine its direction relative to the anchor,
using the same eight-direction output space as the camera-anchored task.
Anchors are restricted to ScanNet categories with a meaningful front, and
candidate examples are manually checked for orientation ambiguity.

\paragraph{Distance ranking.}
\label{app:construction_dist}
Each item specifies three object pairs whose members may occur in different
views. The model ranks the pairs by the Euclidean 3D distance between their
centres, and the six possible permutations form the answer space. This task
measures inter-object metric structure and differs from depth ordering, which
is defined along a camera axis.

\paragraph{Object correspondence (OC).}
\label{app:construction_id}
This task uses same-category, size-similar instances that cannot be reliably
distinguished by their names; instance identities and per-view visibility
determine the answer. In the matching variant, the first image marks a target
with a red box and the query view marks candidates with distinct colors; the
model selects the same instance or reports that it is not visible. In the
missing-instance variant, context views assign colors to several instances and
the model identifies which one is absent from the query view. The two variants
are micro-averaged into one score.

\paragraph{Nearest object.}
\label{app:construction_near}
A reference object and four candidates are distributed across three views. The
model selects the candidate whose annotated 3D centre is closest to the
reference, scored by four-way accuracy. Construction records the true distance
margin and screens simple image-space cues such as box area and image centre.

\subsection{Scene Reconstruction}

\paragraph{Construction.}
\label{app:construction_recon}
Indoor scenes come from ScanNet images~\citep{dai2017scannet} with
EmbodiedScan oriented-box annotations~\citep{wang2024embodiedscan}, and from
CA-1M~\citep{lazarow2025cubify} with its own box annotations; tabletop scenes from
YCB-Video~\citep{xiang2018posecnn}, T-LESS~\citep{hodan2017tless},
HomebrewedDB~\citep{kaskman2019homebreweddb}, and NOCS~\citep{wang2019nocs};
and outdoor scenes from nuScenes~\citep{caesar2020nuscenes}. We transform
the object annotations into a gravity-aligned frame anchored to the
reference camera, using the ground or tabletop as the height reference.
Selected views provide complementary observations of the scene, and
projection associates the queried 3D boxes with image regions. Each
instance pairs these multi-view observations and object queries with
oriented-box annotations in the shared frame, requiring the partial
observations to be combined into a joint layout.

\paragraph{Output format and reproducibility.}
\label{app:scoring}
Reconstruction instances query an average of 6.73, 5.64, and 8.62 objects in
indoor, tabletop, and outdoor scenes. Centres and extents are in metres, with
$+x$ to the right, $+y$ upward, and $-z$ along the reference camera's horizontal
viewing direction; the origin lies below the reference camera on the floor,
ground, or table surface, and the front camera serves as the reference view for
nuScenes. Responses must be a JSON object listing each queried object's
identifier, category, centre, size, and yaw; the parser performs no generative
repair.

Predicted boxes are matched by queried object ID. Instances with fewer than
three matched objects receive zero on both headline scores; otherwise the
geometric components use the matched subset. Coverage is reported separately
and does not enter Structure or Metric. Saved empty or unparseable outputs
receive zero on both headline scores. The random-layout reference in
Structure uses a fixed per-instance seed, so all scores are deterministic.

\paragraph{Scorer controls.}
Controls built from the ground truth (Table~\ref{tab:scorer_controls}) confirm
that each score responds to its intended property: random rescaling leaves
Structure at 100.0 but lowers Metric to 76.9, depth flattening lowers both, and
shuffled assignments reduce both to near their floors. A fully corrupted
prediction, with shuffled assignments and a random scale, scores lower on
Structure (8.9) than on Metric (15.2), so the Structure--Metric gap does not
arise from Metric having a lower floor.

\begin{table}[t]
  \centering
  \small
  \caption{Reconstruction scorer controls on all 840 instances, scored with the
  official per-instance seeds and equal domain weights. Each control alters the
  ground truth: shuffled boxes permute object centres, depth flattening sets all
  depths to the scene mean, and random scale applies one log-uniform factor in
  $[1/3,3]$. Stochastic controls average eight construction seeds.}
  \label{tab:scorer_controls}
  \begin{tabular}{lrrrrr}
\toprule
Control & Structure & Metric & Pos. & View IoU & Scale \\
\midrule
GT copy & 100.0 & 100.0 & 100.0 & 100.0 & 100.0 \\
Shuffled boxes & 8.7 & 23.2 & 12.8 & 8.2 & 48.6 \\
Depth flattened & 50.2 & 63.9 & 34.3 & 57.3 & 100.0 \\
Random scale & 100.0 & 76.9 & 100.0 & 100.0 & 30.6 \\
Shuffled + scale & 8.9 & 15.2 & 12.9 & 8.4 & 24.4 \\
\bottomrule
\end{tabular}

\end{table}

\paragraph{Sensitivity to scoring choices.}
\label{app:scorer_sensitivity}
Model rankings remain stable under alternative component weights and scale
tolerances, as shown in Table~\ref{tab:scorer_sensitivity}. All variants use
the same 840-instance scores, retain failed outputs as zero, and weight the
three domains equally. Removing the scale term changes the headline ranking
most: position and overlap already allow a fitted scale, so removing its
penalty also removes the requirement for physical calibration.

As a completeness check, we multiply each instance's Structure and Metric
by the fraction of queried objects matched, retaining the same domain
aggregation. Rankings on both scores and their reconstruction average
remain unchanged; the average decreases by at most 0.17 percentage
points, and Structure exceeds Metric in every model--domain combination.

\begin{table}[t]
  \centering
  \small
  \caption{Spearman correlation between the eighteen-model ranking under each
  alternative scoring convention and the headline Structure or Metric ranking
  on the full reconstruction suite. Default scoring weights each Structure
  component and each Metric component (Pos.\ / View IoU / Scale) equally, with
  a $2\times$ scale tolerance in Eq.~\eqref{eq:scale_score}.}
  \label{tab:scorer_sensitivity}
  \begin{tabular}{llr}
\toprule
Alternative scoring convention & Ranking & Spearman $\rho$ \\
\midrule
Structure: five channels weighted equally & Structure & 1.000 \\
Metric: weights 0.4 / 0.2 / 0.4 & Metric & 1.000 \\
Metric: weights 0.25 / 0.25 / 0.5 & Metric & 0.994 \\
Metric: weights 0.5 / 0.5 / 0 (no scale term) & Metric & 0.953 \\
Scale tolerance $1.5\times$ & Metric & 0.996 \\
Scale tolerance $2\times$ (default) & Metric & 1.000 \\
Scale tolerance $3\times$ & Metric & 0.988 \\
\bottomrule
\end{tabular}

\end{table}

\subsection{Navigation}
\label{app:navigation_scoring}

\paragraph{Offline path prediction.}
\label{app:construction_nav}
Navigation episodes use first-person images and camera trajectories from
ScanNet~\citep{dai2017scannet}, with obstacle maps derived from
EmbodiedScan object geometry~\citep{wang2024embodiedscan}. We organize the
observations into a numbered tour sequence and select a starting viewpoint
and a target object. The start is specified by an image frame, with a
centred scene object indicating the initial heading; the goal is specified
by its object name, frame index, and image bounding box. Its metric
coordinates must be inferred from the observations.

The task is to predict a path that avoids furniture and obstacles and
reaches the target's vicinity. The output is an ordered sequence of
ground-plane waypoints in metres, with the origin at the start, $x$
pointing right, and $y$ forward (with $+z$ upward for 3D box maps, unlike the
$y$-up reconstruction convention), serialized as
\begin{quote}
\raggedright\ttfamily
<bev\_task\_pred coords="0 x1 y1; 0 x2 y2; ... 0 xn yn">\\
</bev\_task\_pred>
\end{quote}
Here, \texttt{0} is the fixed track identifier of the moving agent.
Reference routes are obtained from recorded camera-trajectory segments
or planned on the obstacle map and expressed in the same start-centred
frame. Evaluation assesses target arrival and collision avoidance,
allowing alternative feasible routes.

\paragraph{Collision checking.}
The avoidability-aware check uses annotated ground-plane obstacle footprints.
At a sampled path point inside an obstacle, a shallow graze is tolerated if
free floor exists within 0.2 metres on either side, perpendicular to the
direction of travel. A crossing fails when neither side offers such an escape.
Room-boundary crossings are always failures, and the target object is excluded
from collision checking. This tolerance accounts for small lateral errors;
it is not a requirement for a 0.2-metre empty buffer around the path.
Any path satisfying both arrival and collision conditions is accepted,
without requiring agreement with a particular reference traversal.
Unparseable paths score zero for all models, without model-specific parser
relaxation.

\clearpage
\subsection{Task examples}
\label{app:task_examples}

Each example groups its visual input, prompt, and reference answer.

\begin{casecard}{01}{Pixel correspondence}{Geometric perception}
{\centering \includegraphics[height=88pt,width=0.493\linewidth,keepaspectratio]{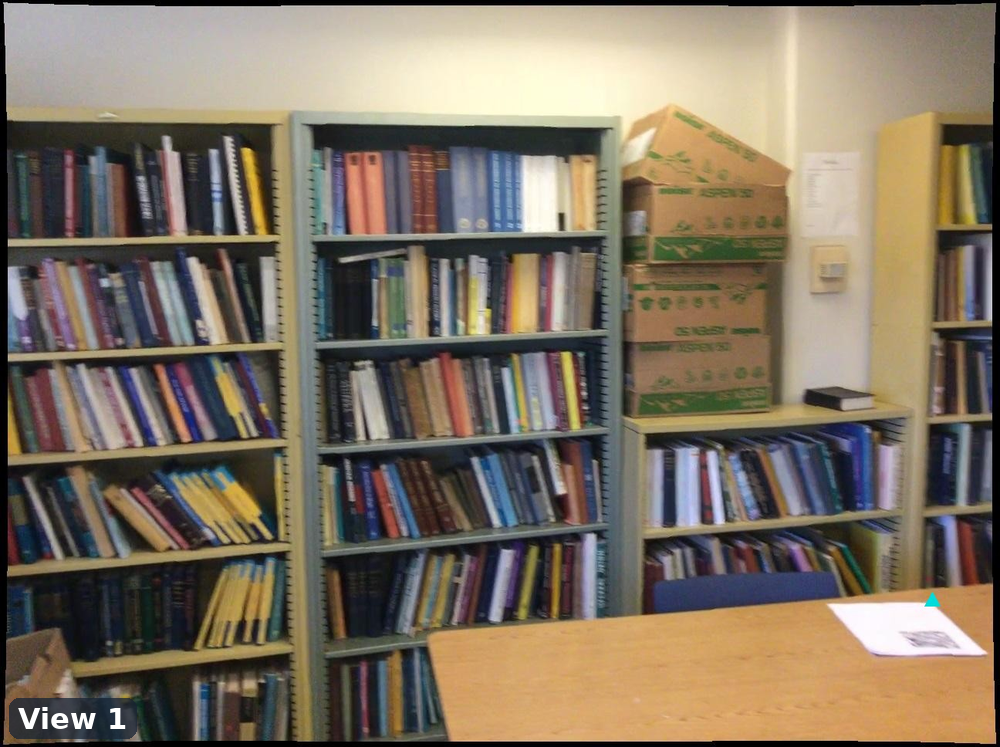}\hspace{6pt}\includegraphics[height=88pt,width=0.493\linewidth,keepaspectratio]{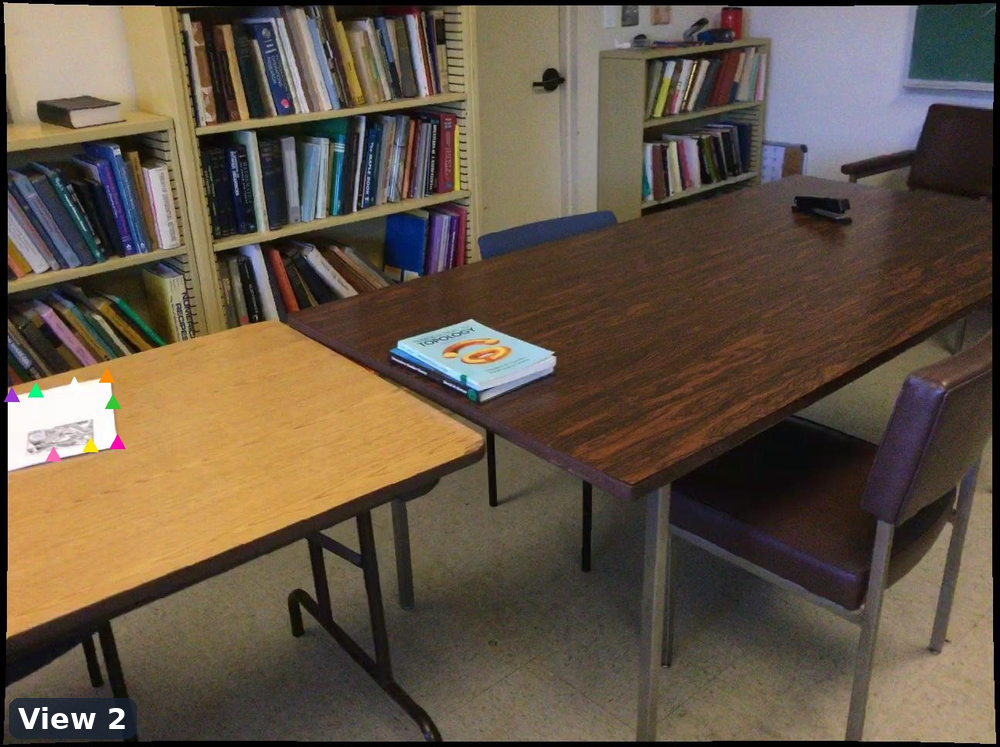}\par}
\casesection{Prompt}
\begin{promptbox}
The Cyan triangle in the first image marks a specific location. Which colored marker in the second image shows the same physical point?

\caseoptions{{\setlength{\tabcolsep}{0pt}\renewcommand{\arraystretch}{1.15}\begin{tabular}{@{}*{4}{p{0.250\linewidth}}@{}}\textbf{A.}\,Lime & \textbf{B.}\,Magenta & \textbf{C.}\,White & \textbf{D.}\,Purple \\ \textbf{E.}\,Green & \textbf{F.}\,Yellow & \textbf{G.}\,Pink & \caseright{\textbf{H.}\,Orange}\end{tabular}}}

Please answer with the correct option.
\end{promptbox}
\casesection{Reference answer}
\caseanswer{H. Orange}
\end{casecard}

\begin{casecard}{02}{Camera pose}{Geometric perception}
{\centering \includegraphics[height=88pt,width=0.493\linewidth,keepaspectratio]{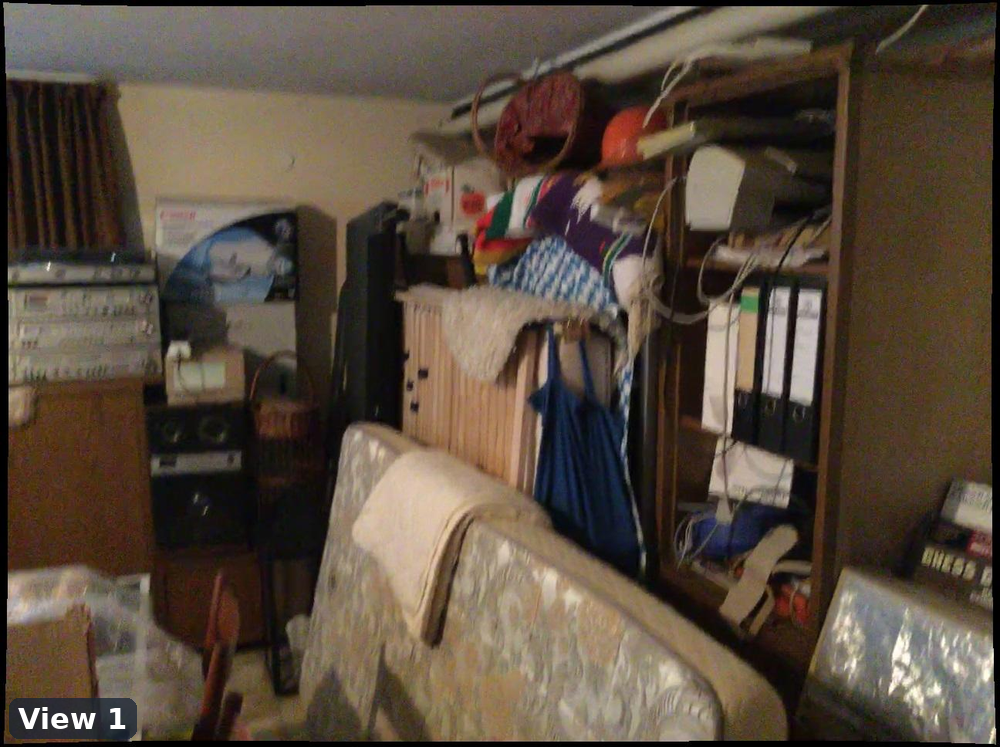}\hspace{6pt}\includegraphics[height=88pt,width=0.493\linewidth,keepaspectratio]{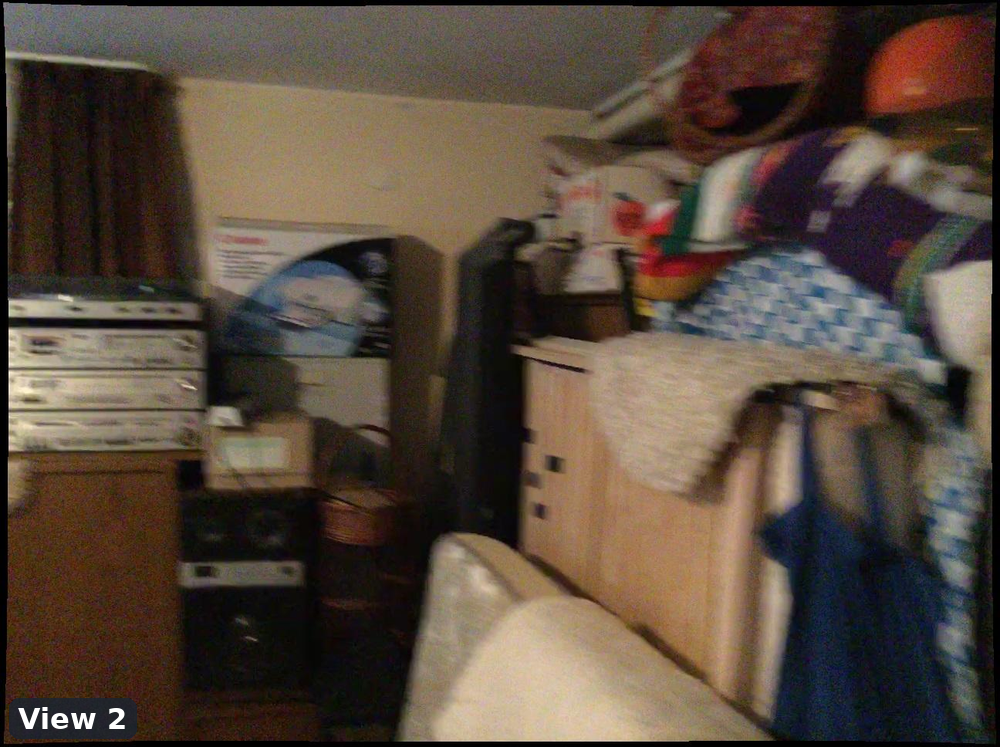}\par}
\casesection{Prompt}
\begin{promptbox}
These two images were captured by a moving camera: the first image is viewpoint 1 and the second image is viewpoint 2 of the same scene. Relative to the FIRST camera (looking along its own forward direction), in which direction did the camera MOVE from viewpoint 1 to viewpoint 2?

\caseoptions{{\setlength{\tabcolsep}{0pt}\renewcommand{\arraystretch}{1.15}\begin{tabular}{@{}*{4}{p{0.250\linewidth}}@{}}\caseright{\textbf{A.}\,Forward} & \textbf{B.}\,Backward-left & \textbf{C.}\,Left & \textbf{D.}\,Forward-left \\ \textbf{E.}\,Backward & \textbf{F.}\,Right & \textbf{G.}\,Backward-right & \textbf{H.}\,Forward-right\end{tabular}}}

Please answer directly with only the letter of the correct option and nothing else.
\end{promptbox}
\casesection{Reference answer}
\caseanswer{A. Forward}
\end{casecard}

\clearpage

\begin{casecard}{03}{Depth ordering}{Geometric perception}
{\centering \includegraphics[height=88pt,width=0.493\linewidth,keepaspectratio]{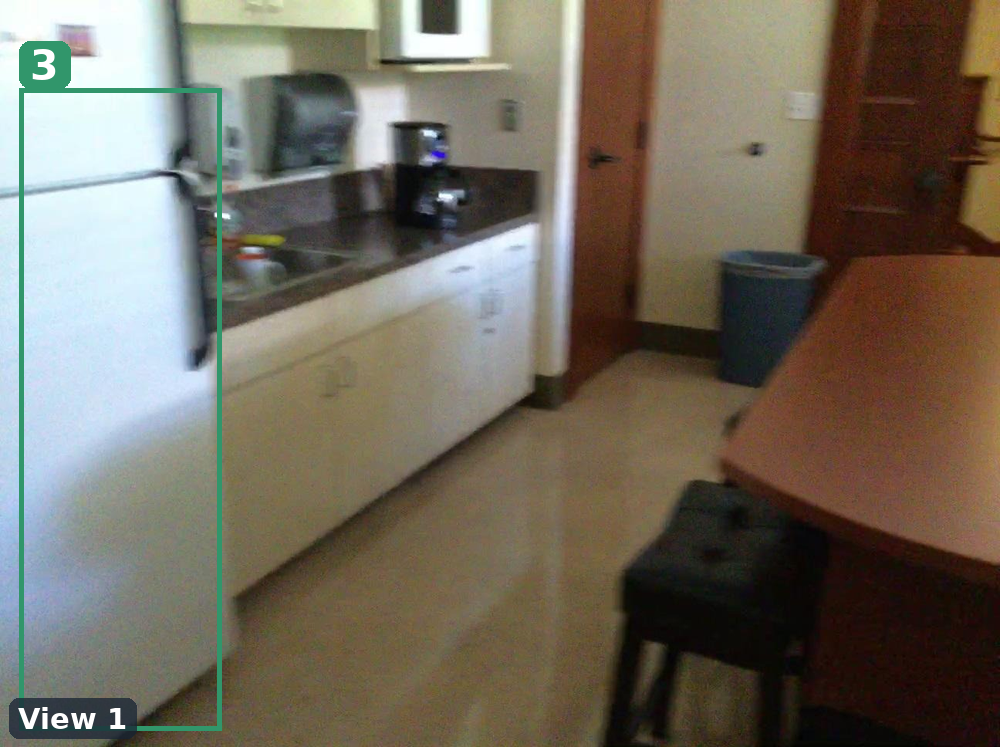}\hspace{6pt}\includegraphics[height=88pt,width=0.493\linewidth,keepaspectratio]{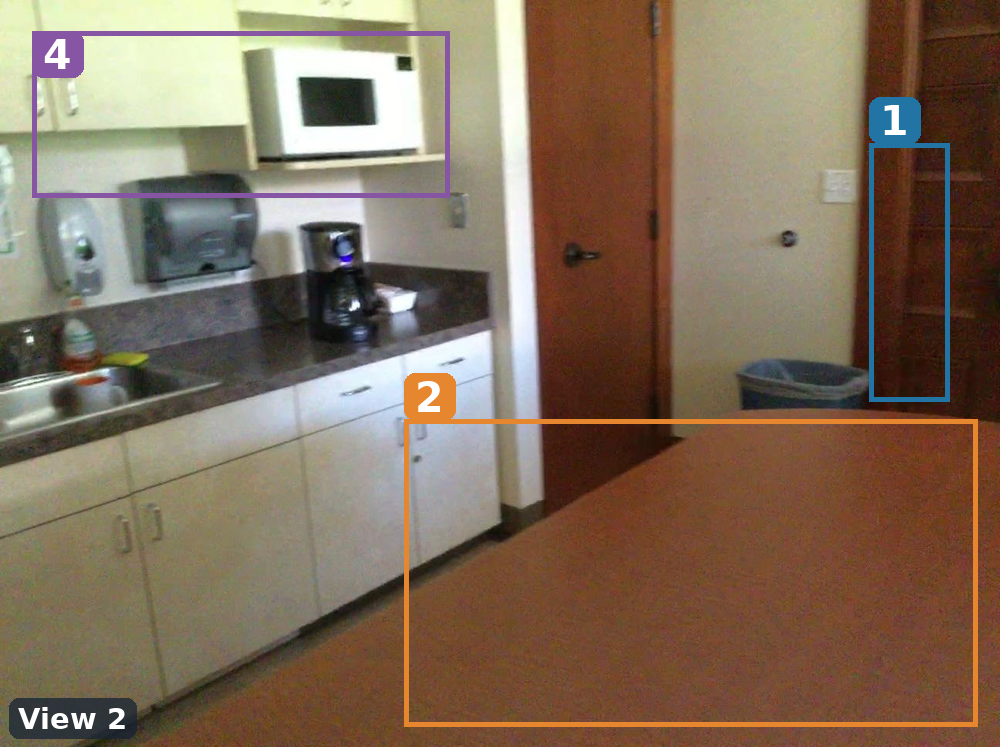}\par}
\casesection{Prompt}
\begin{promptbox}
Two photos show the same room from different viewpoints. Consider these 4 objects: object 1 = the doorframe (box [869, 192, 949, 537] in image 2); object 2 = the table (box [404, 562, 977, 972] in image 2); object 3 = the refrigerator (box [19, 118, 221, 978] in image 1); object 4 = the cabinet (box [32, 42, 449, 264] in image 2). Order them by DEPTH along IMAGE 2's viewing direction, from NEAREST (closest to the image-2 camera) to FARTHEST. Each option lists object numbers nearest-first.

\caseoptions{{\setlength{\tabcolsep}{0pt}\renewcommand{\arraystretch}{1.15}\begin{tabular}{@{}*{4}{p{0.250\linewidth}}@{}}\textbf{A.}\,(4, 3, 1, 2) & \textbf{B.}\,(3, 2, 4, 1) & \textbf{C.}\,(3, 4, 1, 2) & \textbf{D.}\,(1, 2, 4, 3) \\ \textbf{E.}\,(1, 2, 3, 4) & \textbf{F.}\,(1, 4, 3, 2) & \caseright{\textbf{G.}\,(2, 3, 4, 1)} & \textbf{H.}\,(4, 3, 2, 1)\end{tabular}}}

Answer with a single option letter.
\end{promptbox}
\casesection{Reference answer}
\caseanswer{G. (2, 3, 4, 1)}
\end{casecard}

\begin{casecard}{04}{Metric depth}{Geometric perception}
{\centering \includegraphics[height=88pt,width=0.493\linewidth,keepaspectratio]{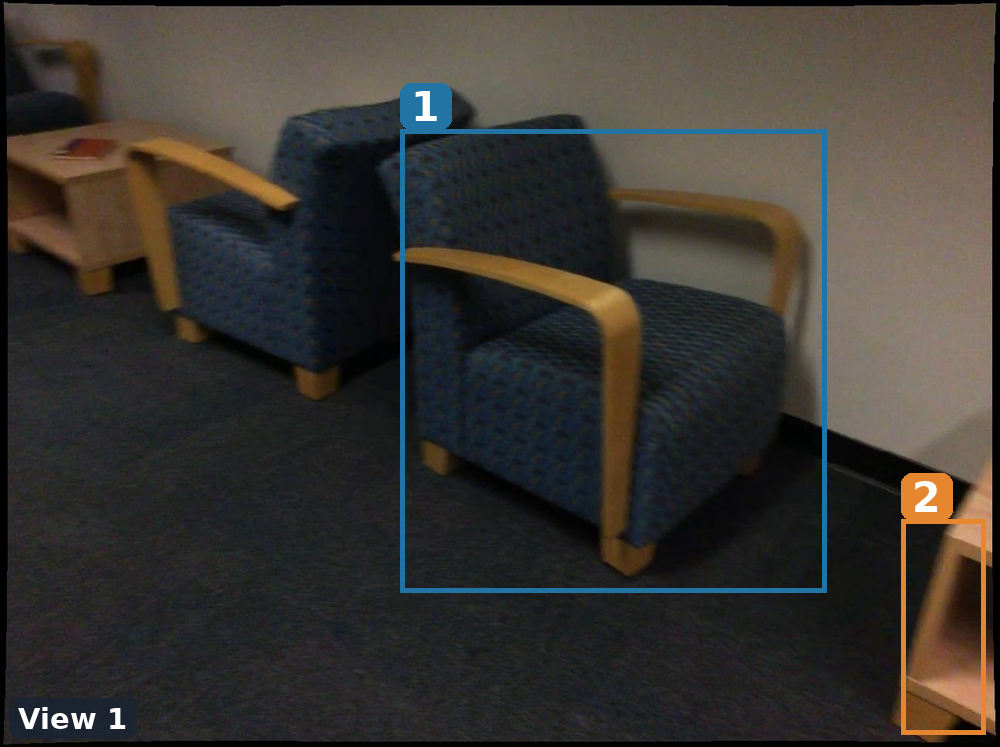}\hspace{6pt}\includegraphics[height=88pt,width=0.493\linewidth,keepaspectratio]{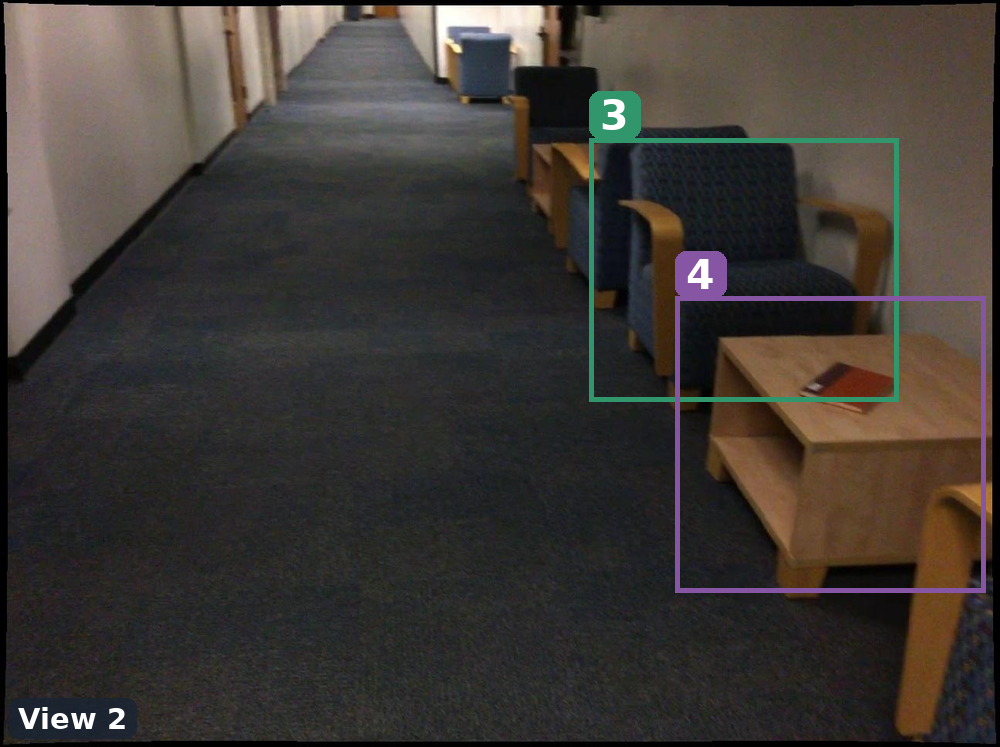}\par}
\casesection{Prompt}
\begin{promptbox}
Two photos show the same room from different viewpoints. Consider these 4 objects: object 1 = the chair (box [400, 173, 826, 793] in image 1); object 2 = the table (box [901, 696, 985, 983] in image 1); object 3 = the chair (box [589, 185, 898, 537] in image 2); object 4 = the table (box [675, 397, 985, 793] in image 2). Boxes are [x0, y0, x1, y1] in pixel coordinates normalized to 0-1000 (origin at the top-left, x rightward, y downward).

For EACH object, estimate its DEPTH along IMAGE 1's viewing direction: how far in metres the object's centre is in front of the IMAGE-1 camera, measured along that camera's viewing direction (i.e. its depth, NOT the straight-line distance to the camera).

Answer with real metric values in metres, to two decimals. Do not answer with ranges or words.

Output ONE JSON object and NOTHING else:\newline {\casemono \char123{}\char34{}depth\char95{}m\char34{}:\ \char123{}\char34{}1\char34{}:\ \char60{}number\char62{},\ \char34{}2\char34{}:\ \char60{}number\char62{},\ \char34{}3\char34{}:\ \char60{}number\char62{},\ \char34{}4\char34{}:\ \char60{}number\char62{}\char125{}\char125{}}
\end{promptbox}
\casesection{Reference answer}
\caseanswercode{\char123{}\char34{}depth\char95{}m\char34{}:\ \char123{}\char34{}1\char34{}:\ 2.0,\ \char34{}2\char34{}:\ 1.25,\ \char34{}3\char34{}:\ 4.03,\ \char34{}4\char34{}:\ 3.46\char125{}\char125{}}
\end{casecard}

\clearpage

\begin{casecard}{05}{Camera-anchored relation}{Spatial relations}
{\centering \includegraphics[height=88pt,width=0.493\linewidth,keepaspectratio]{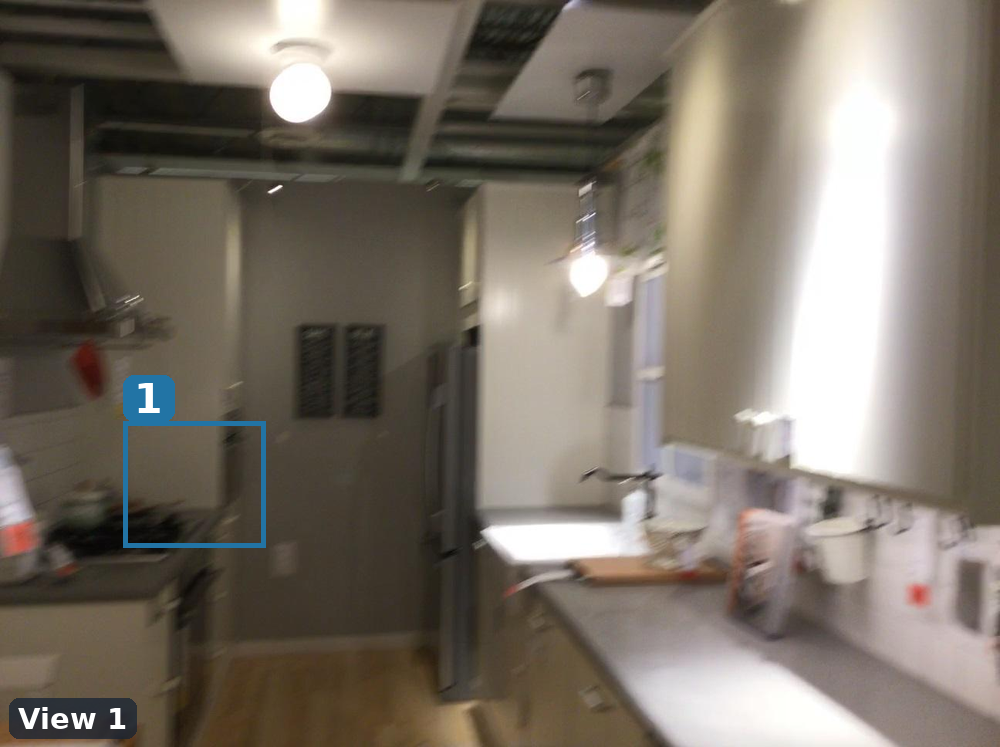}\hspace{6pt}\includegraphics[height=88pt,width=0.493\linewidth,keepaspectratio]{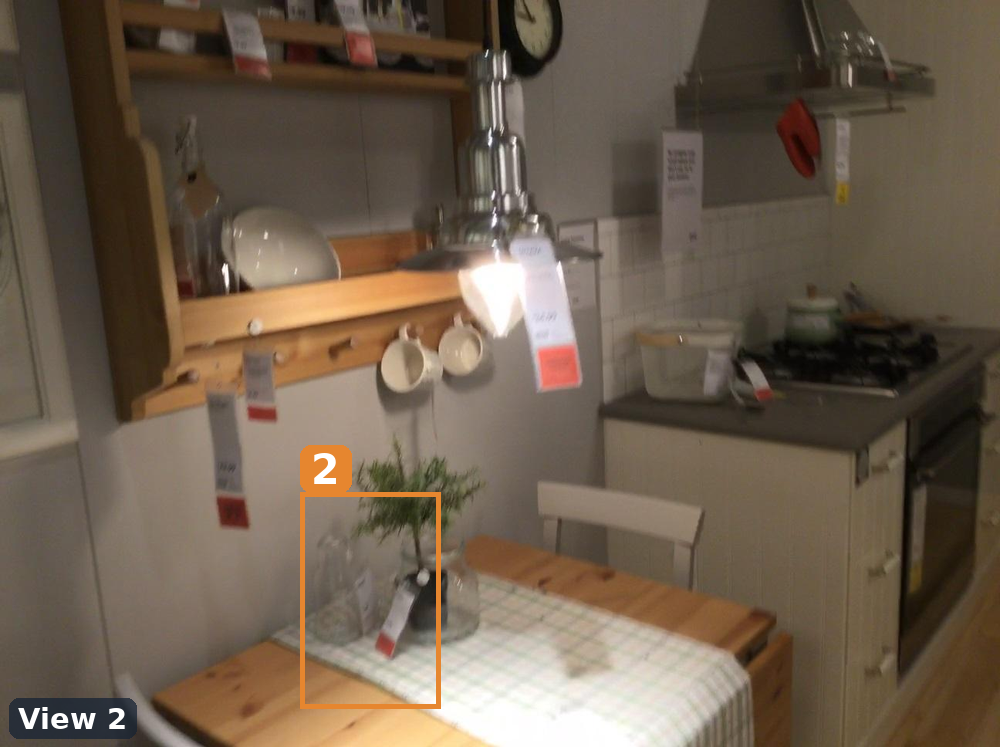}\par}
\casesection{Prompt}
\begin{promptbox}
The two pictures are two camera views of the same room, taken from different spots but sharing some overlapping content. Each object below shows up in only ONE of the pictures and is marked by a bounding box [x0, y0, x1, y1] (normalized 0-1000; origin top-left, x to the right, y downward).\newline - Object 1 = the oven, box [123, 564, 265, 733] in the FIRST picture.\newline - Object 2 = the plant, box [300, 659, 440, 948] in the SECOND picture.\newline Put yourself in the shoes of whoever took the FIRST picture, facing the way that picture points. Relative to Object 2, which way is Object 1? (front = further from you into the scene, back = closer to you, left / right = your own left / right.)

\caseoptions{{\setlength{\tabcolsep}{0pt}\renewcommand{\arraystretch}{1.15}\begin{tabular}{@{}*{4}{p{0.250\linewidth}}@{}}\caseright{\textbf{A.}\,front} & \textbf{B.}\,front-right & \textbf{C.}\,right & \textbf{D.}\,back-right \\ \textbf{E.}\,back & \textbf{F.}\,back-left & \textbf{G.}\,left & \textbf{H.}\,front-left\end{tabular}}}

Reply with a single letter.
\end{promptbox}
\casesection{Reference answer}
\caseanswer{A. front}
\end{casecard}

\begin{casecard}{06}{Object-anchored relation}{Spatial relations}
{\centering \includegraphics[height=88pt,width=0.493\linewidth,keepaspectratio]{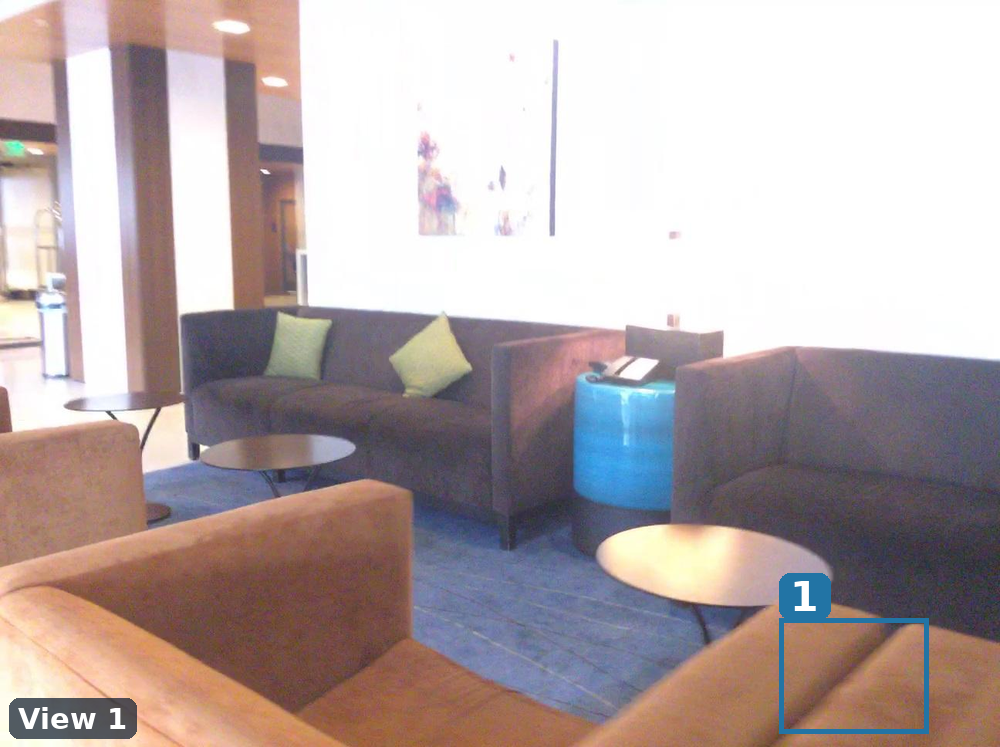}\hspace{6pt}\includegraphics[height=88pt,width=0.493\linewidth,keepaspectratio]{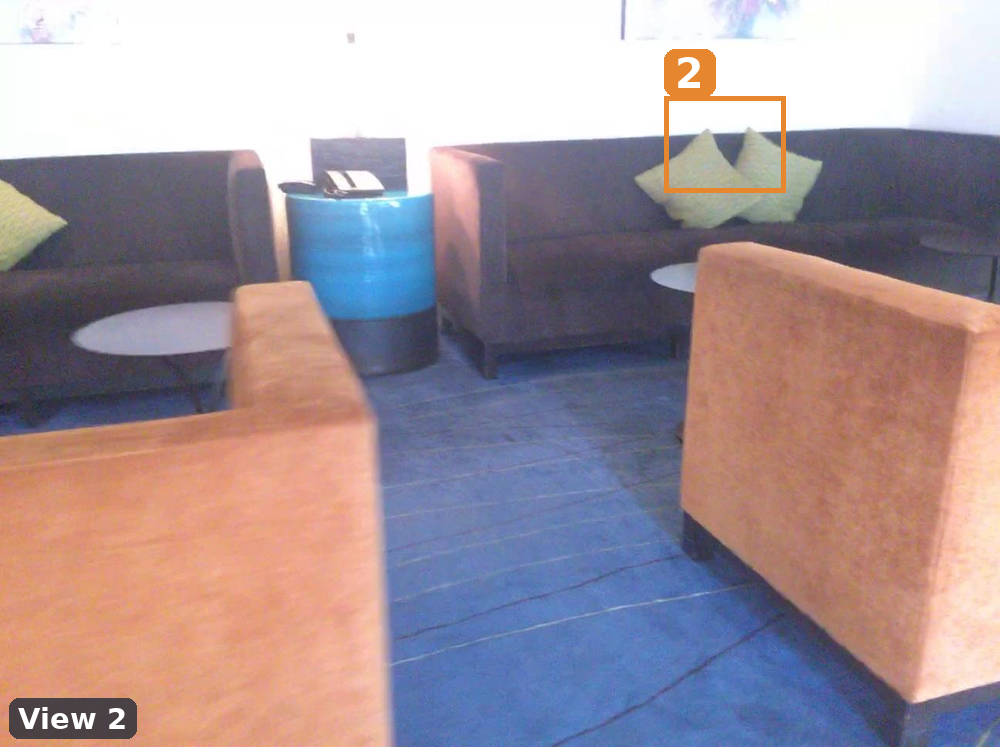}\par}
\casesection{Prompt}
\begin{promptbox}
These two images show the same room from different angles.\newline Object 1 (Image 1): couch, box [779, 828, 928, 982]\newline Object 2 (Image 2): cushion, box [664, 129, 785, 258]\newline Boxes are [x1, y1, x2, y2], normalized to 0-1000.\newline From the perspective of Object 1 (facing its own front), which relative direction is Object 2?

\caseoptions{{\setlength{\tabcolsep}{0pt}\renewcommand{\arraystretch}{1.15}\begin{tabular}{@{}*{4}{p{0.250\linewidth}}@{}}\caseright{\textbf{A.}\,front} & \textbf{B.}\,front-right & \textbf{C.}\,right & \textbf{D.}\,back-right \\ \textbf{E.}\,back & \textbf{F.}\,back-left & \textbf{G.}\,left & \textbf{H.}\,front-left\end{tabular}}}

Reply with a single letter.
\end{promptbox}
\casesection{Reference answer}
\caseanswer{A. front}
\end{casecard}

\clearpage

\begin{casecard}{07}{Distance ranking}{Spatial relations}
{\centering \includegraphics[height=88pt,width=0.493\linewidth,keepaspectratio]{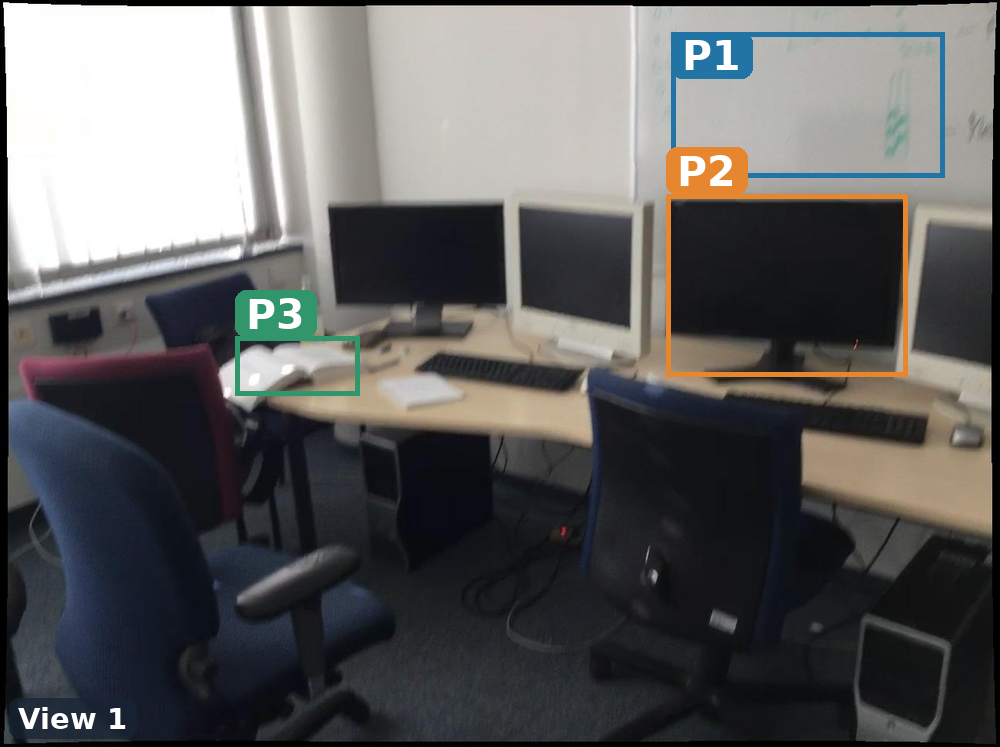}\hspace{6pt}\includegraphics[height=88pt,width=0.493\linewidth,keepaspectratio]{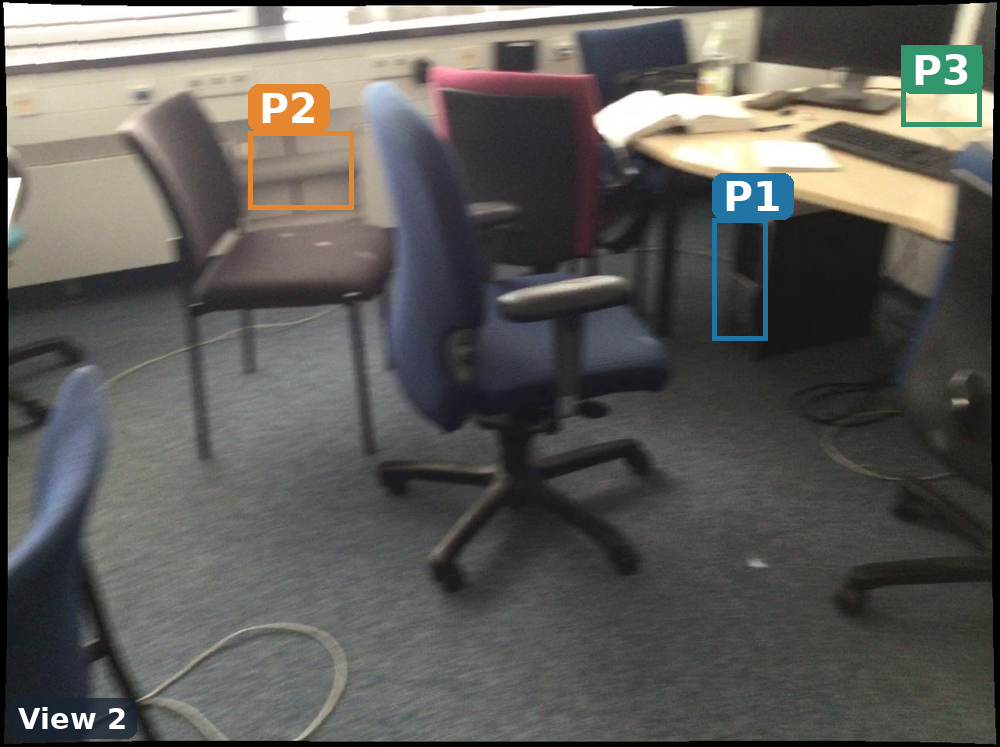}\par}
\casesection{Prompt}
\begin{promptbox}
These two images are different viewpoints of the same room. Boxes are [x0, y0, x1, y1] in pixel coordinates normalized to 0-1000 (origin at the top-left, x rightward, y downward). Consider these three object pairs:\newline - Pair 1: the blackboard in image 1 (at box [671, 43, 944, 238]) and the computer in image 2 (at box [712, 293, 767, 456])\newline - Pair 2: the monitor in image 1 (at box [666, 260, 907, 504]) and the pipe in image 2 (at box [248, 176, 353, 281])\newline - Pair 3: the book in image 1 (at box [235, 451, 359, 530]) and the monitor in image 2 (at box [901, 61, 981, 169])\newline Rank the pairs from CLOSEST to FARTHEST by the real-world 3D distance between the two objects in each pair (each option lists pair numbers closest-first).

\caseoptions{{\setlength{\tabcolsep}{0pt}\renewcommand{\arraystretch}{1.15}\begin{tabular}{@{}*{4}{p{0.250\linewidth}}@{}}\textbf{A.}\,(2, 3, 1) & \textbf{B.}\,(1, 2, 3) & \caseright{\textbf{C.}\,(3, 1, 2)} & \textbf{D.}\,(3, 2, 1) \\ \textbf{E.}\,(2, 1, 3) & \textbf{F.}\,(1, 3, 2) &  & \end{tabular}}}

Answer with the option letter only.
\end{promptbox}
\casesection{Reference answer}
\caseanswer{C. (3, 1, 2)}
\end{casecard}

\begin{casecard}{08}{Nearest object}{Spatial relations}
{\centering \includegraphics[height=80pt,width=0.326\linewidth,keepaspectratio]{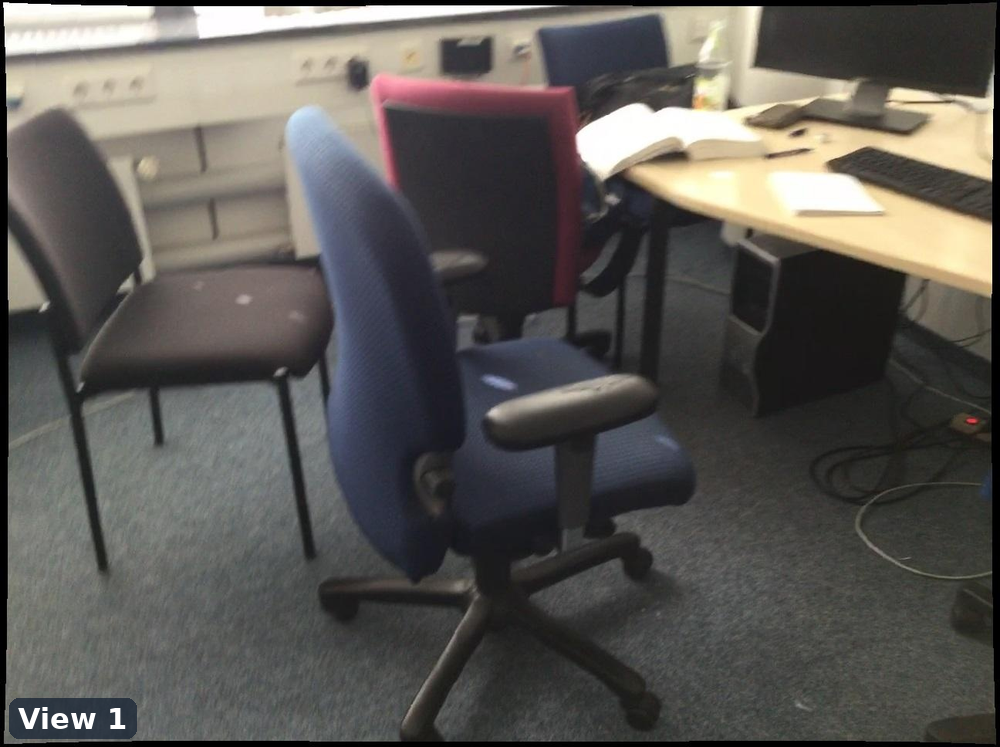}\hspace{6pt}\includegraphics[height=80pt,width=0.326\linewidth,keepaspectratio]{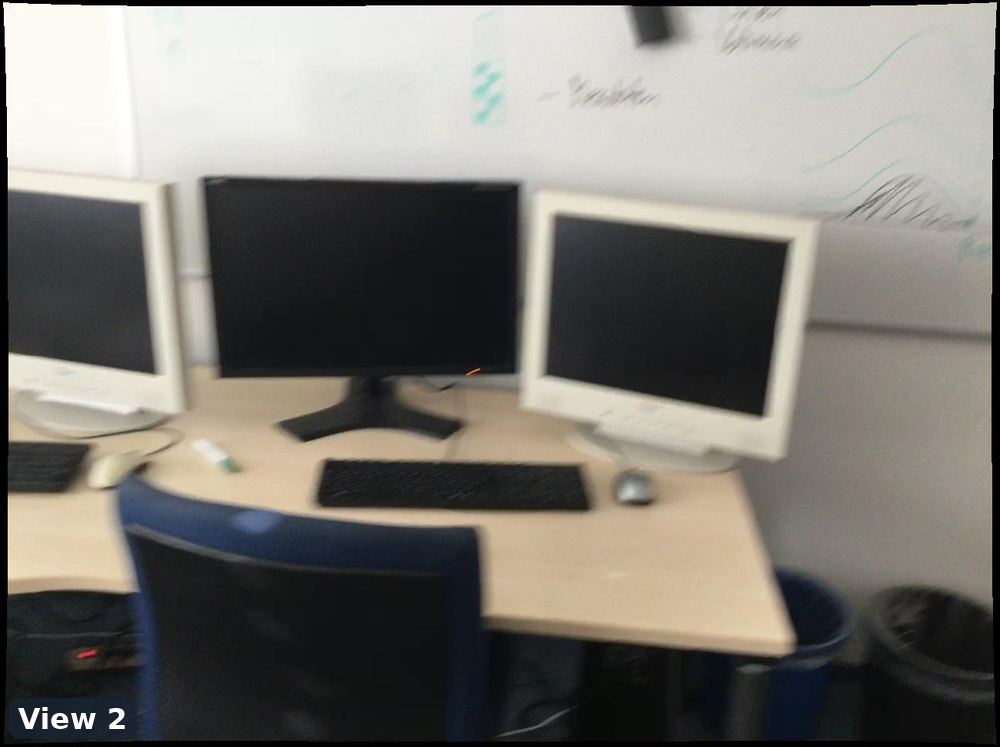}\hspace{6pt}\includegraphics[height=80pt,width=0.326\linewidth,keepaspectratio]{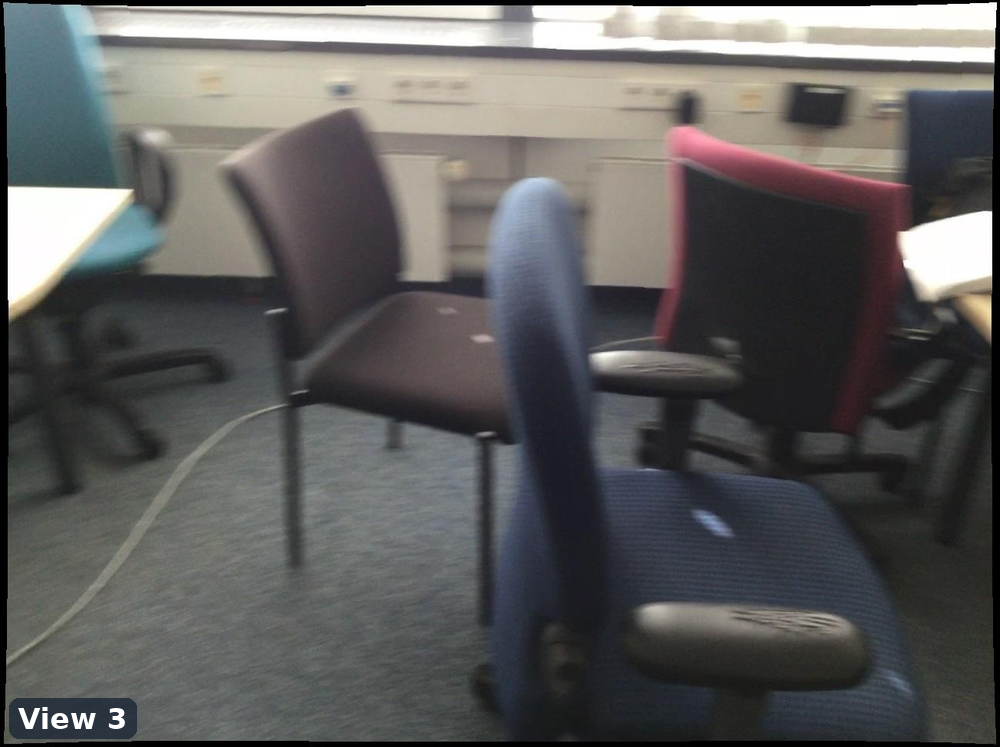}\par}
\casesection{Prompt}
\begin{promptbox}
These 3 images are different viewpoints of the same room. Boxes are [x0, y0, x1, y1] in pixel coordinates normalized to 0-1000 (origin at the top-left, x rightward, y downward).\newline Reference object: the monitor in image 1 (at box [778, 106, 962, 181]).\newline Candidates:\newline - A. the desk in image 3 (at box [46, 652, 132, 886])\newline - B. the blackboard in image 2 (at box [235, 77, 917, 419])\newline - C. the chair in image 3 (at box [62, 126, 233, 614])\newline - D. the bin in image 2 (at box [847, 772, 972, 963])\newline Which candidate is CLOSEST to the monitor in the real 3D room? Each object appears in only one of the images, so you must relate the viewpoints to each other. Answer with the option letter only.
\end{promptbox}
\casesection{Reference answer}
\caseanswer{B. the blackboard in image 2}
\end{casecard}

\clearpage

\begin{casecard}{09}{Object correspondence: matching}{Spatial relations}
{\centering \includegraphics[height=88pt,width=0.493\linewidth,keepaspectratio]{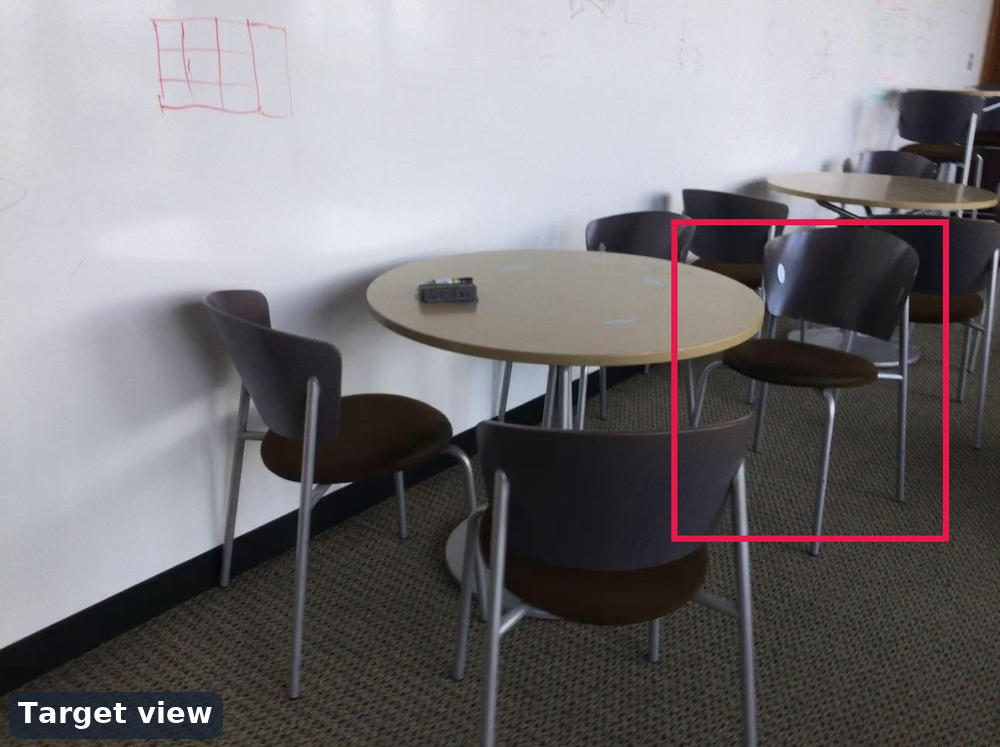}\hspace{6pt}\includegraphics[height=88pt,width=0.493\linewidth,keepaspectratio]{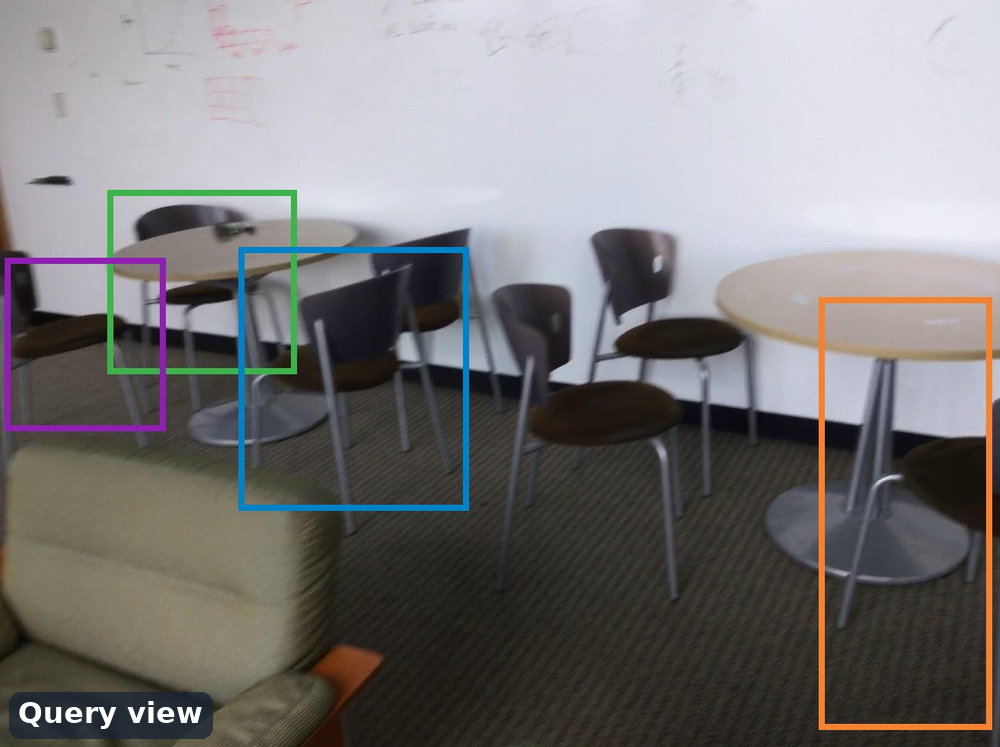}\par}
\casesection{Prompt}
\begin{promptbox}
The first image marks one target chair with a red box. In the last image, several chairs are each marked with a colored box. Which colored chair is the same object as the red-boxed target?

\caseoptions{{\setlength{\tabcolsep}{0pt}\renewcommand{\arraystretch}{1.15}\begin{tabular}{@{}*{4}{p{0.250\linewidth}}@{}}\textbf{A.}\,green & \textbf{B.}\,orange & \textbf{C.}\,purple & \textbf{D.}\,Not visible \\ \caseright{\textbf{E.}\,blue} &  &  & \end{tabular}}}

Please answer directly with only the letter of the correct option and nothing else.
\end{promptbox}
\casesection{Reference answer}
\caseanswer{E. blue}
\end{casecard}

\begin{casecard}{10}{Object correspondence: missing instance}{Spatial relations}
{\centering \includegraphics[height=128pt,width=0.326\linewidth,keepaspectratio]{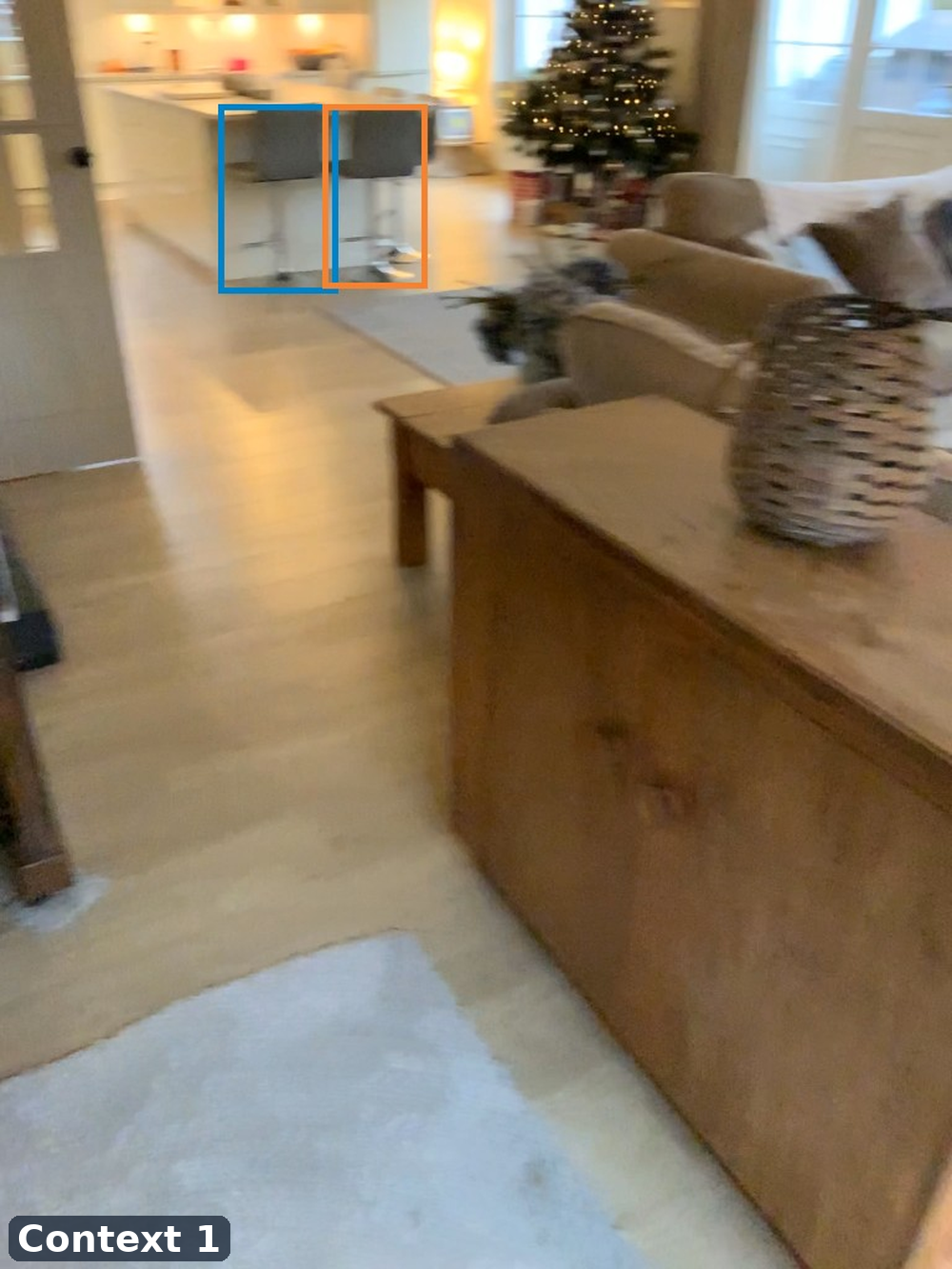}\hspace{6pt}\includegraphics[height=128pt,width=0.326\linewidth,keepaspectratio]{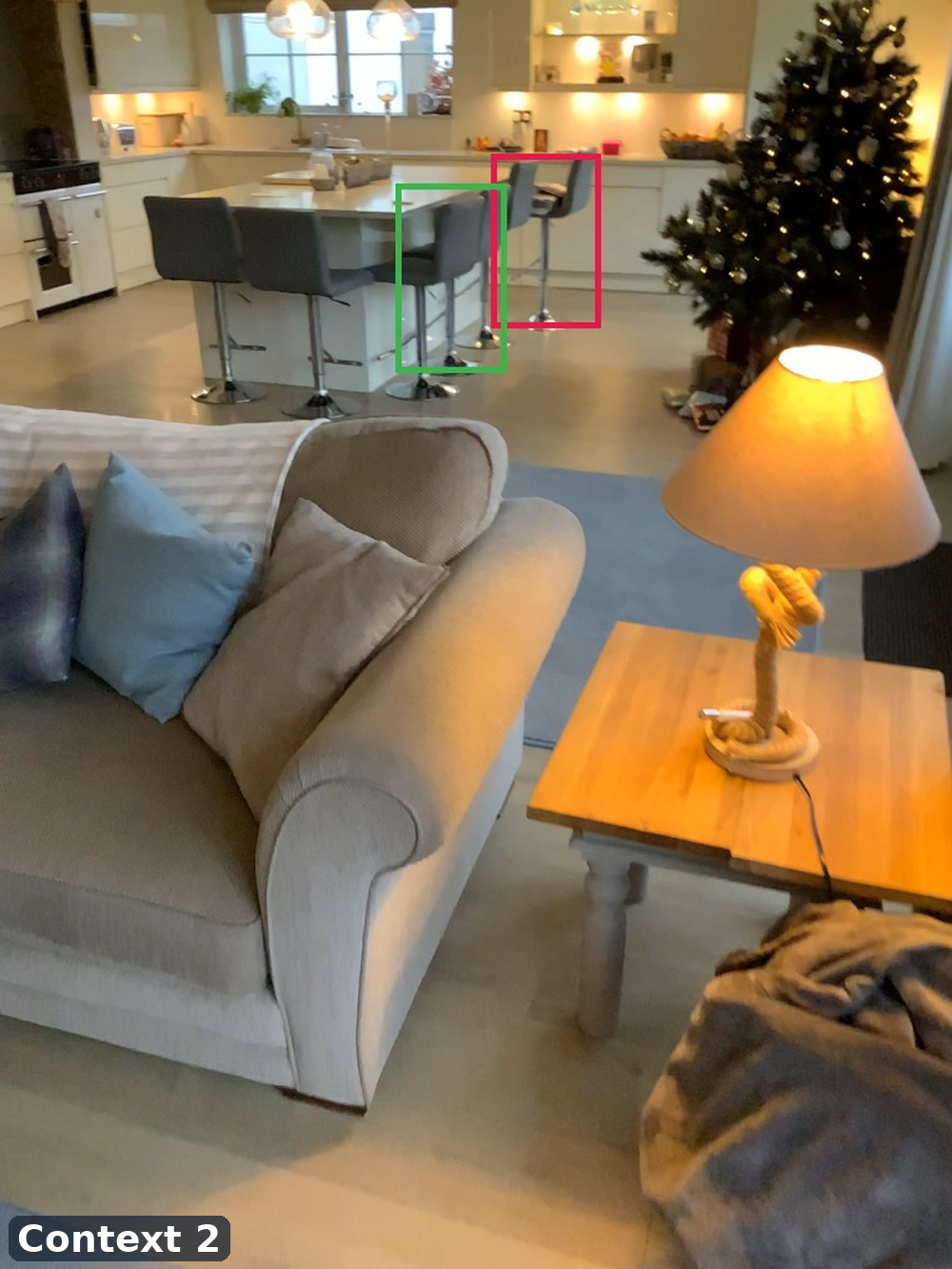}\hspace{6pt}\includegraphics[height=128pt,width=0.326\linewidth,keepaspectratio]{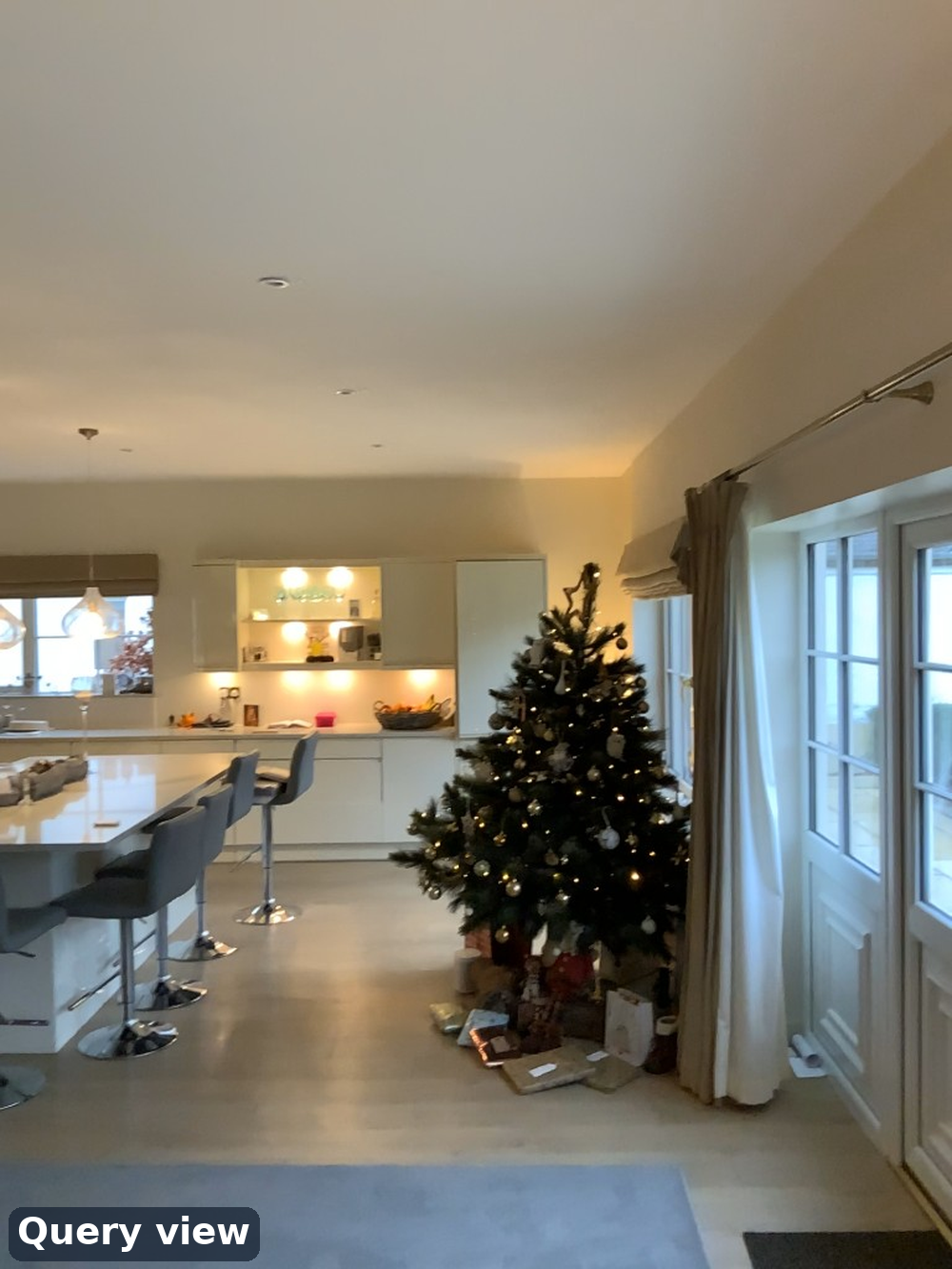}\par}
\casesection{Prompt}
\begin{promptbox}
The first 2 image(s) mark 4 different bar stools, each with a colored box. The last image shows the same room from another viewpoint. Which colored bar stool is missing from the last image?

\caseoptions{{\setlength{\tabcolsep}{0pt}\renewcommand{\arraystretch}{1.15}\begin{tabular}{@{}*{4}{p{0.250\linewidth}}@{}}\textbf{A.}\,red & \textbf{B.}\,green & \caseright{\textbf{C.}\,blue} & \textbf{D.}\,orange\end{tabular}}}

Please answer directly with only the letter of the correct option and nothing else.
\end{promptbox}
\casesection{Reference answer}
\caseanswer{C. blue}
\end{casecard}

\clearpage
\definecolor{reconobj1}{HTML}{2274A5}
\definecolor{reconobj2}{HTML}{E6862F}
\definecolor{reconobj3}{HTML}{31966B}
\definecolor{reconobj4}{HTML}{8656A4}
\definecolor{reconobj5}{HTML}{D44F56}
\begin{casecard}{11}{Joint scene reconstruction}{Scene reconstruction}
{\centering \includegraphics[height=120pt,width=0.493\linewidth,keepaspectratio]{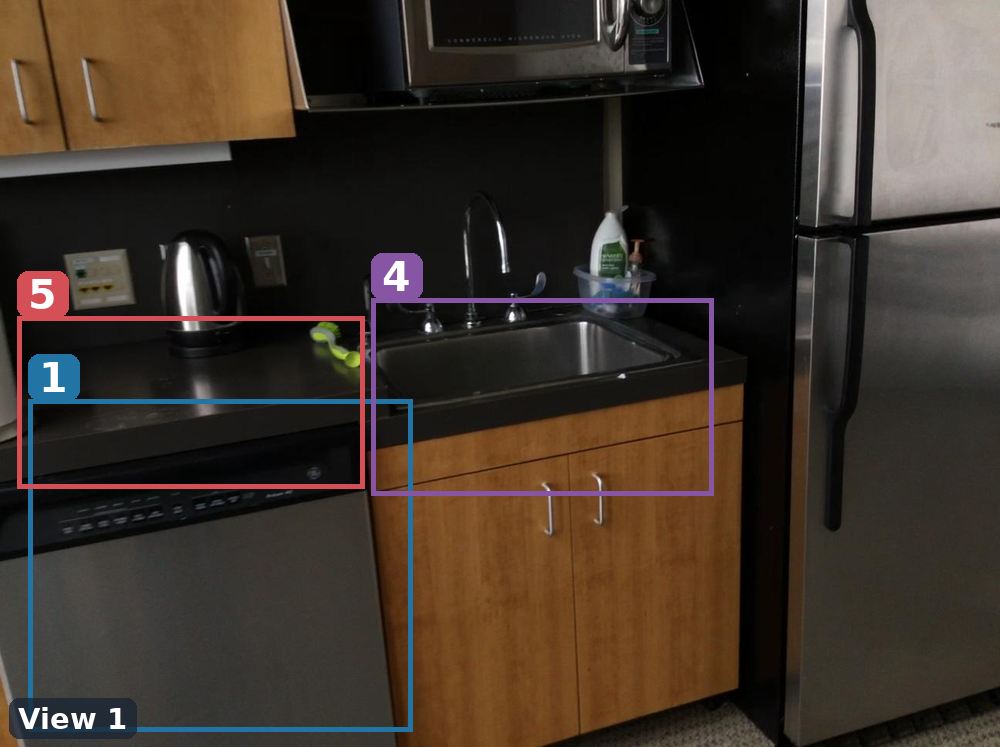}\hspace{6pt}\includegraphics[height=120pt,width=0.493\linewidth,keepaspectratio]{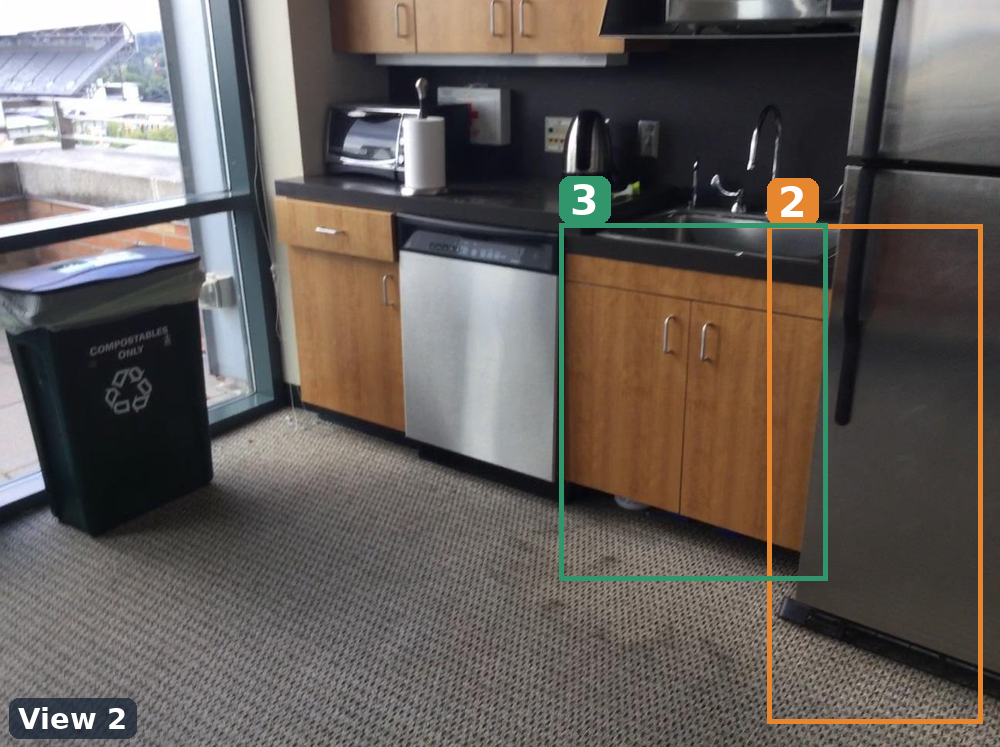}\par}
\casesection{Prompt}
\begin{promptbox}
{\casemono
You\ are\ given\ two\ photographs\ of\ the\ same\ indoor\ room,\ taken\ from\ two\ nearby\\
viewpoints\ with\ overlapping\ fields\ of\ view.\ Reconstruct\ the\ 3D\ layout\ of\ the\ listed\\
objects\ as\ a\ structured\ scene.\\[3pt]
COORDINATE\ SYSTEM\ (right-handed,\ three.js/glTF,\ units\ =\ METERS):\\
-\ +Y\ is\ up\ (against\ gravity);\ y\ =\ 0\ is\ the\ FLOOR;\ an\ object\char13{}s\ y\ is\ the\ height\ of\ its\\
\ \ box\ CENTER\ above\ the\ floor.\\
-\ The\ horizontal\ origin\ (x=0,\ z=0)\ is\ directly\ below\ the\ camera\ of\ IMAGE\ 1.\\
-\ -Z\ is\ the\ direction\ IMAGE-1\char13{}s\ camera\ looks\ (forward);\ +X\ is\ its\ right.\\[3pt]
OBJECTS\ TO\ PLACE\ (each\ is\ marked\ by\ a\ 2D\ box\ in\ ONE\ image;\ box\ is\ normalized\ 0-1000,\\
origin\ top-left,\ x-\char62{}right,\ y-\char62{}down).\ Place\ every\ listed\ object:\\
\reconobj{1}id\ \char34{}dishwasher\char95{}1\char34{}\ \char124{}\ dishwasher\ \char124{}\ IMAGE\ 1\ \char124{}\ box\ [28,\ 535,\ 412,\ 979]\\
\reconobj{2}id\ \char34{}refrigerator\char95{}1\char34{}\ \char124{}\ refrigerator\ \char124{}\ IMAGE\ 2\ \char124{}\ box\ [767,\ 301,\ 982,\ 969]\\
\reconobj{3}id\ \char34{}cabinet\char95{}1\char34{}\ \char124{}\ cabinet\ \char124{}\ IMAGE\ 2\ \char124{}\ box\ [559,\ 299,\ 827,\ 777]\\
\reconobj{4}id\ \char34{}sink\char95{}1\char34{}\ \char124{}\ sink\ \char124{}\ IMAGE\ 1\ \char124{}\ box\ [371,\ 400,\ 713,\ 663]\\
\reconobj{5}id\ \char34{}counter\char95{}1\char34{}\ \char124{}\ counter\ \char124{}\ IMAGE\ 1\ \char124{}\ box\ [17,\ 424,\ 364,\ 654]\\[3pt]
FOR\ EACH\ listed\ object\ output\ its\ 3D\ oriented\ bounding\ box:\\
-\ center\ [x,\ y,\ z]\ \ (meters;\ y\ =\ box-center\ height\ above\ the\ floor)\\
-\ size\ \ \ [w,\ h,\ d]\ \ (meters,\ in\ the\ object\char13{}s\ OWN\ axes:\ w\ =\ width\ along\ its\ local\ X,\\
\ \ \ \ \ \ \ \ \ \ \ \ \ \ \ \ \ \ \ \ \ h\ =\ height\ along\ Y/up,\ d\ =\ depth\ along\ its\ local\ Z/front-back)\\
-\ yaw\char95{}deg\ \ \ \ \ \ \ \ \ \ \ (rotation\ about\ +Y;\ 0\ =\ object\char13{}s\ front\ (+local\ Z)\ faces\ -Z;\\
\ \ \ \ \ \ \ \ \ \ \ \ \ \ \ \ \ \ \ \ \ positive\ =\ counter-clockwise\ seen\ from\ above)\\[3pt]
GUIDANCE:\\
-\ Use\ the\ two\ OVERLAPPING\ views\ to\ triangulate\ depth\ and\ relative\ position.\\
-\ Estimate\ REAL\ metric\ sizes\ from\ commonsense\ (a\ dining\ chair\ seat\ \char126{}0.45\ m,\ an\ interior\\
\ \ door\ \char126{}2\ m).\ The\ whole\ scene\ must\ be\ self-consistent\ and\ at\ real\ scale.\\
-\ Put\ objects\ at\ plausible\ heights\ (on\ the\ floor,\ on\ a\ table,\ wall-mounted...).\\[3pt]
You\ may\ reason\ first.\ Then\ output\ ONE\ JSON\ object\ and\ NOTHING\ after\ it,\ using\ the\ exact\\
ids\ given\ above:\\
\char18{}\char18{}\char18{}json\\
\char123{}\char34{}objects\char34{}:[\char123{}\char34{}id\char34{}:\char34{}\char60{}id\char62{}\char34{},\char34{}category\char34{}:\char34{}\char60{}cat\char62{}\char34{},\char34{}center\char34{}:[x,y,z],\char34{}size\char34{}:[w,h,d],\char34{}yaw\char95{}deg\char34{}:a\char125{}]\char125{}\\
\char18{}\char18{}\char18{}\par}
\end{promptbox}
\casesection{Reference answer}
{\footnotesize\setlength{\tabcolsep}{4pt}\begin{tabular*}{\linewidth}{@{\extracolsep{\fill}}llll@{}}\toprule
{\sffamily\scriptsize\bfseries Object} & {\casemono center} {\sffamily\scriptsize(m)} & {\casemono size [w, h, d]} {\sffamily\scriptsize(m)} & {\casemono yaw\char95{}deg} \\ \midrule
\reconobj{1}{\casemono dishwasher\char95{}1} & {\casemono [-0.502, 0.448, -1.374]} & {\casemono [0.680, 0.892, 0.114]} & {\casemono -158.490} \\
\reconobj{2}{\casemono refrigerator\char95{}1} & {\casemono [0.869, 0.841, -1.677]} & {\casemono [0.809, 1.661, 0.281]} & {\casemono -158.490} \\
\reconobj{3}{\casemono cabinet\char95{}1} & {\casemono [0.159, 0.435, -1.603]} & {\casemono [0.810, 0.869, 0.245]} & {\casemono -158.490} \\
\reconobj{4}{\casemono sink\char95{}1} & {\casemono [0.067, 0.789, -1.829]} & {\casemono [0.729, 0.243, 0.573]} & {\casemono -158.490} \\
\reconobj{5}{\casemono counter\char95{}1} & {\casemono [-0.845, 0.868, -1.501]} & {\casemono [1.234, 0.132, 0.580]} & {\casemono -158.490} \\
\bottomrule\end{tabular*}}
\end{casecard}

\clearpage

\begin{casecard}{12}{Offline navigation}{Navigation}
\noindent\includegraphics[width=0.194\linewidth]{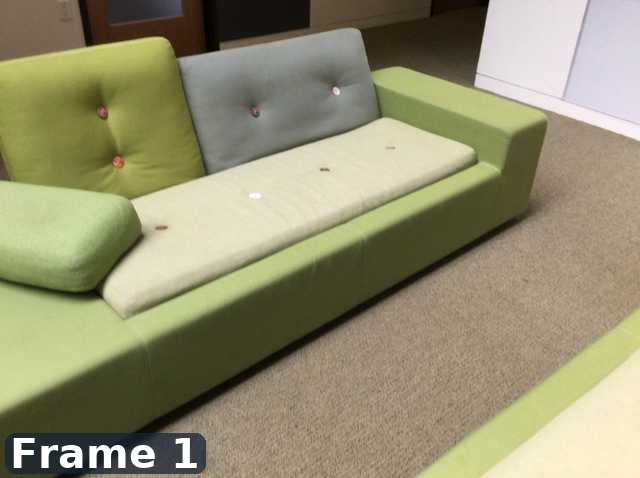}\hfill\includegraphics[width=0.194\linewidth]{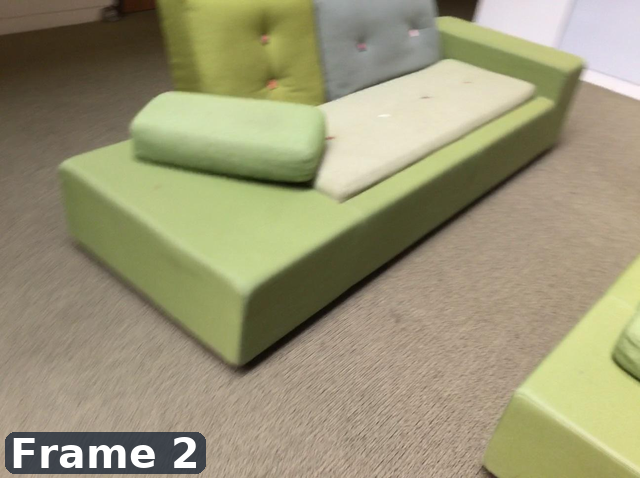}\hfill\includegraphics[width=0.194\linewidth]{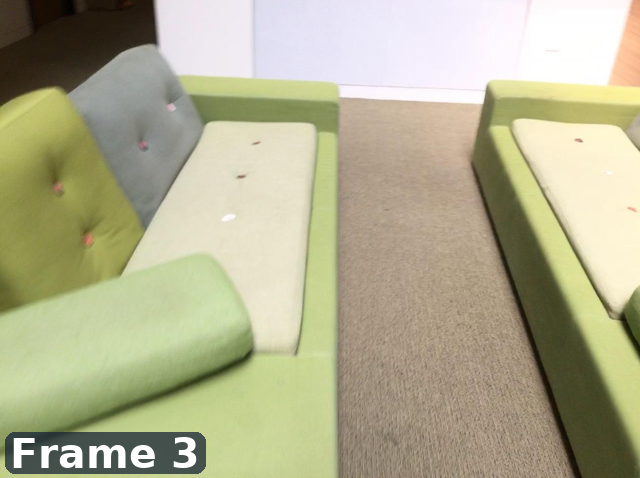}\hfill\includegraphics[width=0.194\linewidth]{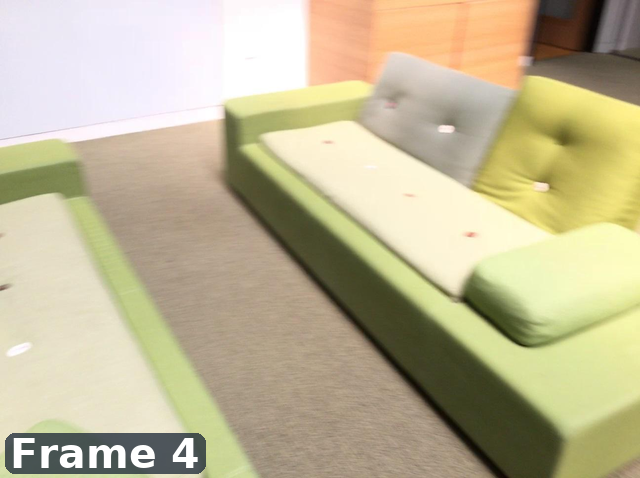}\hfill\includegraphics[width=0.194\linewidth]{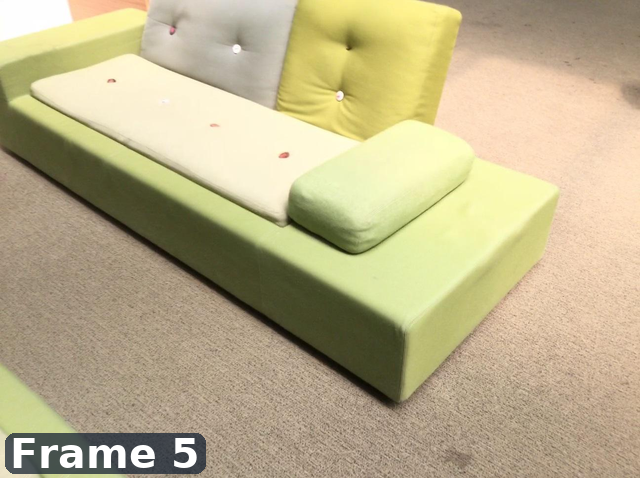}\par\vspace{2pt}
\noindent\includegraphics[width=0.194\linewidth]{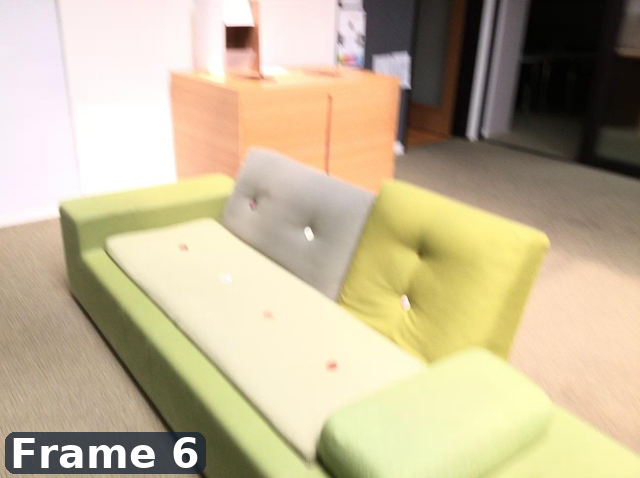}\hfill\includegraphics[width=0.194\linewidth]{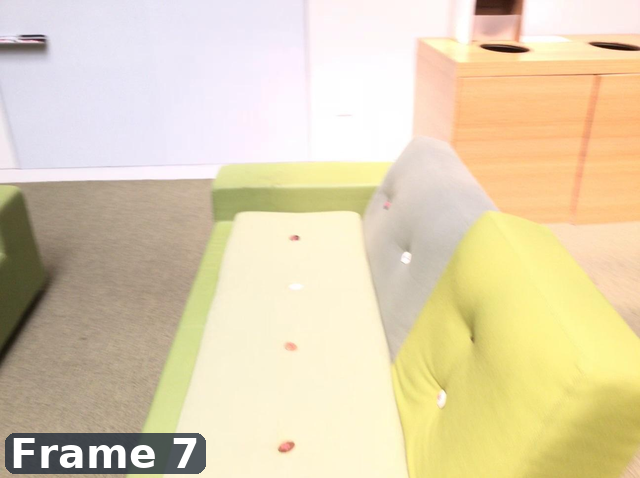}\hfill\includegraphics[width=0.194\linewidth]{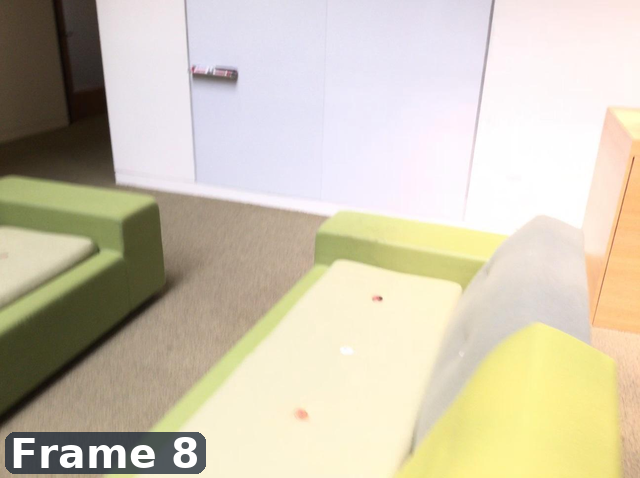}\hfill\includegraphics[width=0.194\linewidth]{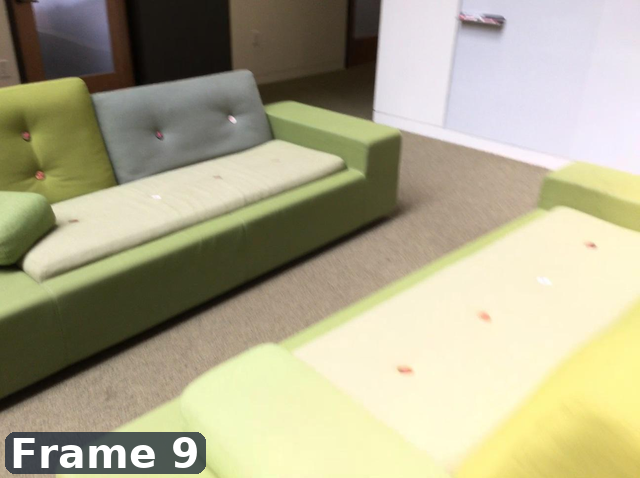}\hfill\includegraphics[width=0.194\linewidth]{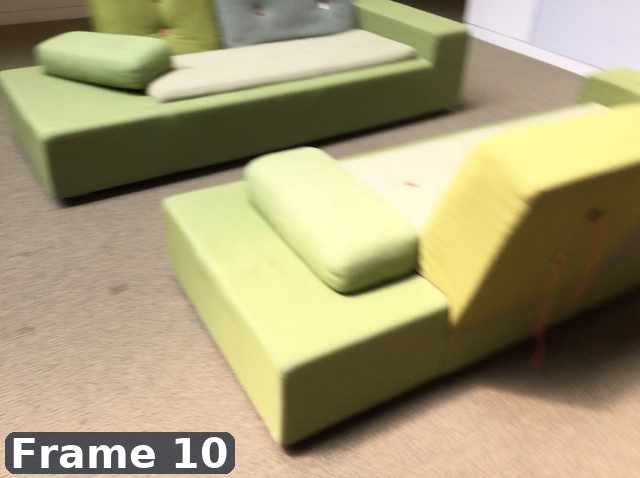}\par\vspace{2pt}
\noindent\includegraphics[width=0.194\linewidth]{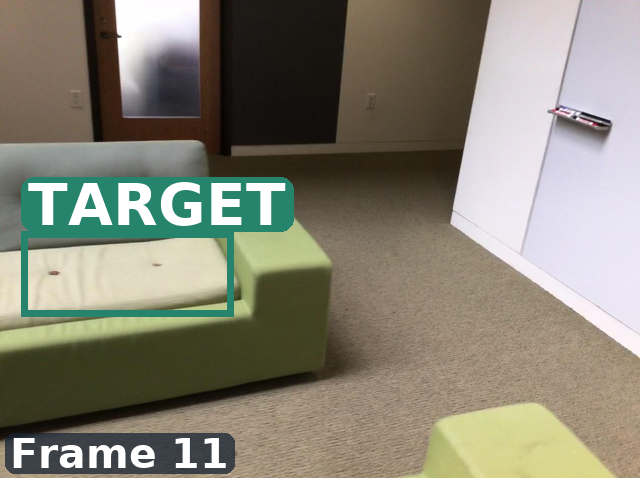}\hfill\includegraphics[width=0.194\linewidth]{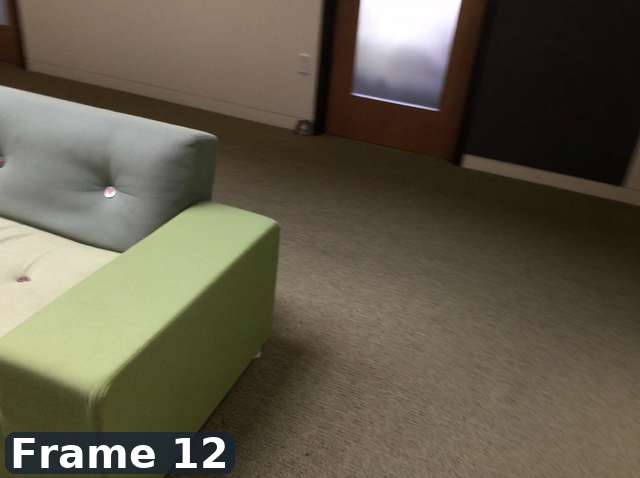}\hfill\includegraphics[width=0.194\linewidth]{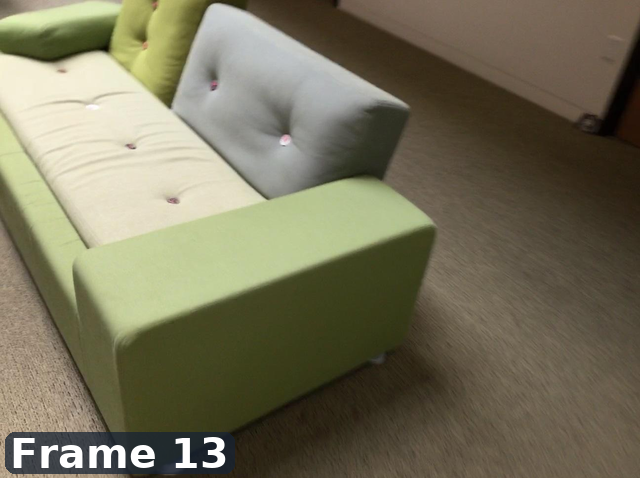}\hfill\includegraphics[width=0.194\linewidth]{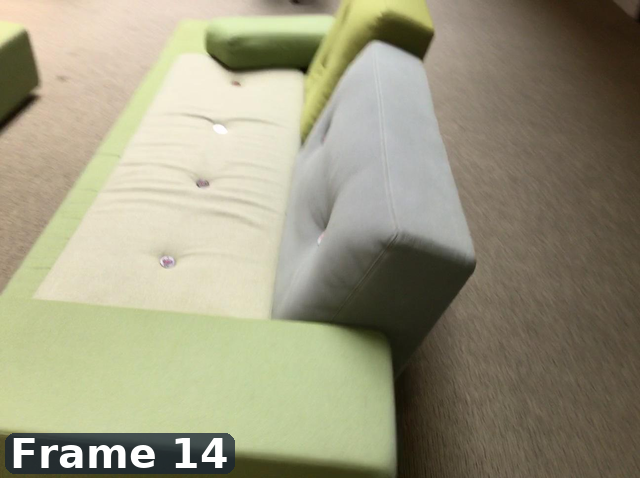}\hfill\includegraphics[width=0.194\linewidth]{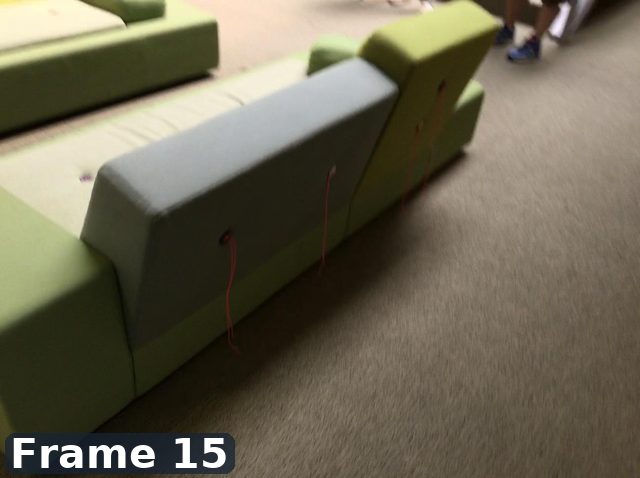}\par\vspace{2pt}
\noindent\includegraphics[width=0.194\linewidth]{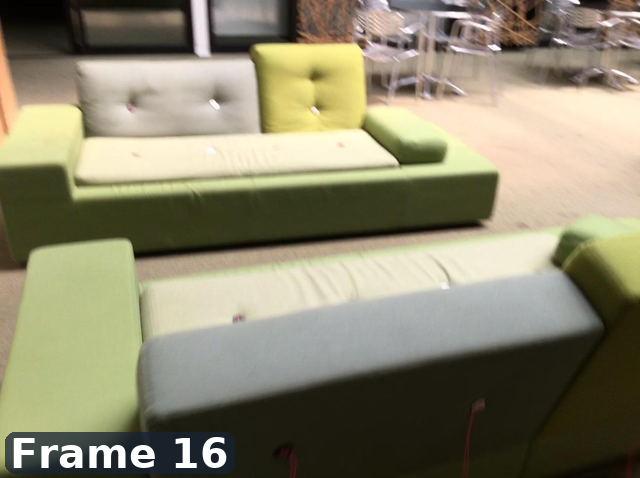}\hfill\includegraphics[width=0.194\linewidth]{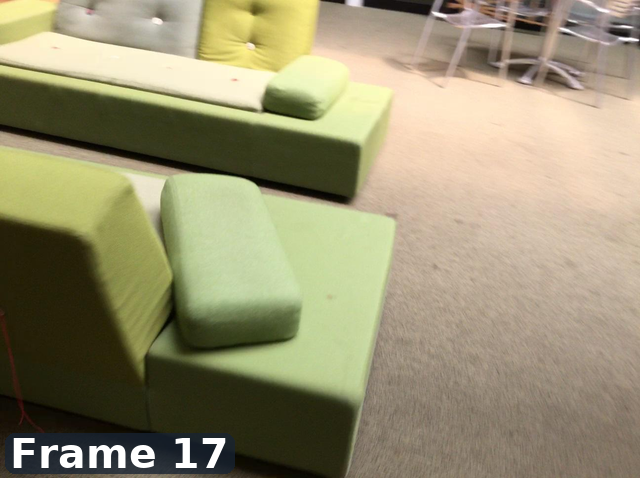}\hfill\includegraphics[width=0.194\linewidth]{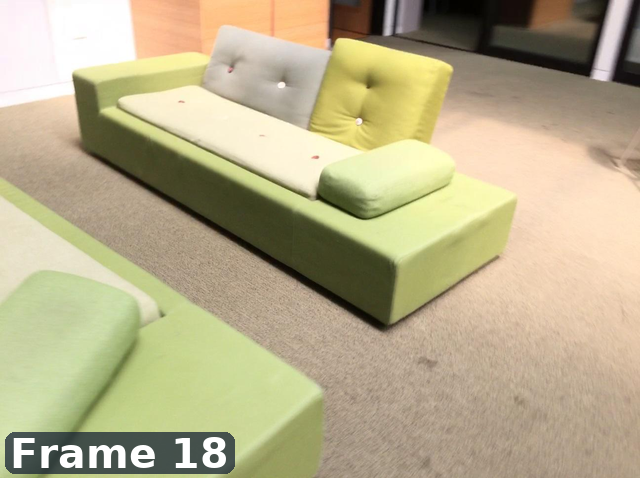}\hfill\includegraphics[width=0.194\linewidth]{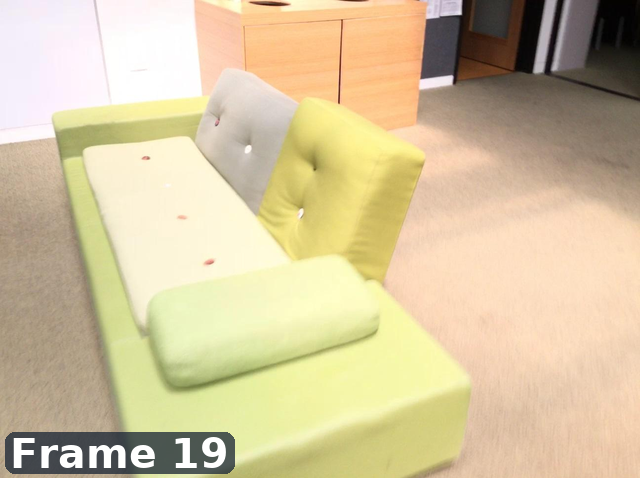}\hfill\hspace*{0.194\linewidth}\par\vspace{2pt}
\casesection{Prompt}
\begin{promptbox}
You are given 19 discrete first-person frames (numbered 1-19) of an indoor tour.

Starting Position: You are at the viewpoint of Frame \#19, facing the couch centered in the image.

Coordinate System: Origin (0.00, 0.00) is your current position. +y is straight ahead, +x is to your right (in meters). Each waypoint is a coordinate (x, y) in this local frame.\newline Navigation Goal: Walk to the mat shown in Frame \#11 (bounding box [33, 484, 365, 662], normalized pixels 0-1000).

No numeric goal coordinate is provided \textemdash{} estimate the distances yourself from the tour images. Report the path in meters as best you can (only its overall direction and shape are scored).\newline Task: Plan a collision-free path to the goal, routing around furniture and obstacles rather than heading straight for it (a direct line is often blocked).

Output Format: Output ONLY the planned path XML tag. Use the prefix 0 (a fixed track id for you, the mover) before each waypoint:\newline {\casemono \char60{}bev\char95{}task\char95{}pred\ coords=\char34{}0\ x1\ y1;\ 0\ x2\ y2;\ ...\ 0\ xn\ yn\char34{}\char62{}\char60{}/bev\char95{}task\char95{}pred\char62{}}\newline (Each x, y in meters to 2 decimal places. The first waypoint should be near 0.00 0.00, and the last should be close to the goal.)
\end{promptbox}
\casesection{Reference answer}
\caseanswercode{\char60{}bev\char95{}task\char95{}pred\ coords=\char34{}0\ 0.00\ 0.00;\ 0\ -0.38\ 0.68;\ 0\ -1.06\ 2.77;\ 0\ -2.10\ 2.70\char34{}\char62{}\char60{}/bev\char95{}task\char95{}pred\char62{}}
\end{casecard}

\clearpage

\newsavebox{\fitwidthbox}
\newcommand{\fitwidth}[1]{\sbox{\fitwidthbox}{#1}%
  \ifdim\wd\fitwidthbox>\linewidth\resizebox{\linewidth}{!}{\usebox{\fitwidthbox}}%
  \else\usebox{\fitwidthbox}\fi}

\section{Additional Analyses}
\label{app:analyses}

This appendix details the analyses in Section~\ref{sec:evaluation},
in the order they appear in the main text.

\subsection{Per-domain results}
\label{app:domain_results}

Table~\ref{tab:domain_results_full} decomposes reconstruction by domain.
Structure exceeds Metric for every model in every domain, but the size of the
gap depends on the domain. It is largest on tabletop scenes (20.6--58.2
points), where Structure is highest but Metric often collapses: with small
objects, a modest scale error becomes a large relative error. Indoor scenes
yield the lowest Structure for all but one model, and outdoor scenes the
highest Metric for most models. Rankings also shift across domains; for
example, Gemini-3.5-Flash ranks fifth on tabletop Metric despite a low
overall reconstruction score. ScanNet and CA-1M are pooled into one indoor
average.

\begin{table}[t]
  \centering
  \small
  \caption{Structure (S) and Metric (M) reconstruction scores across indoor,
  tabletop, and outdoor scenes. Scores are percentages.}
  \label{tab:domain_results_full}
  \begin{tabular}{lrrrrrr}
\toprule
Model & \multicolumn{2}{c}{Indoor} & \multicolumn{2}{c}{Tabletop} & \multicolumn{2}{c}{Outdoor} \\
\cmidrule(lr){2-3}\cmidrule(lr){4-5}\cmidrule(lr){6-7}
& S & M & S & M & S & M \\
\midrule
GPT-6-Astra & 72.5 & 50.9 & 87.4 & 55.8 & 84.1 & 69.4 \\
GPT-5.6-Sol & 56.8 & 34.9 & 82.2 & 41.1 & 74.8 & 48.4 \\
Claude-Fable-5 & 54.8 & 29.8 & 74.5 & 39.4 & 76.8 & 49.1 \\
Claude-Opus-5 & 53.9 & 28.4 & 76.8 & 38.3 & 73.7 & 50.0 \\
GLM-5.3-Flash & 56.5 & 32.6 & 74.6 & 34.8 & 73.9 & 44.3 \\
Qwen3.8-Max & 56.3 & 34.2 & 77.0 & 28.5 & 72.1 & 47.4 \\
Gemini-3.8-Flash & 55.0 & 33.5 & 74.7 & 34.8 & 69.4 & 44.6 \\
Kimi-K3 & 54.9 & 33.6 & 77.9 & 19.8 & 73.7 & 44.6 \\
Doubao-Seed-2.1-Pro & 54.9 & 33.8 & 76.4 & 24.6 & 59.9 & 32.3 \\
GPT-5.5 & 54.4 & 32.1 & 73.6 & 22.5 & 67.0 & 31.2 \\
Grok-4.6 & 50.5 & 30.4 & 71.6 & 19.2 & 67.9 & 38.6 \\
Gemini-3.1-Pro & 53.8 & 30.1 & 69.6 & 20.4 & 63.5 & 38.1 \\
Gemini-3.7-Flash & 45.1 & 28.9 & 58.2 & 28.6 & 57.7 & 43.3 \\
Gemini-3.5-Flash & 36.2 & 25.4 & 58.1 & 37.5 & 53.0 & 35.9 \\
Claude-Opus-4.8 & 40.8 & 19.2 & 60.0 & 16.9 & 65.6 & 36.7 \\
Claude-Opus-4.7 & 41.4 & 22.0 & 61.5 & 10.4 & 58.7 & 30.8 \\
Qwen3.7-Plus & 43.2 & 15.6 & 72.3 & 14.1 & 42.6 & 29.1 \\
MiniMax-M3 & 38.6 & 20.9 & 54.8 & 12.8 & 58.9 & 20.8 \\
\bottomrule
\end{tabular}

\end{table}

\subsection{Cross-level associations}
\label{app:cross_level}

\begin{figure}[t]
  \centering
  \includegraphics[width=\linewidth]{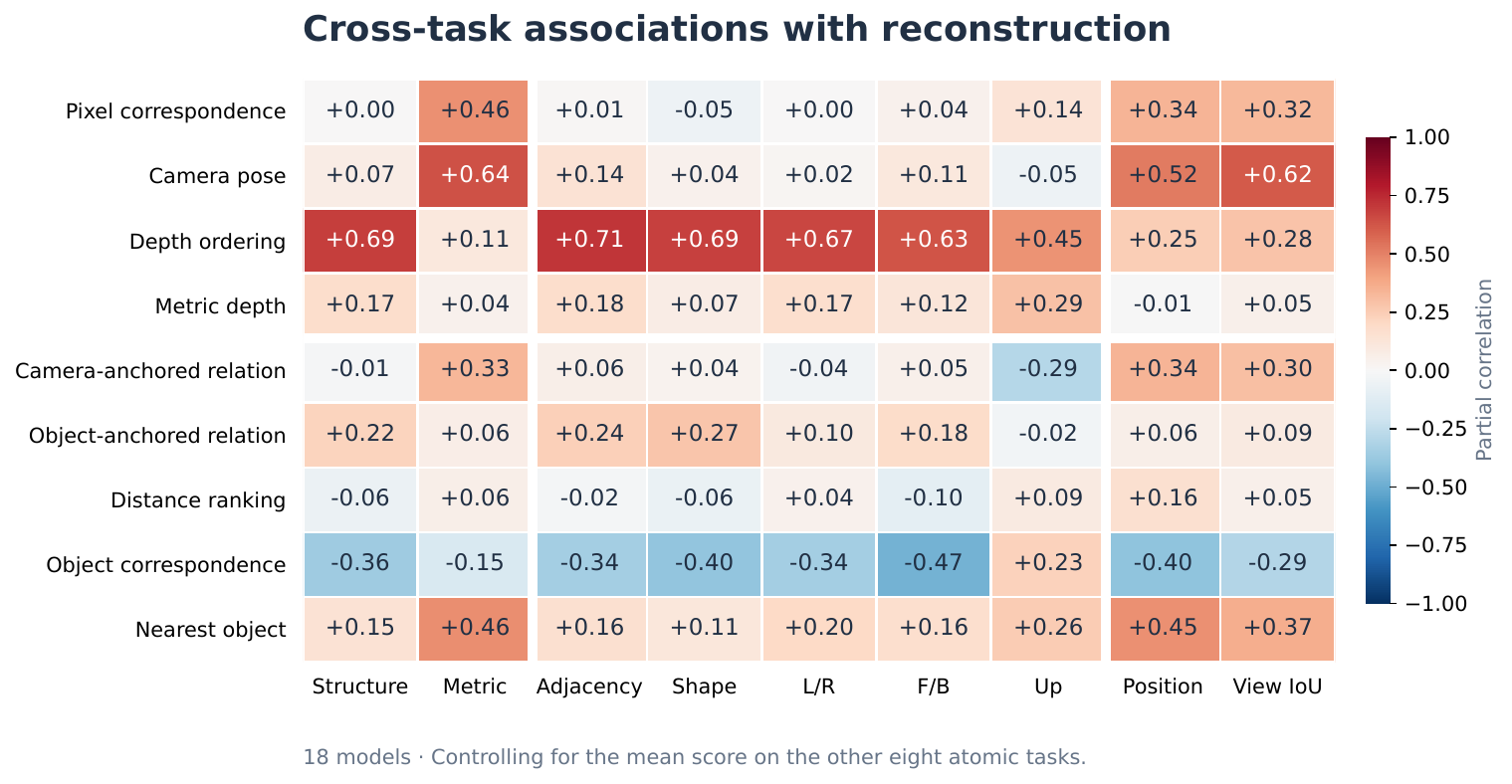}
  \caption{Cross-level partial correlations across eighteen models on the
  full reconstruction suite with three-domain aggregation. Each cell controls
  for the mean score on the other eight atomic tasks.}
  \label{fig:partial_correlation}
\end{figure}

Raw task-score correlations combine differences in overall model performance
with task-specific associations.
For each atomic score $x$ and reconstruction component $y$, let $z$ be the mean
of the other eight atomic scores across the same eighteen models. We regress
$x$ and $y$ separately on an intercept and $z$, then correlate the residuals:
\begin{equation}
  r_{\mathrm{partial}}=\operatorname{corr}
  \bigl(x-\widehat{x}(z),\;y-\widehat{y}(z)\bigr).
\end{equation}
Figure~\ref{fig:partial_correlation} reports the full matrix, and
Table~\ref{tab:cross_level} compares selected raw and partial correlations.
Raw correlations are uniformly high, but most fall sharply after adjustment,
indicating that they largely reflect overall model strength. Depth ordering
and front--back order remain the most stable pair (partial $r=0.63$, robust to
removing any single model), whereas atomic metric depth falls from $r=0.53$ to
$0.04$. These associations are descriptive rather than causal.

\begin{table}[t]
  \centering
  \small
  \caption{Selected cross-level associations over eighteen models. Partial
  correlation controls for the mean of the other eight atomic tasks; LOO
  gives the range after leaving out each model. Task abbreviations follow
  Table~\ref{tab:main_results}.}
  \label{tab:cross_level}
  \fitwidth{%
    \begin{tabular}{llrrrr}
\toprule
Atomic unit & Reconstruction axis & Pearson & Spearman & Partial & LOO range \\
\midrule
PC & metric & +0.86 & +0.81 & +0.46 & [+0.38, +0.53] \\
PC & posAP & +0.78 & +0.77 & +0.34 & [+0.24, +0.42] \\
CP & vIoU & +0.88 & +0.80 & +0.62 & [+0.29, +0.72] \\
DO & FB & +0.78 & +0.75 & +0.63 & [+0.51, +0.75] \\
MD & metric & +0.53 & +0.43 & +0.04 & [-0.10, +0.09] \\
Rel & LR & +0.63 & +0.58 & -0.04 & [-0.19, +0.07] \\
Rel & posAP & +0.79 & +0.86 & +0.34 & [+0.23, +0.49] \\
Dist & structure & +0.60 & +0.58 & -0.06 & [-0.18, +0.09] \\
Near & posAP & +0.82 & +0.79 & +0.45 & [+0.32, +0.56] \\
\bottomrule
\end{tabular}

  }
\end{table}

\subsection{Geometric information injection}
\label{app:interventions}

The experiment covers the complete ScanNet two-view reconstruction set:
159 instances from 60 distinct scenes. GPT-5.6-Sol, GPT-6-Astra, Qwen3.8-Max,
and Gemini-3.8-Flash receive metric depth cues or relational cues appended to
the reconstruction prompt. Relational cues contain object depth order,
pairwise directions, and distance ranks, plus the camera motion with its
translation in metres. The baseline is the same prompt without added cues,
taken from the leaderboard predictions.

Table~\ref{tab:fact_injection_cross_model} reports the resulting changes. Both
cue types raise Structure ($+7.3$ to $+20.1$ points), while metric depth cues
raise Metric consistently across all four models ($+9.5$ to $+18.1$);
relational cues change Metric by only $-4.5$ to $+4.3$, even though the camera
translation conveys the scene scale.

\begin{table}[t]
  \centering
  \small
  \ifdefined\XeTeXversion
    \renewcommand{\bfdefault}{bx}
    \fontencoding{OT1}\fontfamily{cmr}\fontsize{10}{12}\selectfont
  \fi
  \caption{Score changes from geometric information injection relative to the
  same prompts without added cues, in percentage points.}
  \label{tab:fact_injection_cross_model}
  \fitwidth{%
    \begin{tabular}{llrr}
\toprule
Model & Supplied information & $\Delta$Structure & $\Delta$Metric \\
\midrule
GPT-5.6-Sol & Metric depth cues & +14.7 & +18.1 \\
 & Relational cues & +20.1 & $-$4.5 \\
\addlinespace
GPT-6-Astra & Metric depth cues & +7.3 & +9.5 \\
 & Relational cues & +11.4 & $-$2.1 \\
\addlinespace
Qwen3.8-Max & Metric depth cues & +9.5 & +13.9 \\
 & Relational cues & +9.9 & $-$2.9 \\
\addlinespace
Gemini-3.8-Flash & Metric depth cues & +11.7 & +16.8 \\
 & Relational cues & +17.3 & +4.3 \\
\bottomrule
\end{tabular}

  }
\end{table}

\paragraph{Depth and camera pose.}
Supplying both true depth and the second camera's pose to GPT-5.6-Sol raises
Metric to 63.6 and reduces mean object-position error from 0.65~m to 0.28~m.
Lateral placement, extent, orientation, and cross-object consistency still
require inference.

\subsection{Layout verification and revision}
\label{app:verification}

\begin{table}[t]
  \centering
  \small
  \caption{Layout verification and reconstruction revision. \textbf{(a)}
  Accuracy in selecting the rendered layout that matches the scene, with and
  without the photographs. \textbf{(b)} Scores of the original and revised
  reconstructions.}
  \label{tab:verification}
  \begin{tabular}[t]{@{}lrr@{}}
\multicolumn{3}{@{}l}{\textbf{(a)} Layout selection accuracy (\%)} \\
\toprule
Model & With photos & Without photos \\
\midrule
GPT-5.6-Sol & 95.0 & 72.5 \\
Claude-Opus-4.8 & 97.5 & 60.0 \\
Gemini-3.1-Pro & 97.4 & 59.5 \\
Gemini-3.5-Flash & 97.2 & 54.1 \\
\bottomrule
\end{tabular}
\hfill
\begin{tabular}[t]{@{}lrr@{}}
\multicolumn{3}{@{}l}{\textbf{(b)} Revision, GPT-5.6-Sol} \\
\toprule
Condition & Structure & Metric \\
\midrule
Original & 65.5 & 43.5 \\
Text-only revision & 65.9 & 44.0 \\
Rendered revision & 65.8 & 43.7 \\
\bottomrule
\end{tabular}

\end{table}

The layout probe evaluates GPT-5.6-Sol, Claude-Opus-4.8, Gemini-3.1-Pro,
and Gemini-3.5-Flash.
Models select which of two renderings matches the photographs: one retains
the reference geometry; the other changes object arrangement while preserving
sizes. Object names link photographs to rendered boxes, and perturbations
causing interpenetration or unsupported floating are excluded.
Table~\ref{tab:verification}(a) compares selection accuracy with and without
photographs; image evidence improves layout discrimination across all four
models.

For revision, GPT-5.6-Sol receives the original photographs and its predicted
scene JSON and is asked to correct the reconstruction; one
condition additionally provides a rendering of that JSON. Metric changes
only slightly from the original prediction ($+0.4$ points for text-only and
$+0.2$ for rendered revision), and adding the rendering changes
Metric by only $-0.3$ points (Table~\ref{tab:verification}(b)). Layout selection and revision are
separate tasks, testing error recognition and correction of the model's own
geometry, respectively.

\subsection{Navigation map injection}
\label{app:map_injection}

The study in Section~\ref{sec:map_injection} evaluates GPT-5.6-Sol,
GPT-6-Astra, Qwen3.8-Max, and Gemini-3.8-Flash on all 346 episodes, each model
serving as its own planner. Maps use the reconstruction task's object-centric
parameterization: IDs, categories, centres, dimensions, and yaw measured
counterclockwise from $+x$ to the width axis. In the self-map condition, the
model first selects navigation-relevant objects and predicts their boxes from
the episode observations, then plans from those boxes; GT boxes come from
EmbodiedScan~\citep{wang2024embodiedscan} annotations, with synthetic
room-boundary boxes of nominal height 2.5~m. The GT map leaves the model to
associate the visual target with a map entry, whereas GT + target ID
identifies that entry. Paths are evaluated
without endpoint rescaling, $\mathbf q_t=\widehat{\mathbf p}_t$, so the
supplied metric geometry must support planning directly; for \diagmodel{},
image-only success is 21.7\% under the scale-normalized leaderboard protocol
and 15.6\% here.

Table~\ref{tab:map_intervention_full} reports success and its components by
episode tier. GT maps improve success over self-generated maps for all four
models. Self-generated maps increase success for GPT-6-Astra, Qwen3.8-Max,
and Gemini-3.8-Flash, while GPT-5.6-Sol remains unchanged. Explicit target
identification raises success for GPT-5.6-Sol and Qwen3.8-Max, changes
GPT-6-Astra little, and lowers success for Gemini-3.8-Flash.

The outcome components show how the maps help. With images alone, failures
stem mainly from collisions: paths reach the target in 31--62\% of episodes
but are collision-free in only 16--29\%. GT maps raise both components above
85\% for three of the four models, whereas self-generated maps leave
GPT-5.6-Sol's success unchanged at 15.6\%. Target identification lifts reach
to 99--100\% for all models, so the remaining failures are collisions, which
explains the slight drop for Gemini-3.8-Flash. Detour episodes (Tier~2) are
not harder than human-traversal episodes once GT maps are supplied.

\begin{table}[t]
  \centering
  \small
  \ifdefined\XeTeXversion
    \renewcommand{\bfdefault}{bx}
    \fontencoding{OT1}\fontfamily{cmr}\fontsize{10}{12}\selectfont
  \fi
  \caption{Detailed 3D box map results on 346 offline navigation episodes
  per model, using unscaled metric paths. Tier 1 uses human traversals;
  Tier 2 requires algorithmically generated detours. Human/solver reference
  success is 97.7\%; a straight-to-goal path succeeds on 14.2\%. Scores are
  percentages; each model serves as its own planner in all four conditions.}
  \label{tab:map_intervention_full}
  \fitwidth{%
    \begin{tabular}{llrrrrr}
\toprule
Model & Input & Success & Reach & Collision-free & Tier 1 & Tier 2 \\
\midrule
GPT-5.6-Sol & Images & 15.6 & 41.3 & 27.7 & 18.7 & 11.8 \\
 & Self-map & 15.6 & 43.1 & 27.7 & 15.5 & 15.7 \\
 & GT map & 85.3 & 92.5 & 92.5 & 82.4 & 88.9 \\
 & GT + target ID & 90.5 & 100.0 & 90.5 & 85.5 & 96.7 \\
\addlinespace
GPT-6-Astra & Images & 22.5 & 61.8 & 29.2 & 21.8 & 23.5 \\
 & Self-map & 35.5 & 66.2 & 49.4 & 37.8 & 32.7 \\
 & GT map & 97.1 & 99.4 & 97.4 & 95.3 & 99.3 \\
 & GT + target ID & 97.4 & 99.7 & 97.7 & 95.3 & 100.0 \\
\addlinespace
Qwen3.8-Max & Images & 9.5 & 31.2 & 16.2 & 9.8 & 9.2 \\
 & Self-map & 11.0 & 34.1 & 21.4 & 12.4 & 9.2 \\
 & GT map & 66.2 & 94.2 & 68.8 & 59.1 & 75.2 \\
 & GT + target ID & 78.6 & 100.0 & 78.6 & 75.1 & 83.0 \\
\addlinespace
Gemini-3.8-Flash & Images & 11.8 & 38.4 & 21.7 & 11.9 & 11.8 \\
 & Self-map & 18.2 & 43.9 & 28.6 & 18.1 & 18.3 \\
 & GT map & 82.1 & 94.8 & 85.3 & 77.7 & 87.6 \\
 & GT + target ID & 79.8 & 99.1 & 80.1 & 71.5 & 90.2 \\
\bottomrule
\end{tabular}

  }
\end{table}

\subsection{Spatial fine-tuning settings}
\label{app:training_benchmarks}

We fine-tune Qwen3.6-35B-A3B separately on 10k, 20k, and 50k examples spanning
Geometric Perception, Spatial Relations, and scene reconstruction, without
general training data. Reasoning traces and answers are generated by
Qwen3.8-Max. Multiple-choice examples are selected by answer correctness,
while reconstruction examples are selected by ranking their reconstruction scores.
Each run starts from the same checkpoint and trains for one epoch with a global
batch size of 512, a learning rate of $7\times 10^{-6}$ with cosine decay; the three runs share the same data recipe and differ only in the
number of examples. We follow source-level training/evaluation splits. All
variants share decoding settings (temperature 0.6 and top-$p$ 0.95) and a
common local inference pipeline.

Spatial evaluations include SITE-Bench and SITE-Bench-Video~\citep{wang2025site},
BLINK~\citep{fu2024blink}, ViewSpatial-Bench~\citep{li2025viewspatial},
ERQA~\citep{gemini2025gemini_robotics}, OmniSpatial~\citep{jia2026omnispatial},
3DSRBench~\citep{ma20253dsrbench}, MMSI-Bench~\citep{yang2026mmsi},
EmbSpatialBench~\citep{du2024embspatial}, and RefSpatialBench~\citep{zhou2025roborefer}.

General evaluations cover the MMMU validation split~\citep{yue2024mmmu},
MMBench-CN v1.1~\citep{liu2024mmbench}, VisualPuzzles with direct
answers~\citep{song2025visualpuzzles}, BabyVision~\citep{chen2026babyvision},
the CharXiv reasoning task~\citep{wang2024charxiv}, AI2D~\citep{kembhavi2016diagram},
RealWorldQA~\citep{xai2024realworldqa}, and HallusionBench~\citep{guan2024hallusionbench}.

\clearpage
\section{Qualitative Case Studies}
\label{app:case_studies}

We show individual predictions to illustrate the patterns behind the aggregate
scores. The cases were chosen as illustrations, including failures of the
strongest model, and do not estimate how often each error occurs. All
reconstructions are plotted in their raw metric coordinates with shared axes.
Section~\ref{sec:case_studies} compares the three reconstruction domains;
the figures below provide larger versions of these examples.
Table~\ref{tab:case_ids} lists the instance identifiers of all cases.

\begin{table}[h]
  \centering
  \small
  \caption{Case identifiers.}
  \label{tab:case_ids}
  \begin{tabular}{@{}ll@{}}
    \toprule
    Case & Instance identifier \\
    \midrule
    Indoor (Fig.~\ref{fig:case_indoor_detailed}) & \path{scannet8f/scene0628_02_tour8} \\
    Tabletop (Fig.~\ref{fig:case_tabletop}) & \path{tabletop/tless_000016_356_486} \\
    Outdoor (Fig.~\ref{fig:case_outdoor}) & \path{nusc_c_far8/3a064c48e47b489a8b442b8081f877b6} \\
    Tabletop failure (Fig.~\ref{fig:case_scale_failure}) & \path{tabletop/tless_000019_1_115} \\
    Navigation (Fig.~\ref{fig:case_navigation}) & \path{scene0047_00_ns_t2_8_01160_a1_s2} \\
    Navigation: lounge (Fig.~\ref{fig:case_navigation_long_stool}) & \path{scene0281_00_ns_t2_15_01760_a2_s2} \\
    Navigation: conference room (Fig.~\ref{fig:case_navigation_long_detour}) & \path{scene0291_02_ns_t1_1_00060_s2} \\
    \bottomrule
  \end{tabular}
\end{table}

\paragraph{Indoor.}
The eight-view indoor example compares the full input sequence with the
reference geometry and each model's reconstruction. Shared metric axes
expose global displacement that can be obscured by individually aligned
visualizations.

\begin{figure}[!htb]
  \centering
  \includegraphics[width=\linewidth]{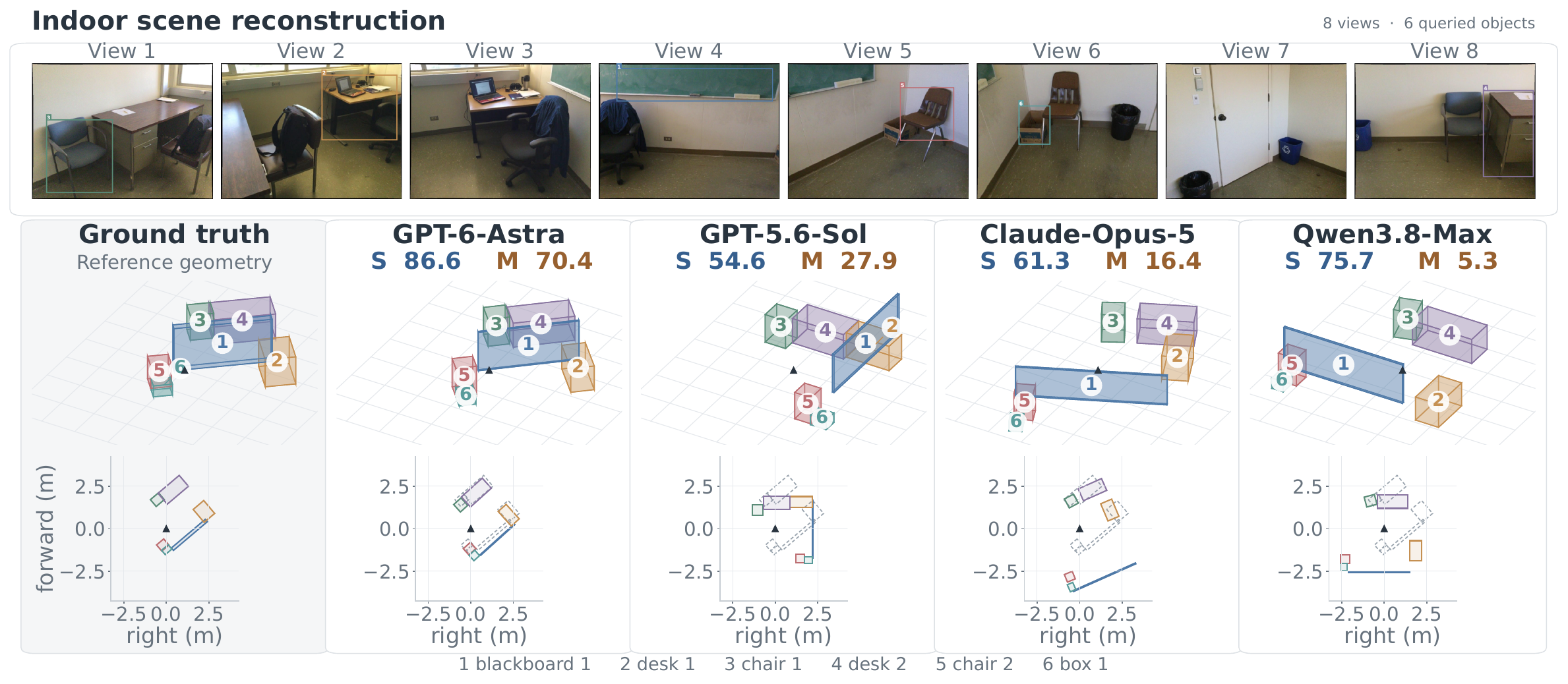}
  \caption{Indoor case with all eight input views. Object colors and
  identifiers match across photographs, 3D boxes, and ground-plane footprints;
  dashed outlines show the ground truth.}
  \label{fig:case_indoor_detailed}
\end{figure}

\paragraph{Tabletop.}
An elevated and a near-horizontal view show eight industrial objects. All four
models recover a plausible arrangement (Structure 65.3--79.6), but only
GPT-6-Astra also places the objects near their true positions (Metric 69.3);
the other three models fall to 15.7--26.9 on Metric.

\begin{figure}[!htb]
  \centering
  \includegraphics[width=\linewidth]{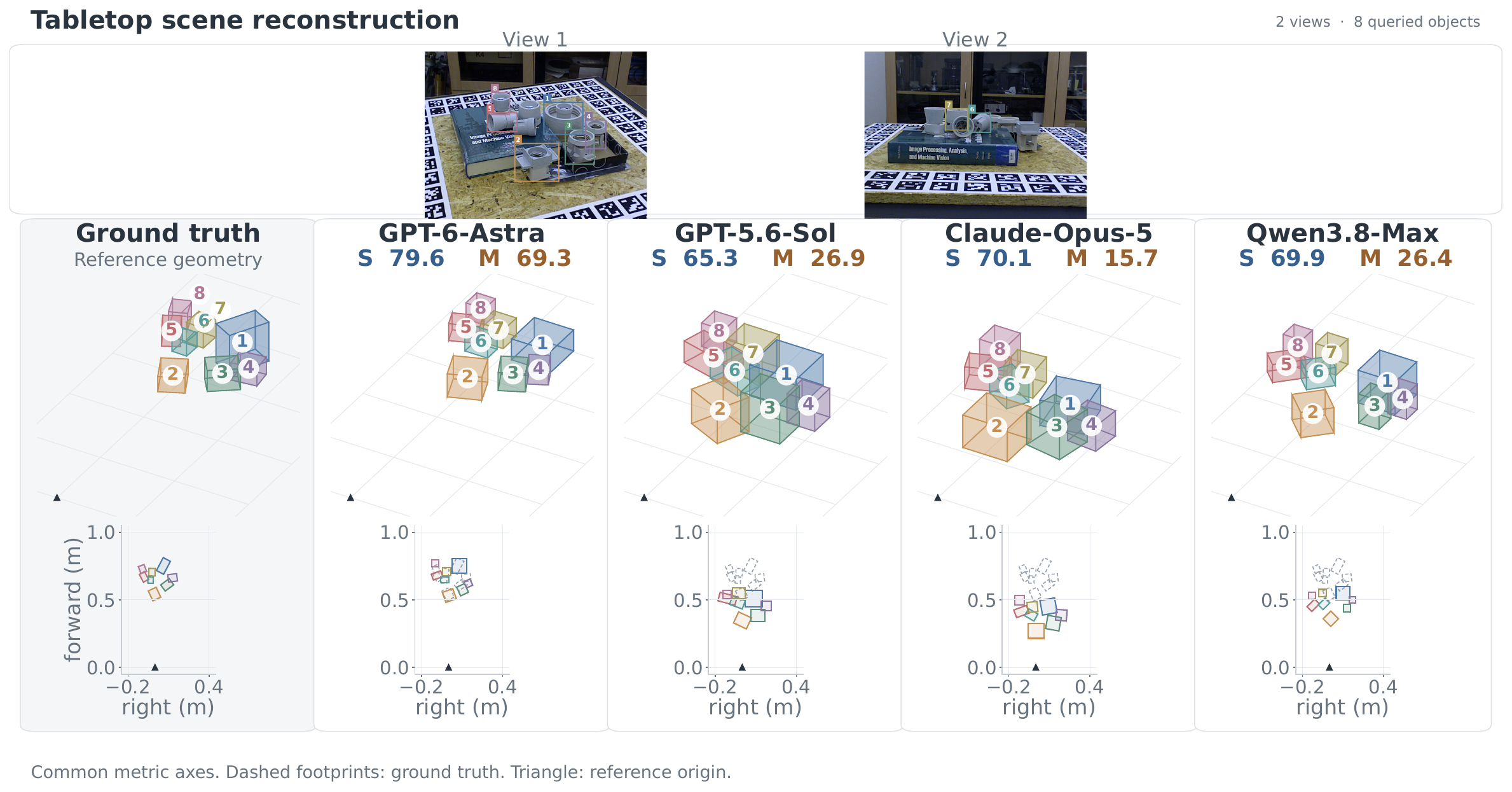}
  \caption{Tabletop case: eight T-LESS objects seen from an elevated and a
  near-horizontal camera. Numbers mark the same objects in the inputs and
  reconstructions; dashed outlines show the ground truth.}
  \label{fig:case_tabletop}
\end{figure}

\paragraph{Outdoor.}
Six surround cameras observe a row of distant barriers. The models keep the
barriers in the right order, but their absolute positions are off by metres:
Qwen3.8-Max scores 86.1 on Structure yet only 24.9 on Metric, with a mean
position error of 9.02~m.

\begin{figure}[t]
  \centering
  \includegraphics[width=\linewidth]{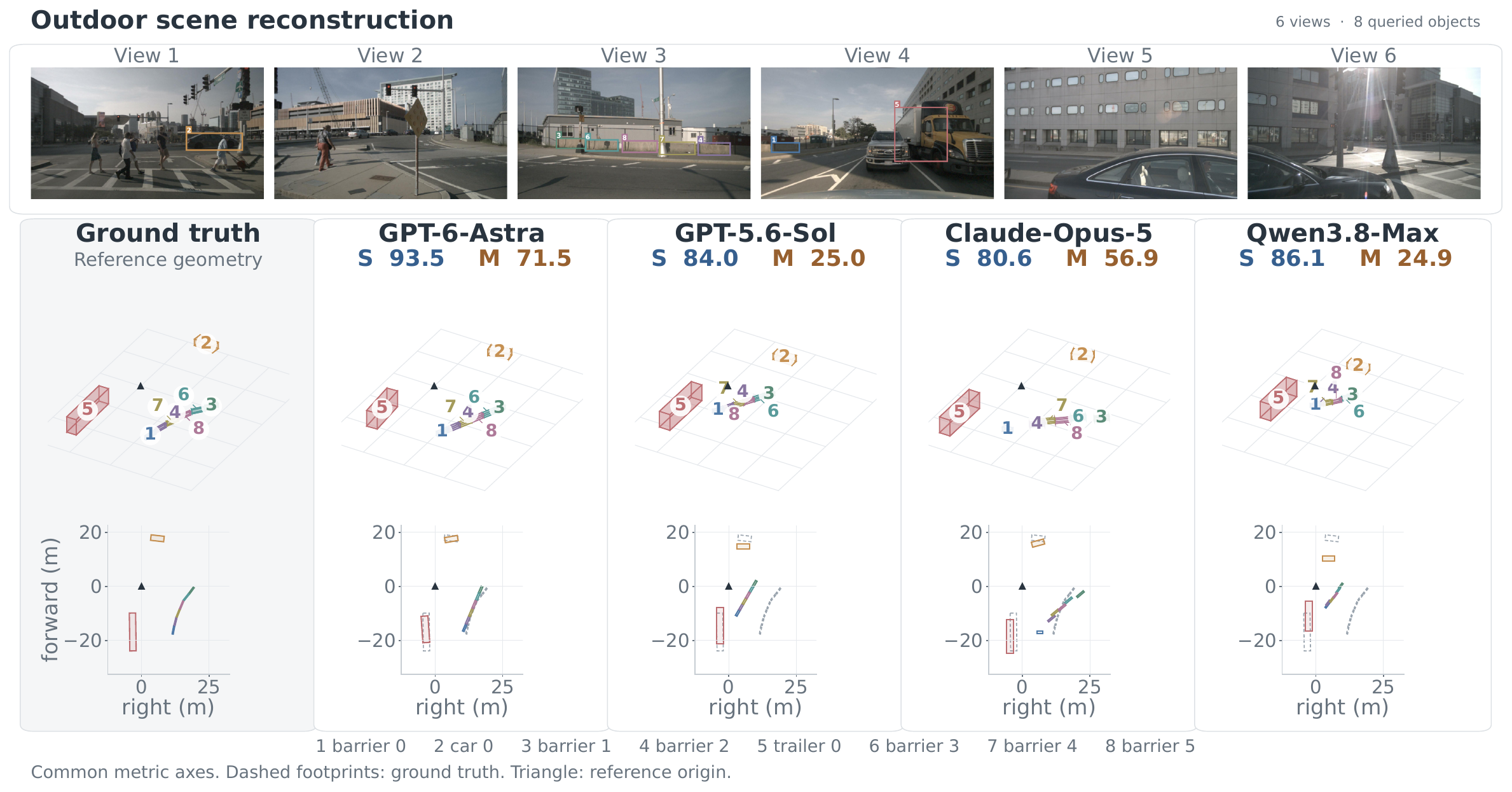}
  \caption{Outdoor case with all six surround-camera inputs. The barriers
  keep a recognizable order across models, while their absolute positions
  differ.}
  \label{fig:case_outdoor}
\end{figure}

\begin{figure}[p]
  \centering
  \includegraphics[width=\linewidth]{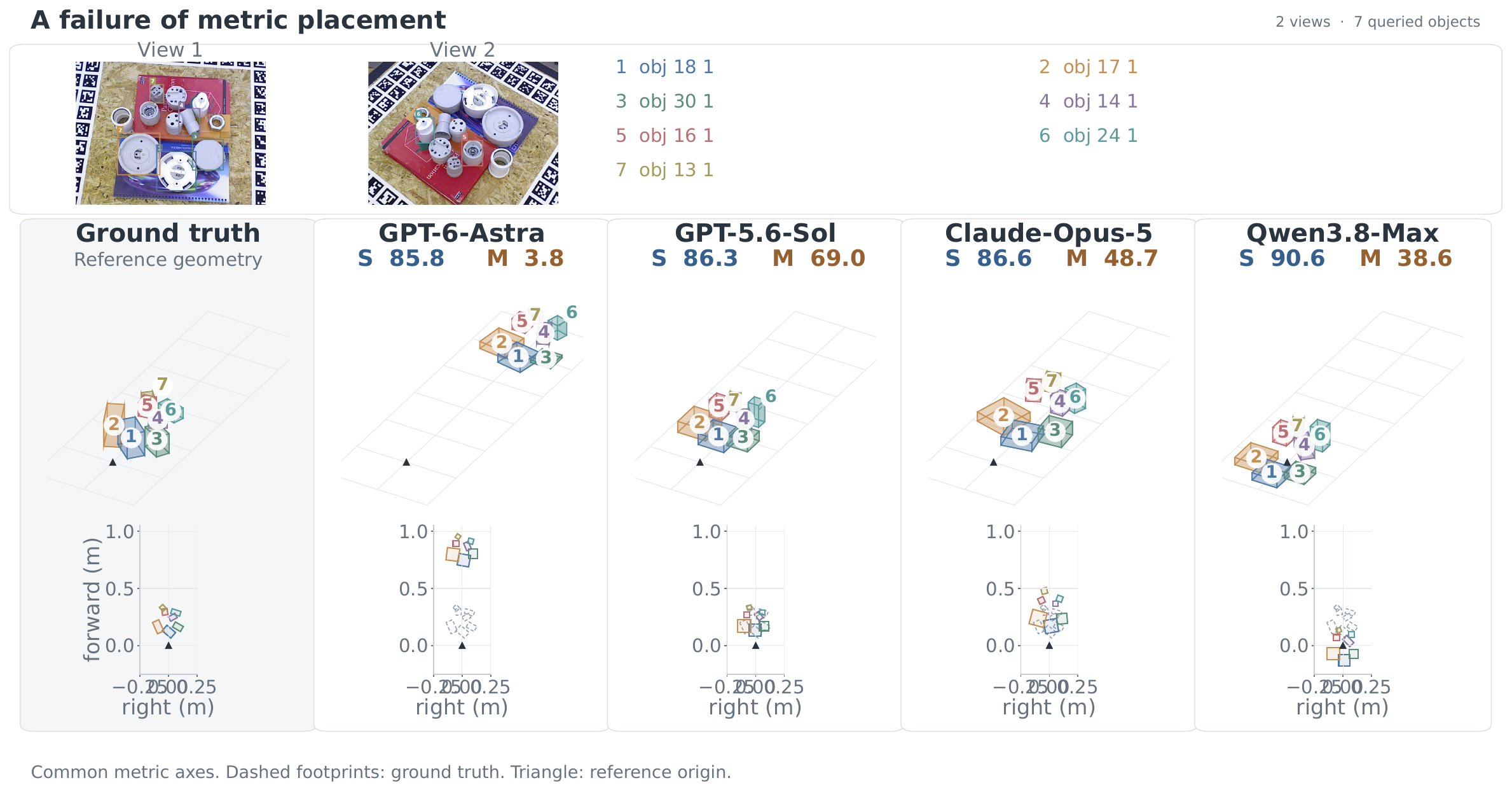}
  \caption{Tabletop failure case. GPT-6-Astra keeps the relative layout but shifts
  the whole group away from the camera; GPT-5.6-Sol stays close to the ground
  truth.}
  \label{fig:case_scale_failure}
\end{figure}

\paragraph{A failure of the strongest model.}
In a second tabletop scene, GPT-6-Astra arranges all seven objects correctly relative to one another
(Structure 85.8) but places the whole group too far from the camera
(Metric 3.8), whereas GPT-5.6-Sol places them much closer to their true
positions (Figure~\ref{fig:case_scale_failure}).

\begin{figure}[p]
  \centering
  \includegraphics[width=\linewidth]{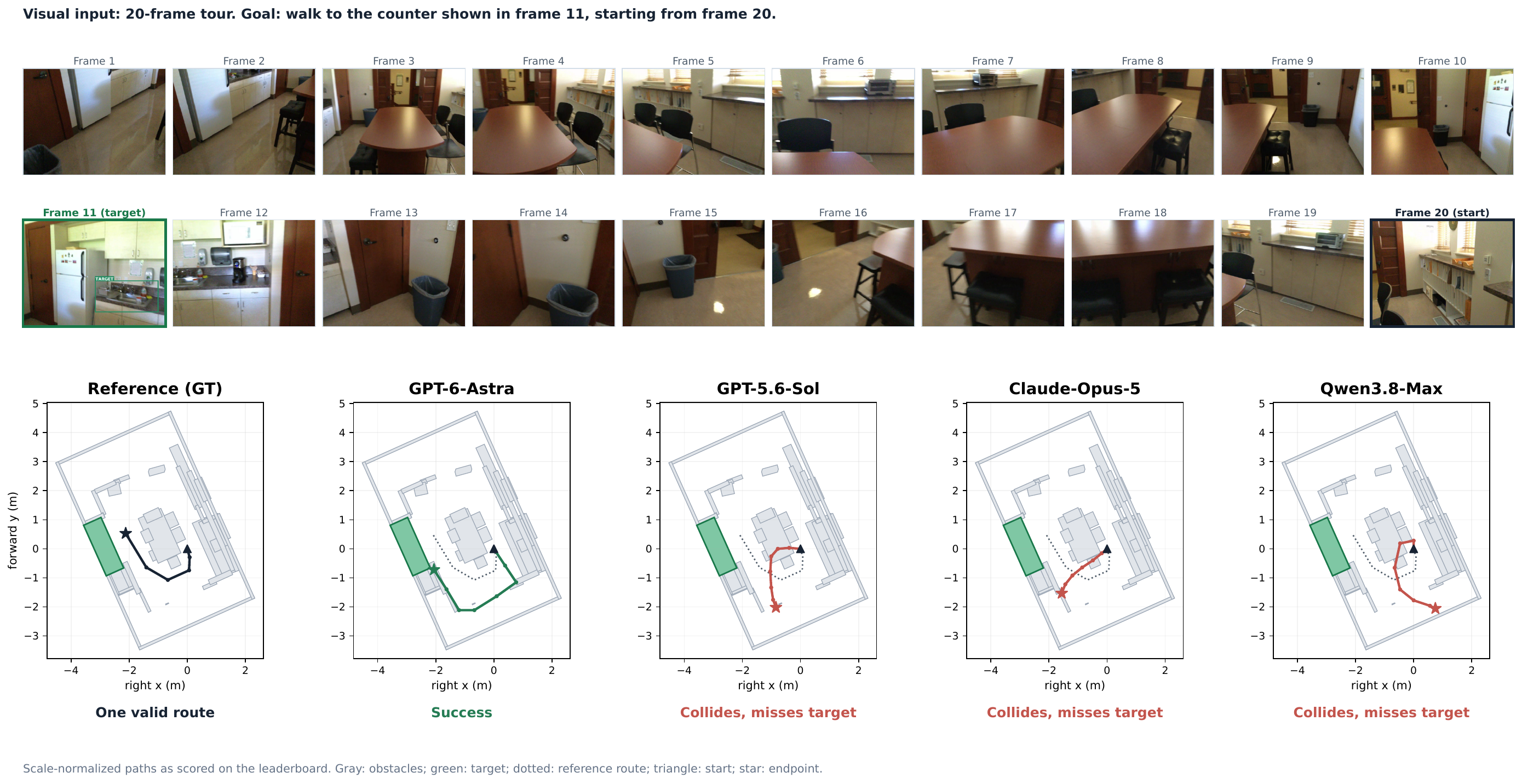}
  \caption{Navigation case requiring a detour. Top: the 20-frame visual input,
  with the start and target frames outlined. Bottom: a valid reference route and
  the paths of four models as scored on the leaderboard. Gray polygons are
  obstacles and the green polygon is the target.}
  \label{fig:case_navigation}
\end{figure}

\paragraph{Navigation.}
In this episode the counter can only be reached by detouring around a central
cluster of tables (Figure~\ref{fig:case_navigation}). GPT-6-Astra takes the
detour and arrives without collision, whereas the other three models cut
through the furniture and stop well short of the counter.

\begin{figure}[!htb]
  \centering
  \begin{subfigure}{\linewidth}
    \centering
    \includegraphics[width=\linewidth]{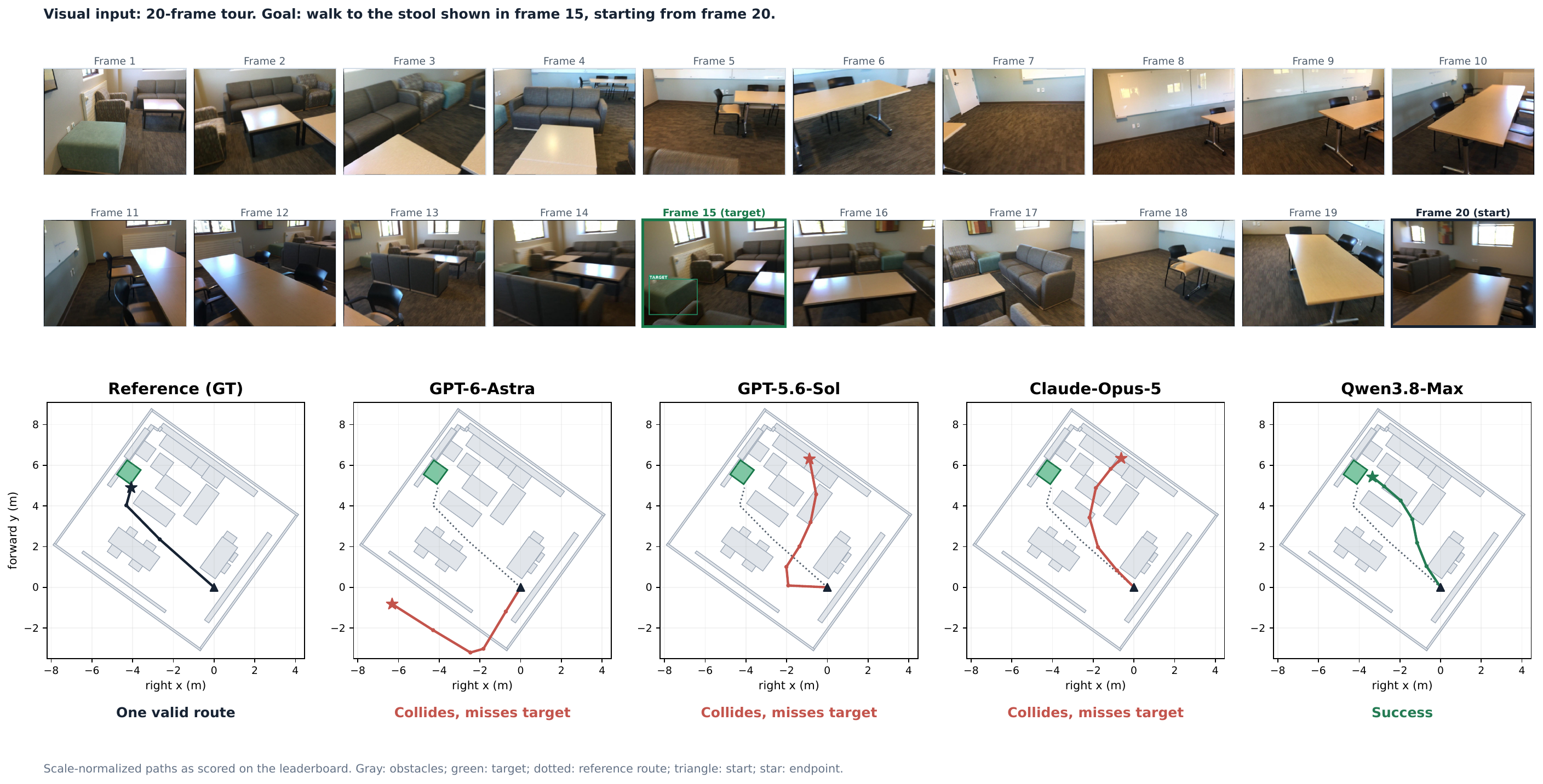}
    \caption{Lounge: 6.81~m reference route and 6.37~m start-to-end displacement.}
    \label{fig:case_navigation_long_stool}
  \end{subfigure}
  \par\vspace{8pt}
  \begin{subfigure}{\linewidth}
    \centering
    \includegraphics[width=\linewidth]{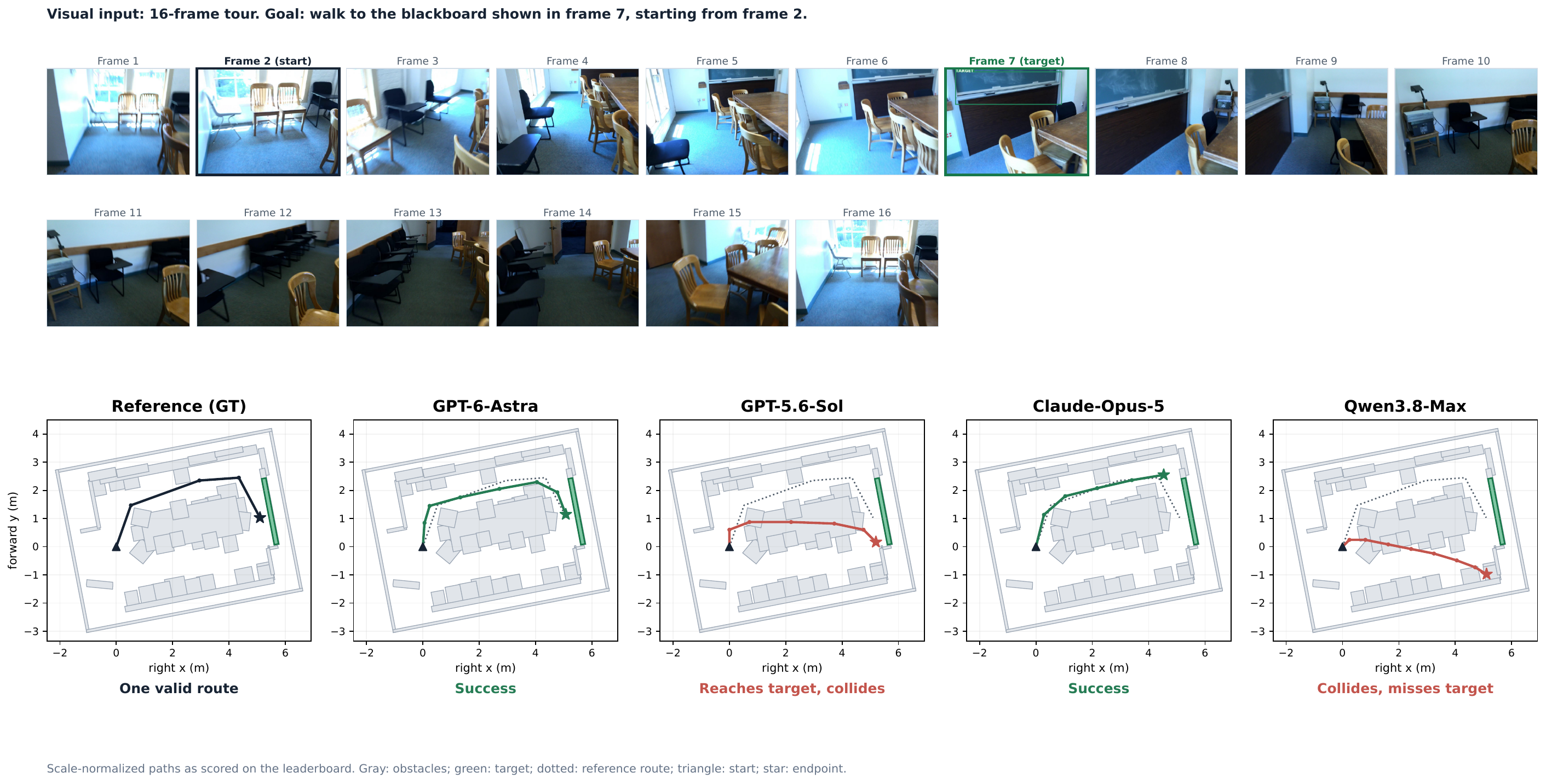}
    \caption{Conference room: 7.15~m reference route and 5.20~m start-to-end displacement.}
    \label{fig:case_navigation_long_detour}
  \end{subfigure}
  \caption{Longer navigation cases, with all 20 input frames in (a) and all
  16 in (b). Trajectories use the same scale normalization and scoring as the
  leaderboard. Gray polygons denote obstacles, green polygons denote the
  target, and dotted lines show a valid reference route.}
  \label{fig:case_navigation_long}
\end{figure}

\clearpage
\paragraph{Longer routes.}
Two longer traversals expose complementary errors
(Figure~\ref{fig:case_navigation_long}). In the lounge, Qwen3.8-Max reaches
the stool without collision, while the other models miss the target and
intersect obstacles. This reverses the outcome of the shorter counter case,
illustrating that reliable target grounding remains scene-dependent. In the
conference room, GPT-6-Astra and Claude-Opus-5 follow the aisle around the
central table. GPT-5.6-Sol reaches the blackboard's vicinity but crosses the
furniture, whereas Qwen3.8-Max both crosses the furniture and misses the goal.
The latter example shows why arrival and collision avoidance must be assessed
together: a plausible endpoint can conceal an infeasible route. Across the
longer traversals, models must preserve both the target location and the
obstacle layout over the full path.

\end{document}